\documentclass{article}

\PassOptionsToPackage{numbers, compress}{natbib}

\usepackage[final]{neurips_2022}

\usepackage[utf8]{inputenc} %
\usepackage[T1]{fontenc}    %
\usepackage[hidelinks]{hyperref}  
\usepackage{url}            %
\usepackage{booktabs}       %
\usepackage{amsfonts}       %
\usepackage{nicefrac}       %
\usepackage{microtype}      %
\usepackage{xcolor}         %
\usepackage{mathtools}

\usepackage{multibib}
\newcites{appendix}{Additional References}

\usepackage[nameinlink]{cleveref}

\usepackage{tikz} %
\usepackage{rotating}
\usepackage{floatrow}
\usepackage{placeins}
\usepackage{subcaption}
\usepackage{wrapfig}
\usepackage{caption}
\crefname{section}{Section}{Sections}
\crefname{subsection}{Section}{Sections}
\crefname{subsubsection}{Section}{Sections}
\crefname{figure}{Figure}{Figures}
\crefname{table}{Table}{Tables}
\crefname{subfigure}{Figure}{Figures}
\crefname{equation}{Equation}{Equations}
\crefname{appendix}{Appendix}{Appendix}
\crefname{algorithm}{Algorithm}{Algorithms}

\newif\iftodos
\todosfalse

\iftodos
    \newcommand\todo[1]{\textcolor{orange}{\textbf{TODO} #1}}
    \newcommand\ar[1]{\textcolor{teal}{\textbf{Aditya:} #1}}
    \newcommand\lk[1]{\textcolor{teal}{\textbf{Louis:} #1}}
    \newcommand\svs[1]{\textcolor{blue}{\textbf{Sjoerd:} #1}}
\else
    \newcommand\todo[1]{}
    \newcommand\ar[1]{}
    \newcommand\lk[1]{}
    \newcommand\svs[1]{}
\fi

\newcommand\fullapproach{random curiosity with general value functions}
\newcommand\approach{RC-GVF}

\usepackage{mathtools}

\newcommand{\psreturnfull}[1]{G_{#1}^z \left ( \lambda_z \right)}
\newcommand{\psreturn}{G_t^z}

\title{Exploring through Random Curiosity \\ with General Value Functions}

\author{
  Aditya Ramesh$^{1}$ ~ Louis Kirsch$^{1}$ ~ Sjoerd van Steenkiste$^{2}$\thanks{Part of this work was done while the author was a Postdoctoral researcher at IDSIA.} ~ J\"urgen Schmidhuber$^{1,3,4}$\\
  $^1$The Swiss AI Lab (IDSIA), University of Lugano (USI) \& SUPSI\\
  $^2$Google Research\\
  $^3$AI Initiative, King Abdullah University of Science and Technology (KAUST)\\
  $^4$NNAISENSE\\
  \texttt{\{aditya, louis, juergen\}@idsia.ch, svansteenkiste@google.com} 
}

\begin{document}

\maketitle

\begin{abstract}
Efficient exploration in reinforcement learning is a challenging problem commonly addressed through \emph{intrinsic} rewards.
Recent prominent approaches are based on \emph{state novelty} or variants of \emph{artificial curiosity}.
However, directly applying them to partially observable environments can be ineffective and lead to premature dissipation of intrinsic rewards.
Here we propose {\fullapproach} ({\approach}), a novel intrinsic reward function that draws upon connections between these distinct approaches.
Instead of using only the current observation’s novelty or a curiosity bonus for failing to predict precise environment dynamics, {\approach} derives intrinsic rewards through predicting temporally extended general value functions.
We demonstrate that this improves exploration in a hard-exploration diabolical lock problem.
Furthermore, {\approach} significantly outperforms previous methods in the absence of ground-truth episodic counts in the partially observable MiniGrid environments.
Panoramic observations on MiniGrid further boost RC-GVF's performance such that it is competitive to baselines exploiting privileged information in form of episodic counts.
\end{abstract}

\section{Introduction}
\label{intro}
How should an agent efficiently explore environments with high-dimensional state spaces and sparse rewards~\citep{thrun1992efficient, amin2021survey}?
An \emph{extrinsic} reward-maximising agent will make little progress until it stumbles upon a rewarding sequence of actions.
In the literature, this issue is commonly addressed by providing additional \emph{intrinsic} reward to guide the agent's exploratory behaviour~\citep{schmidhuber1990making,chentanez2004intrinsically,pathak2017curiosity}.

One class of prominent approaches rely on \emph{state novelty}, where an intrinsic reward in the form of a `novelty bonus' is awarded based on how often a state has been visited~\citep{sutton1990integrated, barto1991computational}.
More recent works have extended these approaches to high dimensional state spaces where tabular counts are inapplicable~\citep{bellemare2016unifying, ostrovski2017count, burda2018exploration}.
Another class of approaches is based on \emph{artificial curiosity}, where agents are rewarded in proportion to the prediction errors or information gains of a predictive world model~\citep{schmidhuber1991curiosity, storck1995reinforcement}.
Curiosity-based techniques have also been scaled up to handle larger state spaces \citep{houthooft2016vime, pathak2017curiosity}.\looseness=-1

\begin{wrapfigure}{r}{0.5\textwidth}
\begin{subfigure}{.99\textwidth}
  \centering
  \includegraphics[width=0.8\textwidth]{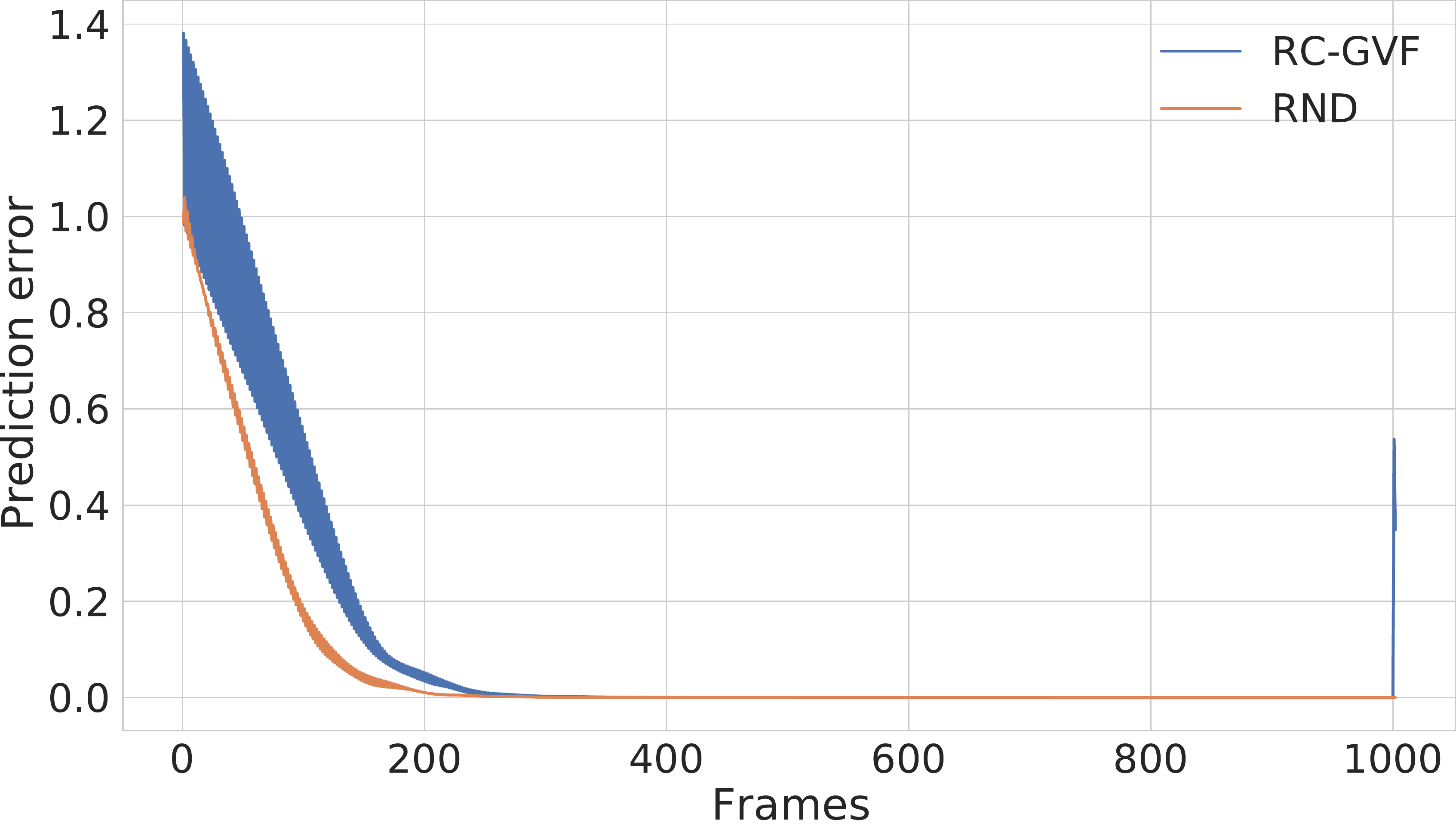}
  \label{fig:altcorr:results}
\end{subfigure}
\begin{subfigure}{.99\textwidth}
  \centering
   \vspace{5pt}
  \includegraphics[width=0.9\textwidth]{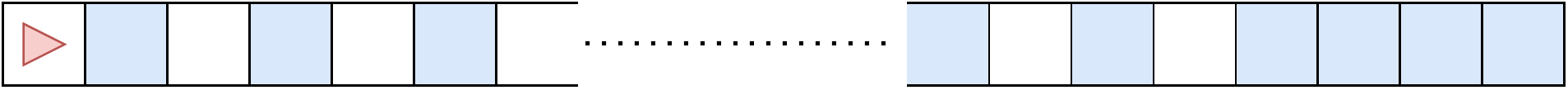}
  \label{fig:altcorr:env}
\end{subfigure}
\caption{
The alternating tile corridor environment.
The agent (red triangle) observes the colour of the tile that it currently occupies (white or blue). 
The extrinsic reward is always zero.
At each time-step (frame), the agent moves one tile forward, until it reaches the last tile.
Unlike our approach (RC-GVF), RND does not generate an intrinsic reward when encountering the surprising last few tiles (step 1000).
}
\label{fig:altcorr}
\end{wrapfigure}

Expecting an agent to have access to the complete environment state is unrealistic.
In such \emph{partially observable} settings~\citep{aastrom1965optimal, kaelbling1998planning}, observations may look alike in different states, and the benefit of approximate state novelty bonuses as intrinsic rewards is unclear.
Consider the environment in \autoref{fig:altcorr} where an agent is required to traverse a series of blue and white tiles, yet it is only able to observe the current tile it is standing on.
The colours of the tiles alternate in a regular fashion until the end of the corridor, where there is a surprising sequence of consecutive blue tiles.
Directly applying state novelty approaches may not be particularly meaningful in this case, as can be seen when using random network distillation (RND; \citep{burda2018exploration}).
In RND, the agent receives a novelty bonus based on the error of its predictions about the output of a fixed randomly initialised neural network applied to the current observation.
As shown in the top panel of \autoref{fig:altcorr}, RND is unable to ascribe novelty when the state pattern changes later in the sequence and the intrinsic reward has vanished. 
Similarly, intrinsic rewards may also vanish prematurely for certain curiosity-based approaches when single-step dynamics are simple \citep{parisi2021interesting}, while longer horizon predictions (as needed in the partially observable case) are susceptible to rapidly compounding errors during sequential rollouts~\citep{talvitie2014model}.

In this paper, we explore a connection between state novelty and artificial curiosity to address
these limitations.
We propose \textit{random curiosity with general value functions (RC-GVF)} as a novel approach to generating intrinsic rewards.
It takes inspiration from exploration through temporally extended experiments that summarise long observation sequences~\citep{Schmidhuber:97interesting}.
Our approach uses random general value functions (GVFs)~\citep{sutton2011horde} to pose questions about the cumulative future value of observation-dependent features.
Specifically, we train an ensemble of predictors to minimise the temporal difference (TD) errors of the general value functions and derive an intrinsic reward based on the TD-errors and disagreements in the long-term predictions.
The effective horizon of the prediction task can be controlled via the discount factor, where a zero discount can be related to state novelty with random network distillation (RND)~\citep{burda2018exploration}.
We hypothesise that predicting information from an temporally extended horizon improves exploration in POMDPs and guards against premature vanishing of intrinsic rewards due to the increase in difficulty of the auxiliary task.
Indeed, in our toy experiment of \autoref{fig:altcorr} it can be seen how RC-GVF manages to generate a spike of intrinsic reward at the end of the sequence.

We evaluate our approach on sparse-reward and partially observable RL problems.
Our results on a hard exploration diabolical lock problem \citep{misra2020kinematic} and the Minigrid suite of environments \citep{gym_minigrid} reveal the benefits of considering extended horizon predictions in comparison to approaches that rely on immediate partial observations to generate exploration bonuses.
Furthermore, existing baselines in Minigrid are heavily reliant on privileged information about episodic state-visitation counts from the simulator, which is commonly used to scale the exploration bonus.
We demonstrate that RC-GVF succeeds in many environments \emph{without} the use of such episodic counts.

\section{Preliminaries}
\label{prelims}
\paragraph{Reinforcement learning in POMDPs}

We consider the scenario where an agent interacts with a Partially Observable Markov Decision Process (POMDP) with time steps $t \in \mathbb{N}$, with environment states $S_t \in \mathcal{S}$, observations $O_t \in \mathcal{O}$, actions $A_t \in \mathcal{A}$, extrinsic rewards $R_e : \mathcal{S} \rightarrow \left[ 0, 1 \right] $, and policies that map histories of different lengths to distributions over actions $\pi : \mathcal{H} \rightarrow  \Delta \left( \mathcal{A}  \right) $ where $H_t = O_{1:t} \in \mathcal{H}$~\citep{kaelbling1998planning}.
The agent only observes $O_t$ at time step $t$, which may not fully describe the MDP state, and thus it becomes necessary to condition the agent on the history $H_t$.

The learning objective is to find the optimal policy $\pi^*$ that maximizes the expected discounted return
\begin{equation}\label{eq:rl_objective}
    J(\pi) = \mathbb{E}_{\pi} \left [\sum_{k=0}^{\infty} \gamma^k R_e(S_k) \right],
\end{equation}
where $\gamma \in \left [ 0, 1 \right)$ is the discount factor and the expectation is over stochastic components in the environment and the policy. %

\paragraph{Random Network Distillation}
In random network distillation (RND) \citep{burda2018exploration} the agent receives a state novelty reward proportional to the error of making predictions $\hat z(o_t)$ where the targets are generated by a randomly initialized neural network $Z_{\phi} : \mathcal{O} \rightarrow \mathbb{R}^d$.
The RND intrinsic reward for an observation is given by 
\begin{equation}\label{eq:rnd}
    R_i(o_t) = \| Z_{\phi}\left( o_t \right) - \hat{z} \left ( o_t \right) \|_2.
\end{equation}
RND-like error signals can be a simple way to obtain effective uncertainty estimates about targets at a given input, which could subsequently be used for exploration in RL or detecting out of distribution samples \citep{ciosek2020conservative}. Beyond its simplicity, an appealing aspect of using RND is that it allows us to implicitly incorporate prior beliefs about useful rewards through the neural network architecture~\citep{osband2016deep}. 

\paragraph{General Value Functions}
A general value function (GVF) \citep{sutton2011horde} is defined by a policy $\pi$, a cumulant or pseudo-reward function $Z : \mathcal{O} \rightarrow \mathbb{R}$, and a discount factor $\gamma_z \in \left [ 0, 1 \right)$. It can be expressed as
\begin{align}\label{eq:gvf}
    v_{\pi, z}(o) = \mathbb{E}_{\pi} \left [\sum_{k=0}^{\infty}  \gamma_z^{k} Z(O_{t+k}) \vert O_t = o \right ].
\end{align}

General value functions extend the concept of predicting expected cumulative values to arbitrary signals beyond the reward.
They can be viewed as `answering' questions about cumulative quantities of interest under a particular policy $\pi$ and discount factor $\gamma_z$.
Predictions from GVFs have previously been used as features for state representation~\citep{schaul2013better} or to specify auxiliary tasks that assist in shaping representations for the main task~\citep{jaderberg2016reinforcement, veeriah2019discovery}.

\section{Random Curiosity with General Value Functions}
\label{method}

In the spirit of artificial curiosity \citep{Schmidhuber:97interesting}, we are interested in predicting long-term outcomes in the environment under a particular policy as a means for exploration. 
Our approach, \emph{RC-GVF}, rewards an agent for taking actions that lead to higher uncertainty about the future.
However, instead of predicting the entire sequence of future states, we consider a set of random numeric questions about the environment and capture uncertainty about their outcome using general value functions (GVFs).

\subsection{General value functions to predict the future}

The central piece of RC-GVF is the prediction of temporally extended outcomes in the environment.
This is different from only predicting a feature of the current state (as in state novelty exploration \citep{sutton1990integrated}) or dynamics models that predict the entire (sequence of) next observation(s) (as in early versions of artificial curiosity \citep{schmidhuber1991curiosity}).
At time step $t$, the current observation $o_t$ is mapped to a collection of pseudo-rewards $z_{t+1} \in \mathbb{R}^d$ which together with the policy and discount factor define a question that can be asked about the future: ``what is the expected discounted cumulative sum of these pseudo-rewards under a given policy?''.
The answer to this question (i.e. the prediction target) is based on outcomes in the environment, as in \autoref{eq:gvf}.
In this paper, we investigate whether a set of random pseudo rewards are sufficient to drive exploration in this way.

Previously it was found that random features extracted by a neural network are often sufficient to capture useful properties of the input~\citep{rahimi2008weighted}.
Similarly, neural network architectures can be used to express prior knowledge about useful features~\citep{ulyanov2018deep, ciosek2020conservative}.
To this end, we generate pseudo-rewards from a fixed and randomly initialised neural network $Z_{\phi} : \mathcal{O} \rightarrow \mathbb{R}^d$ with parameters $\phi$ that maps observations to pseudo-rewards.
This choice also bypasses the difficult problem of discovering meaningful general value functions~\citep{veeriah2019discovery}.\looseness=-1

\paragraph{Training the GVF predictors}
Our exploration mechanism is to reward the agent for taking actions that generate previously unknown outcomes. 
To facilitate this, we train a separate (recurrent) neural network--which we call the \textit{predictor}--to predict these values. 
Concretely, a predictor $\hat{v}_{\pi,z} : \mathcal{H} \rightarrow \mathbb{R}^d$ maps histories (of observations, actions, and pseudo rewards) to pseudo-values. 

The predictor is trained on-policy, implying that the GVFs are evaluated under the current policy. 
One motivating factor for this choice is that it couples the prediction task to the current policy, thus creating an incentive to vary behaviour for additional exploration \citep{flet2021adversarially}.

We use the (truncated) $\lambda$-return as the target for the predictor, which can be recursively expressed as
\begin{equation}\label{eq:lambda_return}
    \psreturnfull{t} = Z_{t+1} + \gamma_z (1 - \lambda_z) \hat{v}_{\pi, z} \left( H_{t+1} \right) 
    +\gamma_z \lambda_z \psreturnfull{t+1}.
\end{equation}
Here, $\lambda_z \in \left[0, 1 \right]$ is the usual parameter that allows balancing the bias-variance trade off by interpolating between TD(0) and Monte Carlo estimates of the pseudo-return~\citep{sutton1988learning}.
The predictor is trained to minimise the mean squared TD-error with the $\lambda$-return target.
For convenience in notation, we will denote $\psreturnfull{t}$ as $\psreturn$, despite its dependence on $\lambda_z$.

\subsection{Disagreement and prediction error as intrinsic reward}
To generate an intrinsic reward, a straightforward choice could be to consider the error between the temporal difference target (from \autoref{eq:lambda_return}) and the predictor's output at the current observation:
\begin{equation}\label{eq:td_error}
    L_{\textrm{TD}}(h_t) = \left [ \psreturn - \hat{v}_{\pi, z} \left ( h_t \right) \right ]^2.
\end{equation}
However, the presence of \emph{aleatoric} uncertainty is a problem that arises as a consequence of extending the horizon of the predictive task. 
From an exploration perspective, the agent should focus on the reducible \emph{epistemic} uncertainty, and not the irreducible aleatoric uncertainty~\citep{der2009aleatory, hullermeier2021aleatoric}.
Directly minimising \autoref{eq:td_error} with the predictor would ignore the inherent variance in the TD-target (due to stochasticity in the policy and environment), and thus using the prediction error of the GVF target as an intrinsic reward does not distinguish between aleatoric and epistemic uncertainty. 

Another way to recognize this is by decomposing the expected prediction error as described by \citet{jain2021deup}.
The expected loss of a predictor $\hat{f}(\cdot)$ at an input $o$, $\mathbb{E}  [ L ( t, \hat{f}(o) )]$, can be split into epistemic ($E$) and aleatoric ($A$) components:
$ \int L ( t, \hat{f}(o)) dP(t | o)= E ( \hat{f},  o ) + A ( o ),$
where $P(t | o)$ is the conditional distribution over targets for input $o$ and $L(t, \cdot)$ is a chosen loss function.
The aleatoric uncertainty at input $o$, $A \left (o \right)$, is defined as the expected prediction error of a Bayes optimal predictor function.
$E ( \hat{f},  o )$ denotes the epistemic uncertainty of a predictor at $o$.

To overcome this issue with aleatoric uncertainty in RC-GVF, we propose to train an \emph{ensemble} of predictors and utilise the variance across their predictions as a multiplicative factor on the prediction error. 
Concretely, we train $K$ predictors $\hat{v}_{\pi, z}^k, k \in \{1, 2 \dots K \}$.  
The prediction target for each member of the ensemble is the same $\lambda$-pseudo-return (Equation \ref{eq:lambda_return}) by using bootstrapped values from that member's predictions.
Using the ensemble of predictors, the intrinsic reward is given by
\begin{align}\label{eq:intrinsic_reward}
    R_{i}(o_t) &= \sum_{j=1}^d \left(\mathbb{E}[L_{\textrm{TD}}^k(h_t)] \odot \mathbb{V}[\hat{v}_{\pi, z_j}^k \left ( h_t \right )] \right)_j \\
    &= \sum_{j=1}^d  \left [ \frac{1}{K} \sum_{k=1}^{K} \left ( G_t^{z_j} - \hat{v}_{\pi, z_j}^k \left ( h_t \right ) \right )^2  \right ] \cdot \left [ \frac{1}{K-1} \sum_{k=1}^{K} \left ( \bar{v}_{\pi, z_j} \left ( h_t \right )- \hat{v}_{\pi, z_j}^k \left ( h_t \right ) \right )^2 \right ],
\end{align}
where $\odot$ corresponds to element-wise multiplication.
In this formulation, even when prediction error remains, the exploration bonus will vanish as the predictors converge to the same expectation.

\subsection{Effective horizon and relation to random network distillation}
The effective horizon over which predictions are considered depends on the choice of the discount factor $\gamma_z$.
It can be shown that for a horizon $H_{\gamma_z, \varepsilon} \in \mathbb{R}$ satisfying
\begin{equation}\label{eq:discount_horizon}
H_{\gamma_z, \varepsilon}  \geq \frac{\log \left ( \frac{1}{\varepsilon \left( 1-\gamma_z \right )} \right )}{\log \left ( \frac{1}{\gamma_z}\right )},
\end{equation}
the discounted sum of (pseudo) rewards beyond this horizon is bounded by $\varepsilon$ \citep{kearns2002sparse, kocsis2006bandit}.

From Equation \ref{eq:lambda_return} we can see that for the special case of $\gamma_z=0$ and any choice of $\lambda_z$, we have $G_t^z = Z_{t+1} = Z_{\phi}\left( o_t \right)$. 
Note that in this case the TD-error between prediction and target is equivalent to the intrinsic reward provided by random network distillation (RND; \autoref{eq:rnd}).
Since the prediction targets are now deterministic, there is no problem with aleatoric uncertainty, and a single predictor suffices (as opposed to an ensemble). 

\section{Related work}
\label{related_work}
\paragraph{Exploration}
In tabular RL, explicit \emph{counts} for states or state-action pairs can be maintained to enable exploration and achieve efficient learning~\citep{strehl2008analysis, auer2008near}.
Count-based bonuses have been scaled to larger state spaces through the use of density models, which provide pseudo-counts~\citep{bellemare2016unifying, ostrovski2017count}.
State novelty bonuses are similarly inspired heuristics that provide a bonus based on estimated novelty of the visited state~\citep{burda2018exploration}.

Perhaps the simplest way of implementing exploration through \emph{`curiosity'}  is to reward the policy for encountering transitions to states that surprise a learning predictor~\citep{schmidhuber1991curiosity}. 
This is suitable for deterministic environments but it suffers from the `noisy TV problem' in stochastic environments~\citep{schmidhuber2010formal, burda2018large}.
To overcome this limitation, intrinsic rewards have been defined by \emph{learning progress} (the first derivative of the prediction error) ~\citep{Schmidhuber:91singaporecur}, \emph{information gain}~\citep{storck1995reinforcement, houthooft2016vime}, or \emph{compression progress}~\citep{schmidhuber2007simple}. Also compare such bonuses to pseudo-counts~\citep{bellemare2016unifying}.
Alternatively, it is possible to mitigate sensitivity to noise by measuring prediction errors in a latent space \citep{Schmidhuber:97interesting, pathak2017curiosity}.

Our method is different from state count and novelty approaches in that it models temporally extended values beyond the current observation.
It includes a state-novelty measure, RND~\citep{burda2018exploration}, as a special case when the GVF discount factors are zero.
Unlike most artificial curiosity approaches, RC-GVF does not model the entire environment dynamics, in line with the idea of conducting temporally extended experiments whose outcomes are abstract summaries (represented in latent variables) of long observation sequences~\citep{Schmidhuber:02predictable}.
We mitigate sensitivity to aleatoric uncertainty by measuring the disagreement of an \emph{ensemble} of predictors, inspired by prior work (for single-step prediction models)~\citep{shyam2019model, pathak2019self}.
Ensembles for exploration have also been explored in the context of posterior sampling~\citep{strens2000bayesian}.
Bootstrapped-DQN~\citep{osband2016deep} uses an ensemble of Q-value functions for a model-free interpretation of posterior sampling~\citep{osband2016generalization}.
Unlike in our approach, these value functions are restricted to the original task rewards, and can not capture arbitrary pseudo-rewards.
Recent works suggest incorporating model-based planning to directly optimise for long-term novelty ~\citep{sun2011planning, sekar2020planning, lambert2022challenges}. Our intrinsic reward--which is generated by considering multiple future steps--could motivate approaches for optimising long-term novelty without explicit rollout-based planning.

\paragraph{General Value Functions}
\citet{sutton2011horde} proposed general value functions (GVFs) as an approach to represent predictive knowledge about the world. Closely related is the vector-valued adaptive critic \cite{Schmidhuber:91nips} which predicts and controls cumulative values of special input vectors. 
Several works have studied the impact of using auxiliary value functions to improve representation learning in RL \citep{jaderberg2016reinforcement, bellemare2019geometric, mcleod2021continual}.
Further, it was previously found that GVFs of randomly generated pseudo-rewards can also be useful for shaping and learning representations~\citep{lyle2021effect, zheng2021learning} (see also related work on forecasts~\citep{schaul2013better}, TD-networks~\citep{sutton2005temporal}, and predictive state representations~\citep{littman2001predictive}).
This can be viewed as `answering' questions about cumulative quantities of interest under a particular policy $\pi$ and discount factor $\gamma_z$.
In our implementation, the predictor is an entirely separate network and thus GVF prediction can not be used to improve the agent's internal representation.
Instead, we use the GVF prediction to derive a useful signal to guide exploration.\looseness=-1

The discovery/selection of useful GVF `questions' is an open problem~\citep{schaul2018barbados}.
Existing approaches learn questions related
to optimizing performance on the main task~\citep{veeriah2019discovery, kearney2021finding}.
However, such an approach might not be suitable for sparse-reward environments.
In our approach, we side-step the GVF discovery problem through the use of randomly generated pseudo-rewards and on-policy predictions.\looseness=-1
\section{Experiments}
\label{experiments}
In this section we present an empirical evaluation of RC-GVF.\footnote{Code is available at \url{https://github.com/Aditya-Ramesh-10/exploring-through-rcgvf}.}
First, we consider a hard exploration diabolical lock problem from the literature~\citep{misra2020kinematic} and demonstrate the benefits of predicting beyond the immediate observation.
We then proceed to experiments with the standard Minigrid suite of partially observable and procedurally generated environments, demonstrating how RC-GVF is able to outperform several strong baselines \citep{flet2021adversarially, zhang2021noveld}.\pagebreak

\subsection{Diabolical locks}

The diabolical lock problem \citep{misra2020kinematic} is considered a difficult environment for exploration due to noisy high-dimensional observations, stochastic dynamics and misleading 
rewards that deter the agent from finding the sparse optimal reward at the end of the lock.
In the literature, RND is often shown to fail on these types of environments due to only considering the immediate future~\citep{misra2020kinematic}.
Due to RC-GVF's policy-conditioned predictions of future quantities, this task is instructive to demonstrate the qualitative advantages of our method and how this results in markedly different performance.

\setlength\tabcolsep{3pt}
\begin{wraptable}{r}{0.5\textwidth}
    \vspace{-5pt}
    \footnotesize
    \centering
    \caption{
    Mean over 10 seeds and (min, max) of the farthest column reached (of $H=100$) at different points on the diabolical lock problem.}
    \label{tab:hgrid_rcgvf_locks}
    \begin{tabular}{lcccc}
      \toprule
      Frames & 5M & 10M & 20M \\
      \midrule
      RC-GVF &  61 (47, 81) & 94 (83, 100)& 100 (100, 100) \\
      RND & 28 (21, 41) & 37 (27, 47) & 46 (33, 58) \\
      \bottomrule
    \end{tabular}
    \vspace{-5pt}
    \label{tab:db_lock_results}
\end{wraptable}

\paragraph{Environment details}
The environment consists of states organised in 3 rows $\{a,b,c\}$ and $H$ columns from $\{1, 2, \dots H \}$ (see \autoref{fig:db_lock} in Appendix \ref{sec:appendix_envs:db_lock}).
In each episode, the agent is initialised in one of two possible starting states, either at $a_1$ or $b_1$.
At each state the agent has $L$ available actions, only one of which is `good' and transitions to a `good' state in the next column with equal probability. 
Taking the good action gives a negative (anti-shaped) reward of $-1/H$, except at the end of the lock (states $a_H$ or $b_H$), where the sparse optimal reward of $10$ is received.
The remaining $L-1$ actions at any state are `bad', causing a deterministic transition to the `dead' row $c$ at the next column (zero reward). 
All actions from a dead state lead to the dead state in the next row, and hence the optimal policy is to take the respective good actions in each state.
The good action at each state is assigned randomly as part of the MDP specification.
The agent never observes the state directly and only has access to a high-dimensional noisy observation (details in Appendix~\ref{sec:appendix_envs:db_lock}).\looseness=-1

In our experiment, we consider a problem with horizon $H=100$, $L=10$, and observation noise $\sigma_o = 0.1$.
As discussed by \citet{misra2020kinematic}, this exploration problem is hard because the probability of reaching the optimal reward through a uniformly random policy is $L^{-H}$ ($10^{-100}$ in our instance). 
Further, the stochastic transitions from good states prevent a solution that relies on memorising a state-independent successful sequence of $H$ actions independent of the state.\looseness=-1

\paragraph{Implementation}
We use Proximal Policy Optimization (PPO)~\citep{schulman2017proximal} in an actor-critic framework as the base agent and use 128 pseudo-rewards for RND and RC-GVF.
For RC-GVF we set $\gamma_z=0.6$ and use two prediction heads in the ensemble.
The agent is trained to maximize the expected sum of a weighted combination of intrinsic and extrinsic rewards.
Further details are in Appendix~\ref{sec:appendix_envs:db_lock}.

\paragraph{Results}
Table~\ref{tab:db_lock_results} compares RND to RC-GVF in this environment. 
It can be seen how {\approach} succeeds in completing the lock on all runs, while the best run of RND only reaches $60\%$ of the lock (and does not complete it).
Indeed, once the agent takes an incorrect action and `falls' into the bottom row, the future consequences (and thus, the GVFs) become highly predictable. 
By staying alive, the agent will encounter new observations and unpredictable transitions which appear interesting.
Further experiments with additional baselines are presented in Appendix \ref{sec:appendix:db_lock_analysis}.
This result is illustrative of RC-GVFs exploration capabilities as we will see next.\looseness=-1

\subsection{MiniGrid}

We evaluate {\approach} on procedurally generated environments from MiniGrid~\citep{gym_minigrid}, which is a standard benchmark in the deep reinforcement learning literature for exploration~\citep{raileanu2020ride, campero2020learning, flet2021adversarially, zhang2021noveld, zha2021rank, parisi2021interesting}.
Exploration in these environments is challenging due to partial observability, extremely sparse rewards, and the procedural generation of mazes and objects.\looseness=-1

\paragraph{Environment Details}
Broadly speaking, we will consider three classes of environments organised according to several difficulty levels. 
First, we study \emph{MultiRoom-N7-S8} and \emph{MultiRoom-N12-S10}, which are navigation tasks in a maze with seven and twelve rooms respectively.
Next, we consider two levels of difficulty in \emph{KeyCorridor-S4R3} and \emph{KeyCorridor-S5R3}. Here, the agent needs to pick up a ball behind a locked door.
Finally, we examine the \emph{ObstructedMaze} set of  environments where the agent has a similar task of picking up a ball behind a locked door, but keys are hidden in boxes (\emph{ObstructedMaze-2Dlh}), and doors can be obstructed (\emph{ObstructedMaze-2Dlhb}).
The observations corresponds to an egocentric view of the cells in the front of the agent.
Further details of the MiniGrid environments are available in \cref{sec:appendix_envs:minigrid}.

\paragraph{Implementation}
We compare {\approach} to RND~\citep{burda2018exploration}, AGAC~\citep{flet2021adversarially}, and NovelD~\citep{zhang2021noveld}.
We use Proximal Policy Optimization (PPO)~\citep{schulman2017proximal} as our base agent for all approaches.
The agent is trained to maximize the expected sum of a weighted combination of intrinsic and extrinsic rewards.
At each time step $t$, the agent receives a reward $R_t = R_e(s_t) + \beta R_i(o_t) $, where $\beta \in \mathbb{R}_+$ balances the contribution of the extrinsic $R_e(s_t)$ and intrinsic $R_i(o_t)$ rewards.
For RC-GVF we set $\gamma_z=0.6$, use two prediction heads in the ensemble of predictors and use 128 pseudo-rewards (same as for RND).
Other important hyper-parameters, such as the intrinsic reward coefficient ($\beta$), entropy coefficient, and learning rate of the predictor are obtained via an extensive hyper-parameter search for all baselines (see Appendices \ref{sec:appendix_implementation:baselines} and \ref{sec:appendix_implementation:hyperparameters} for details, including on our implementation of baselines.).\looseness=-1

We present results averaged over 10 independent runs for each approach in every figure (solid line). 
Unless mentioned otherwise, the shading indicates 95\% bootstrapped confidence intervals.

\begin{figure}[t]
\begin{subfigure}{.32\textwidth}
  \centering
  \includegraphics[width=0.99\textwidth]{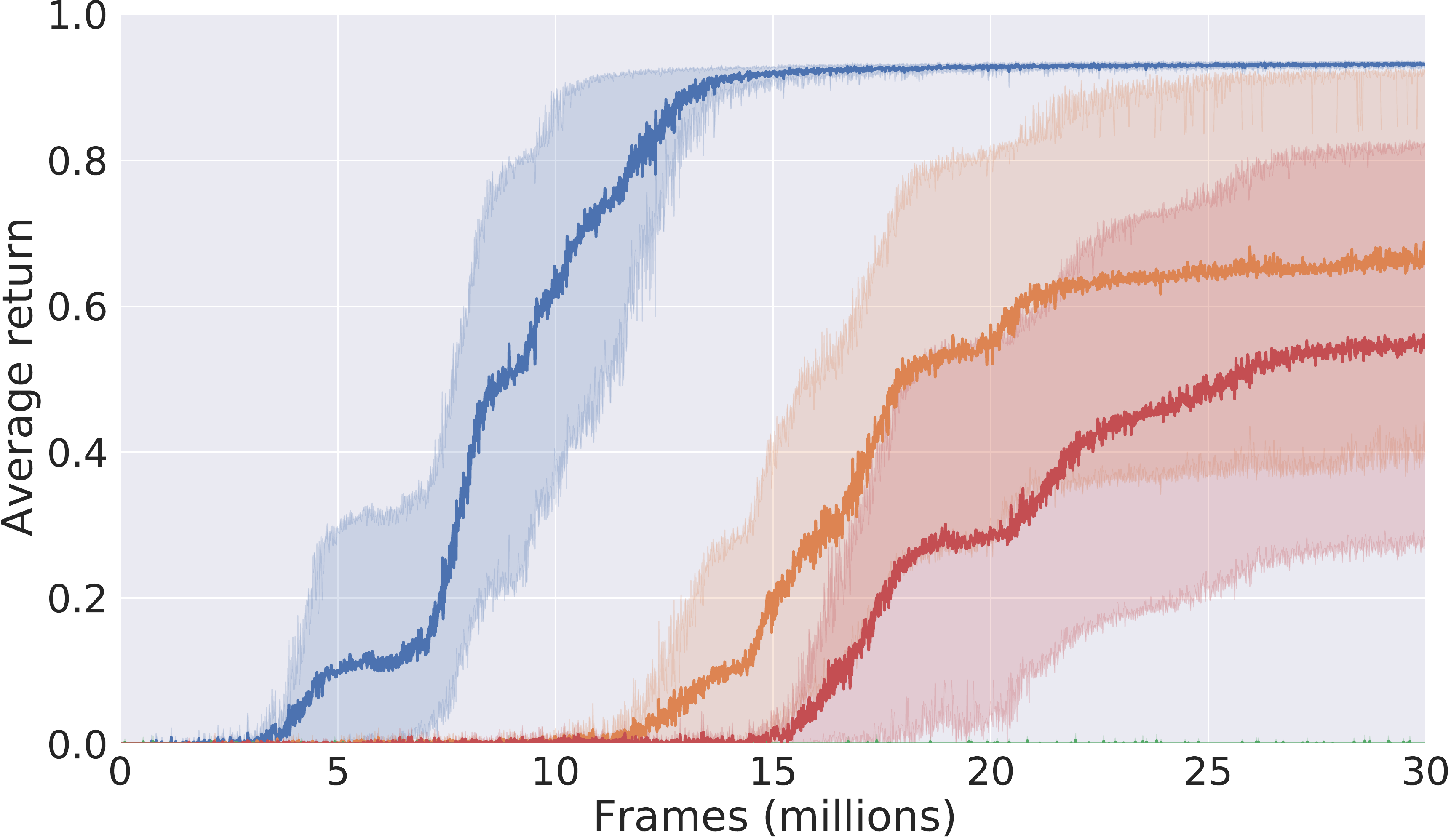}
  \caption{KeyCorridor-S4R3}
  \label{fig:minigrid_results:kcs4r3}
\end{subfigure}
\begin{subfigure}{.32\textwidth}
  \centering
  \includegraphics[width=0.99\textwidth]{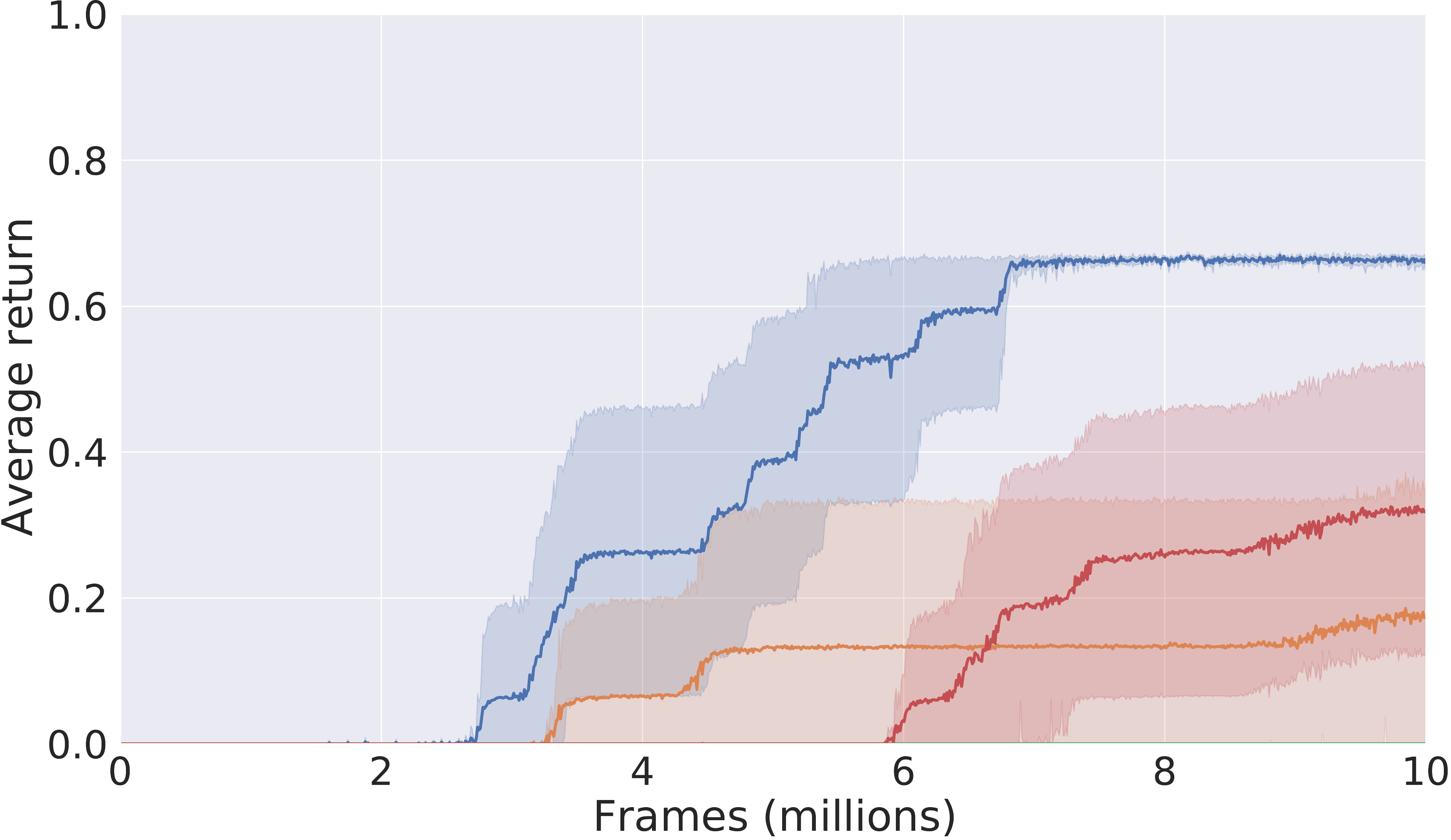}
  \caption{MultiRoom-N7S8}
  \label{fig:minigrid_results:mrn7s8}
\end{subfigure}
\begin{subfigure}{.32\textwidth}
  \centering
  \includegraphics[width=0.99\textwidth]{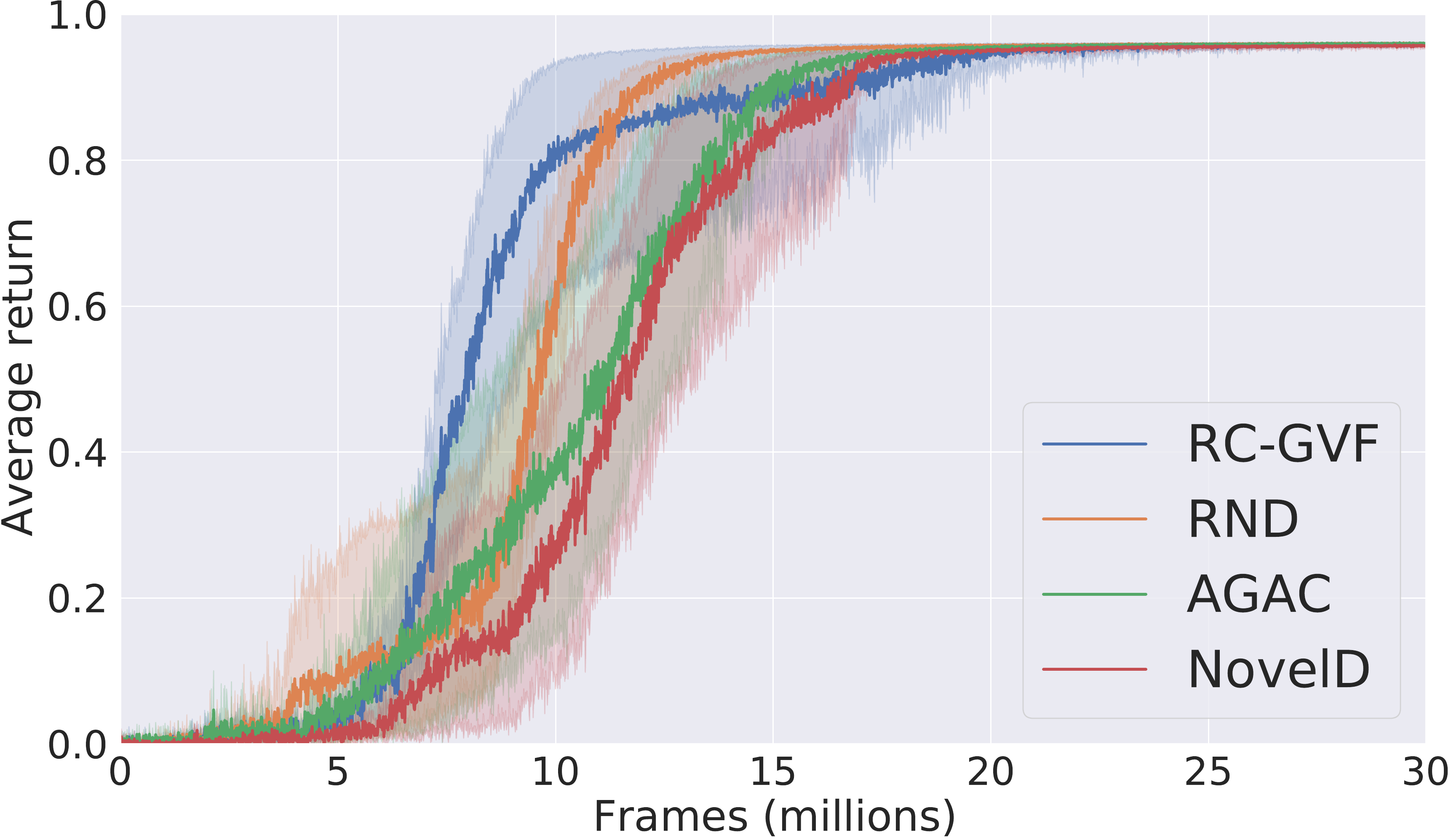}
  \caption{ObstructedMaze-2Dlh}
  \label{fig:minigrid_results:om2dlh}
\end{subfigure}
\begin{subfigure}{.32\textwidth}
  \centering
  \includegraphics[width=0.99\textwidth]{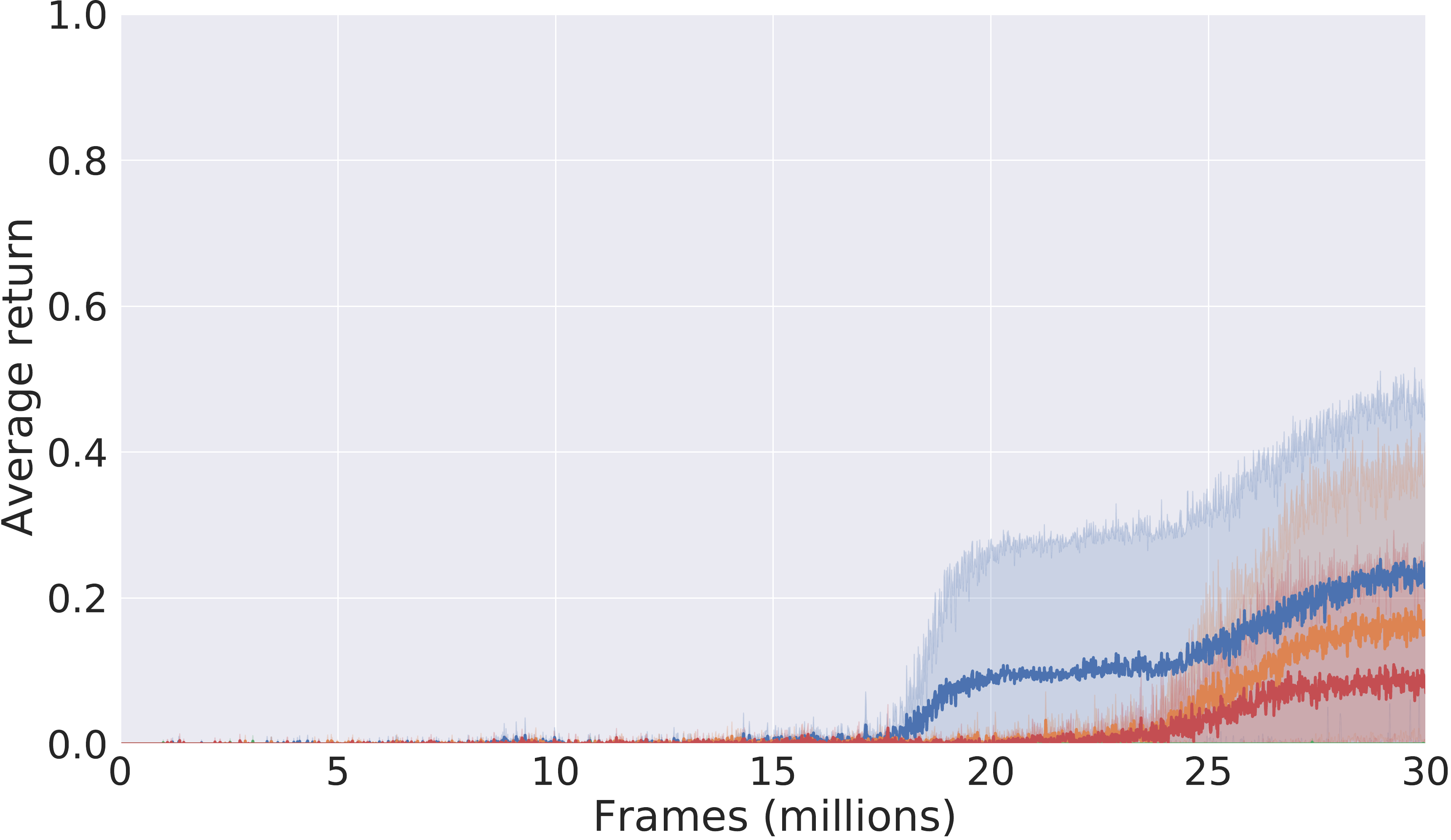}
  \caption{KeyCorridor-S5R3}
  \label{fig:minigrid_results:kcs5r3}
\end{subfigure}
\begin{subfigure}{.32\textwidth}
  \centering
\end{subfigure}
\begin{subfigure}{.32\textwidth}
  \centering
  \includegraphics[width=0.99\textwidth]{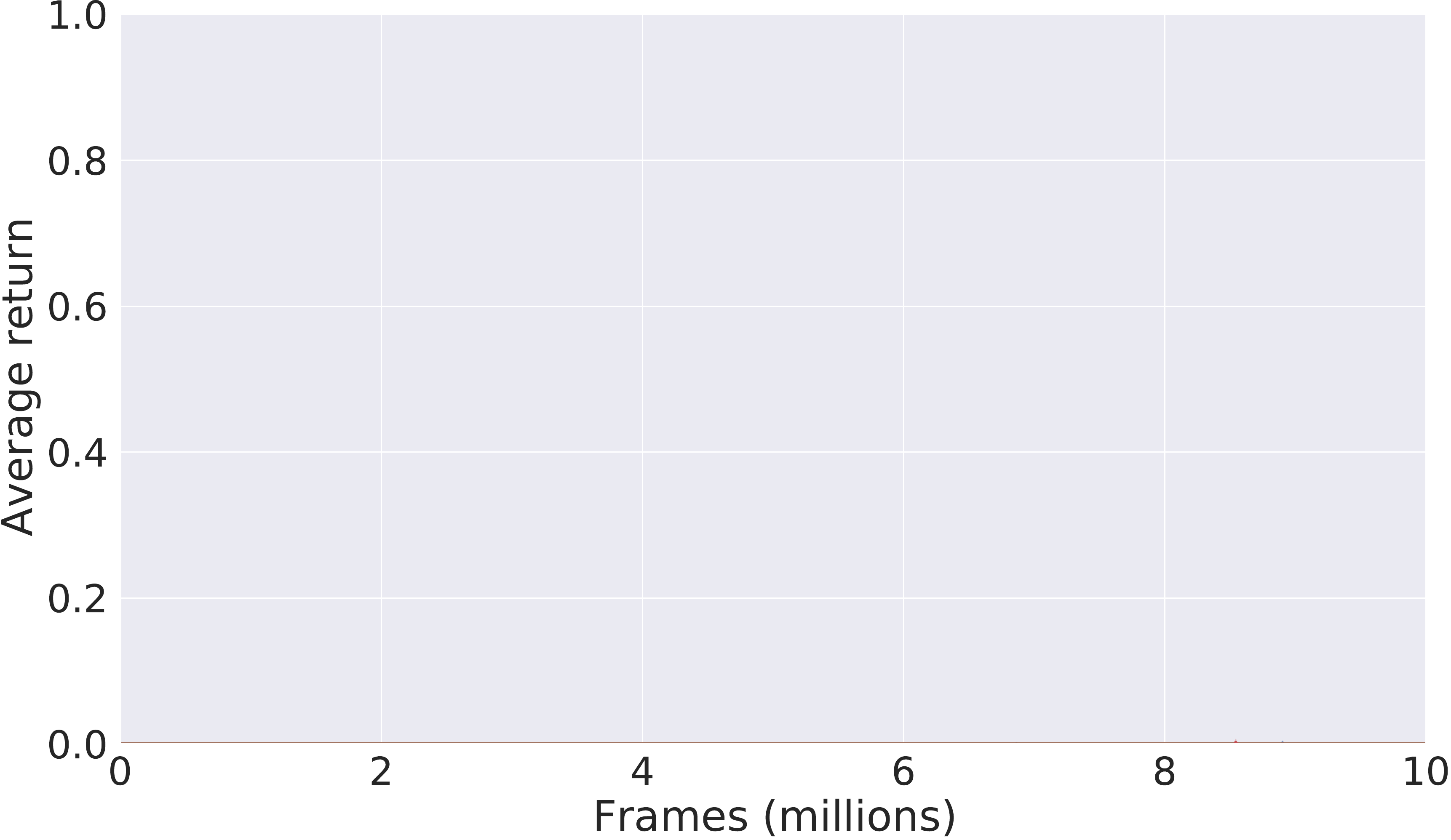}
  \caption{MultiRoom-N12S10}
  \label{fig:minigrid_results:mrn12s10}
\end{subfigure}
\begin{subfigure}{.32\textwidth}
  \centering
  \includegraphics[width=0.99\textwidth]{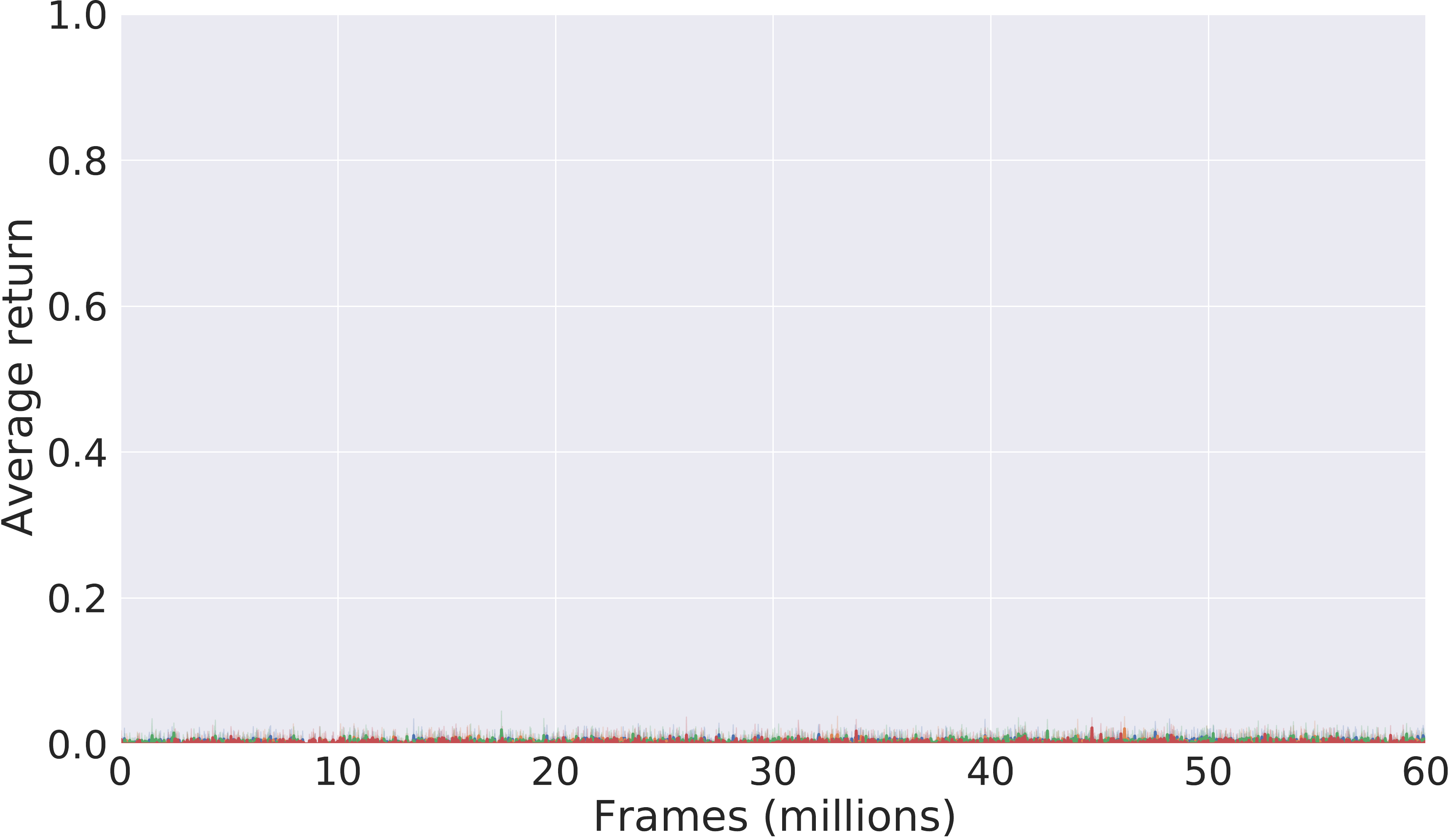}
  \caption{ObstructedMaze-2Dlhb}
  \label{fig:minigrid_results:om2dlhb}
\end{subfigure}
\caption{
Average return of \approach~, RND and other baselines on the selected Minigrid environments (without ground-truth episodic counts for baselines). RC-GVF outperforms baselines in the absence of episodic counts. Some environments prove challenging for all approaches in this setting. 
\vspace{-5mm}
}
\label{fig:minigrid_results}
\end{figure}

\subsubsection{Egocentric observations}  %

We will first consider the usual setting where the agent receives access to the egocentric observations from the environment.
Importantly, we do not provide agents with access to privileged information about the environment in the form of so-called `episodic counts'.
This is different from several recent approaches that have been applied to MiniGrid, which incorporate these values as part of their intrinsic reward~\citep{raileanu2020ride, flet2021adversarially, zhang2021noveld}.

The use of episodic counts obtained through the simulator is problematic for solving exploration problems in a partially observable setting, since it tells the agent precisely how frequently it has been in each state. 
Indeed, the use of episodic counts \emph{alone} can be sufficient for exploration in MiniGrid (Figure~\ref{fig:minigrid_episodic_counts} in Appendix~\ref{sec:appendix_only_counts}).
Meanwhile, there exists no good method for obtaining episodic counts in the absence of a simulator, as it is difficult to accurately estimate pseudo-counts from partial observations (eg. as in \autoref{fig:altcorr}). 
Details about the exact usage of episodic counts by baselines in their intrinsic reward formulation are provided in Appendix~\ref{sec:appendix_implementation:baselines}.

In \autoref{fig:minigrid_results} we compare RC-GVF, RND, AGAC, and NovelD in the absence of episodic counts, while using egocentric observations.
It can be seen that RC-GVF explores faster and succeeds more often than RND and the other baselines in the KeyCorridor-S4R3 environment and MultiRoom-N7-S8 environment.
In comparison to RND and NovelD that utilise prediction errors in immediate random embeddings, it appears that RC-GVF benefits from extended horizon predictions of random pseudo-rewards.
Meanwhile, AGAC only succeeds in ObstructedMaze-2Dlh, where all baselines perform similarly well.
Finally, we observe that all approaches struggle with the most difficult level of considered environments (KeyCorridor-S5R3, MultiRoom-N12-S10 and ObstructedMaze-2Dlhb).

\subsubsection{Panoramic observations}

\begin{figure}[t]
\begin{subfigure}{.32\textwidth}
  \centering
  \includegraphics[width=1.0\textwidth]{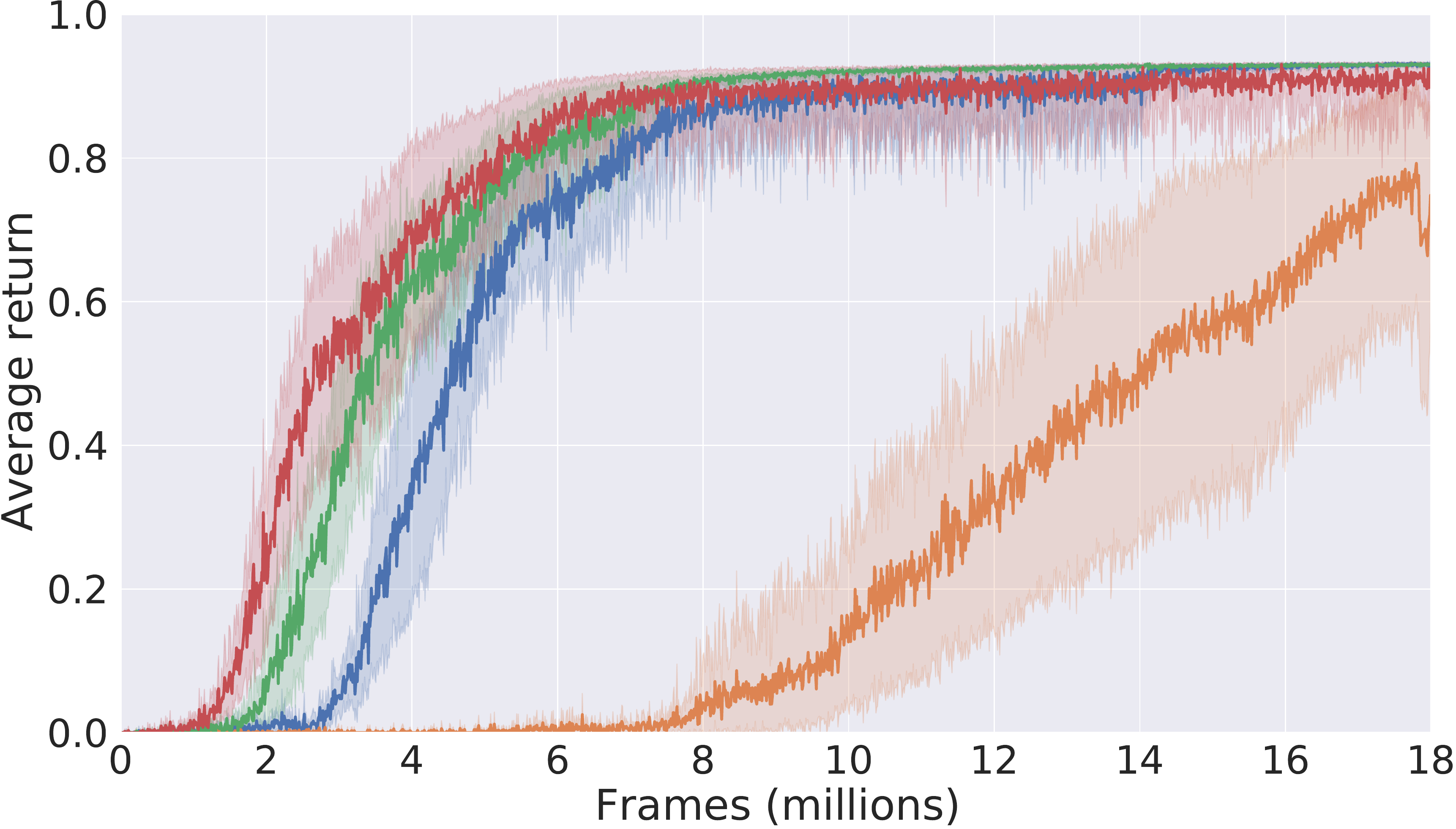}
  \caption{KeyCorridor-S4R3}
  \label{fig:minigrid_results_full:kcs4r3}
\end{subfigure}
\begin{subfigure}{.32\textwidth}
  \centering
  \includegraphics[width=0.99\textwidth]{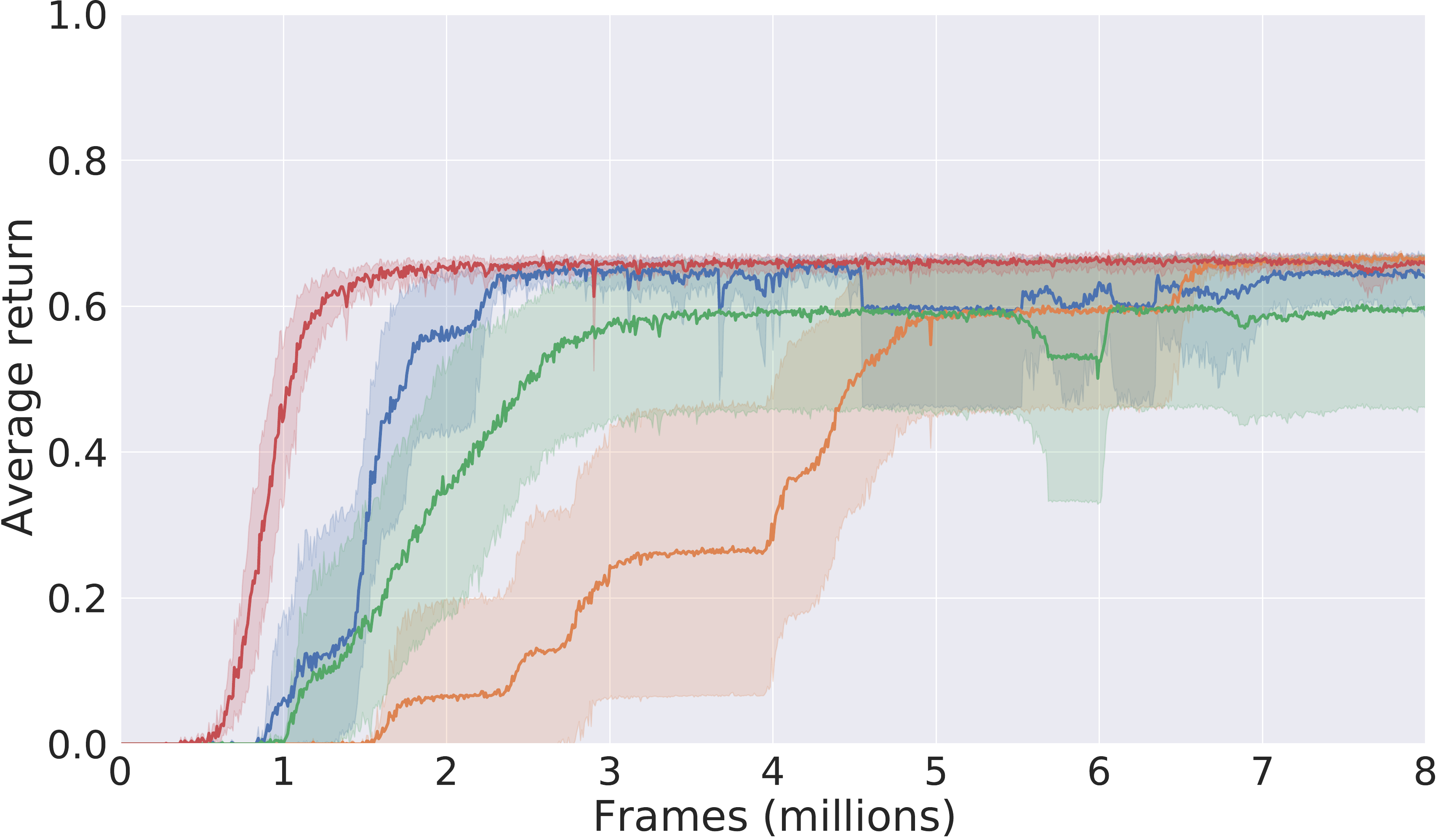}
  \caption{MultiRoom-N7-S8}
  \label{fig:minigrid_results_full:n7s8}
\end{subfigure}
\begin{subfigure}{.32\textwidth}
  \centering
  \includegraphics[width=0.99\textwidth]{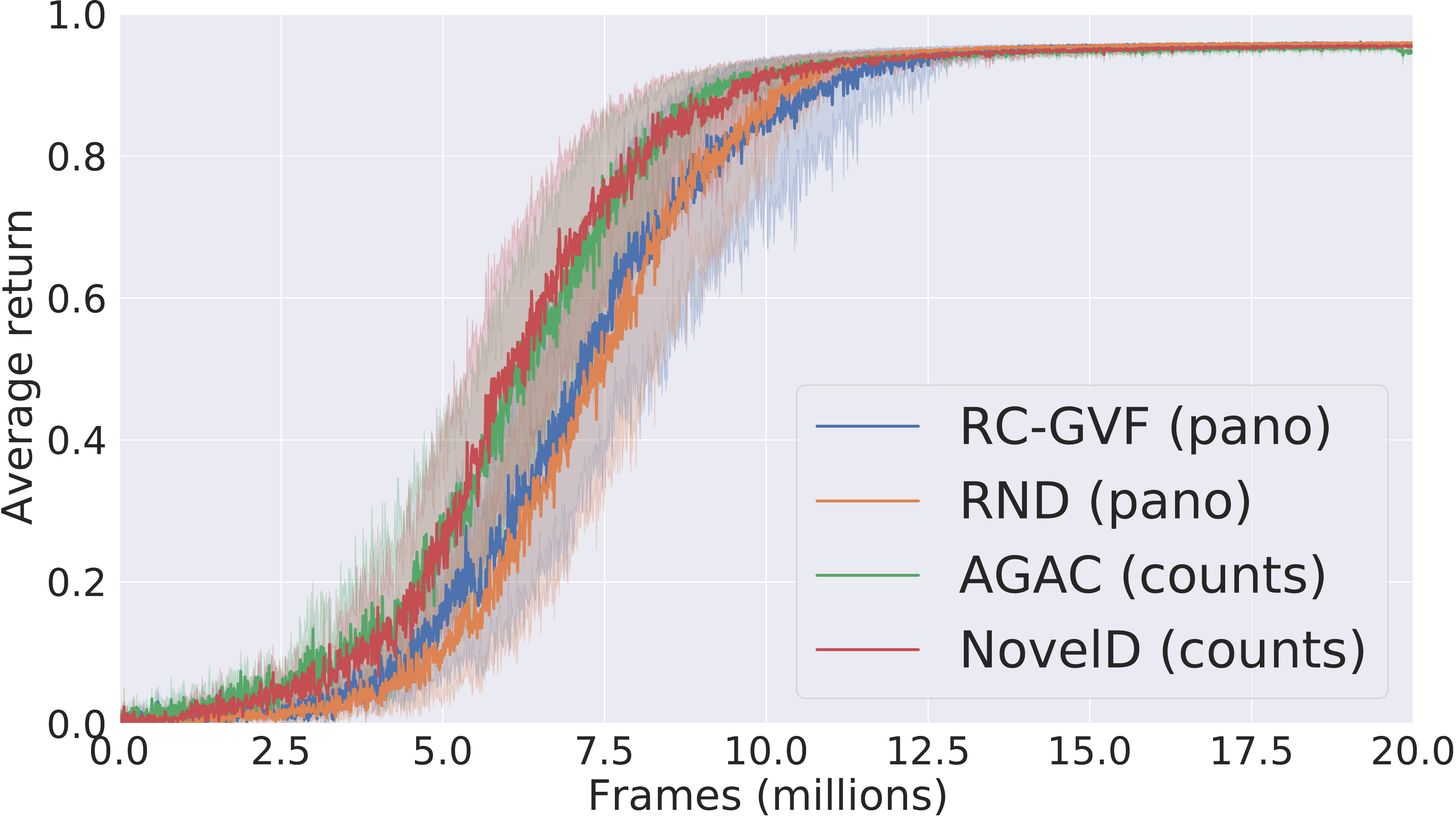}
  \caption{ObstructedMaze-2Dlh}
  \label{fig:minigrid_results_full:om2dlh}
\end{subfigure}
\begin{subfigure}{.32\textwidth}
  \centering
  \includegraphics[width=1.0\textwidth]{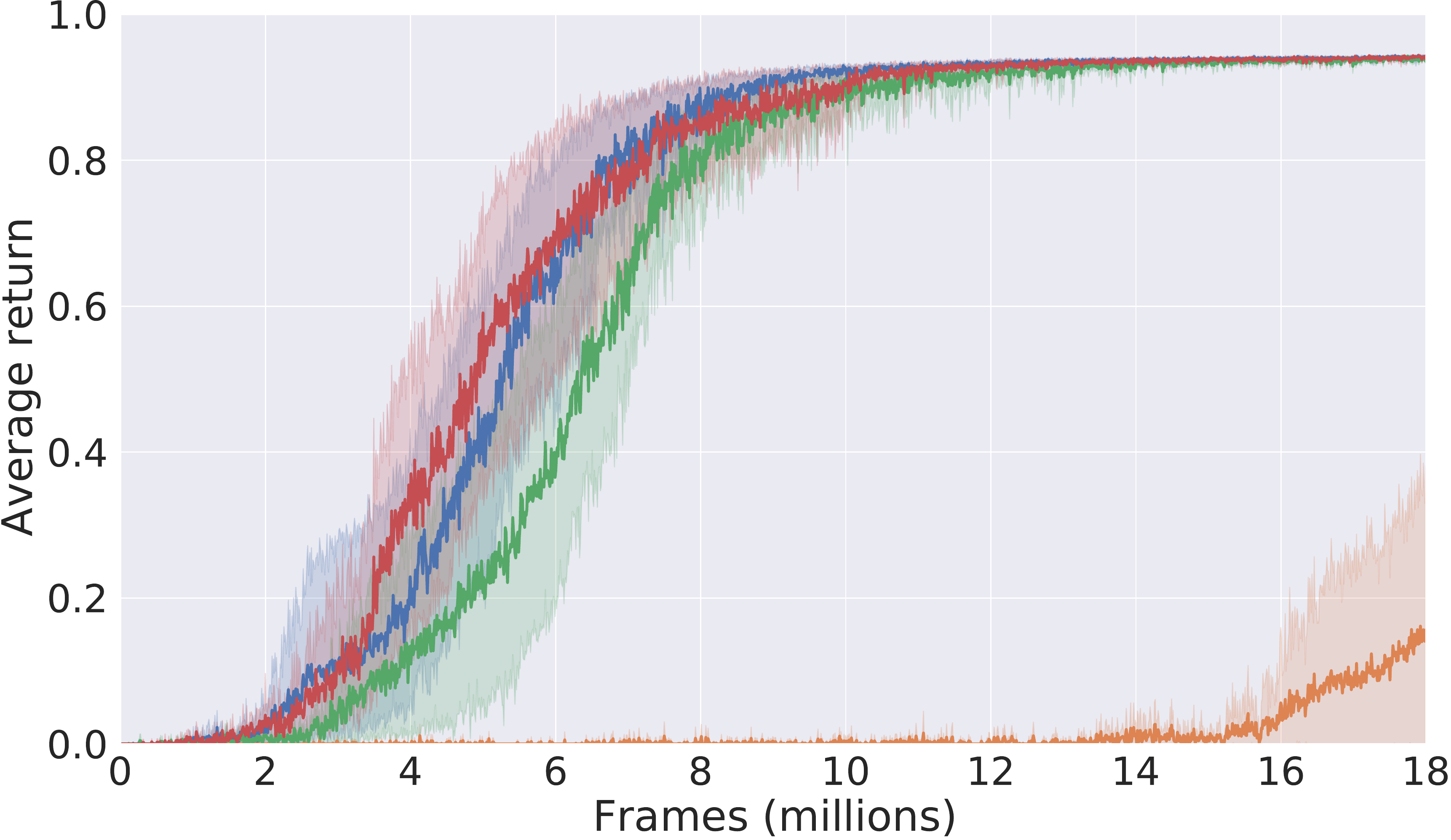}
  \caption{KeyCorridor-S5R3}
  \label{fig:minigrid_results_full:kcs5r3}
\end{subfigure}
\begin{subfigure}{.32\textwidth}
  \centering
  \includegraphics[width=0.99\textwidth]{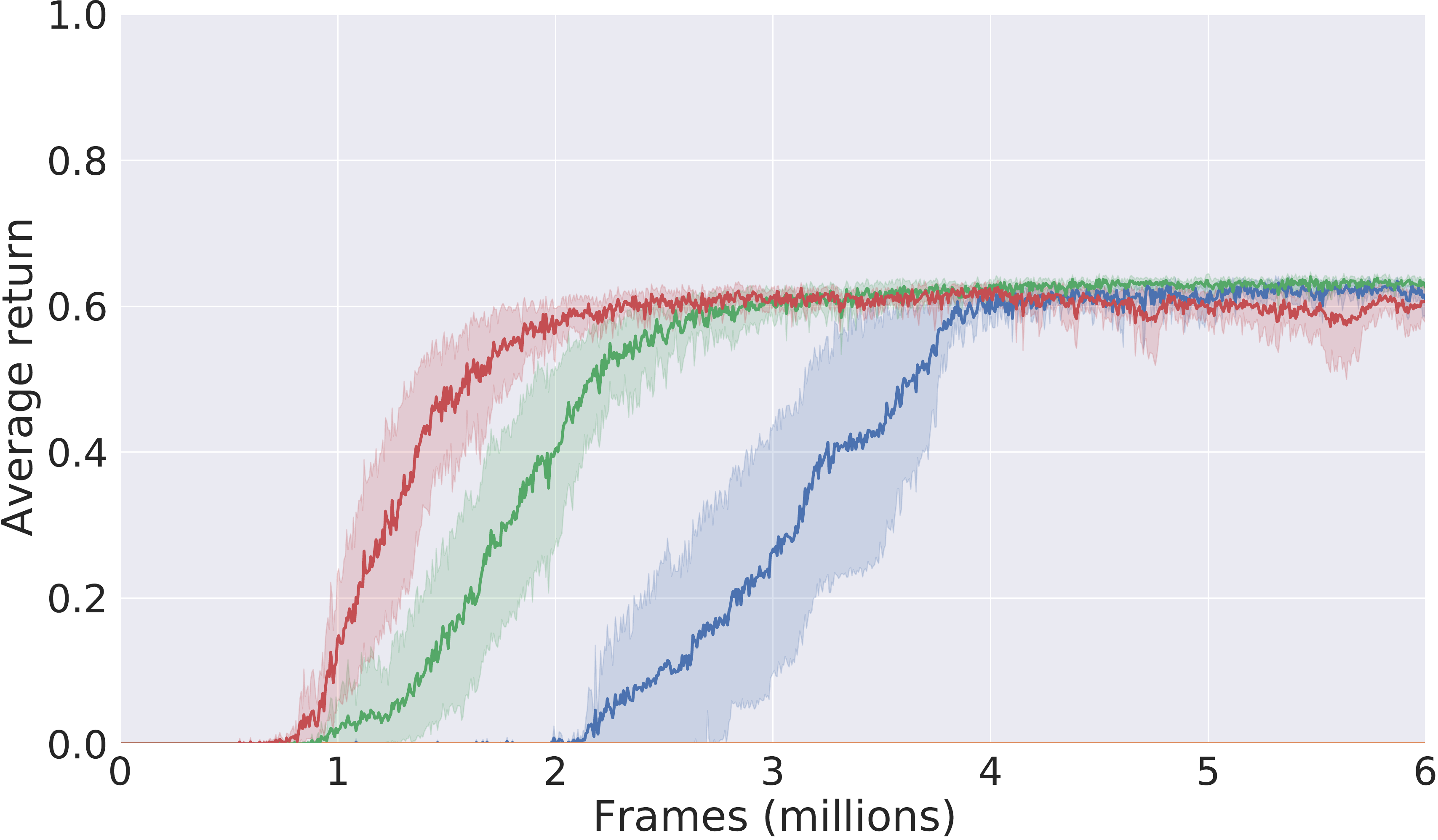}
  \caption{MultiRoom-N12-S10}
  \label{fig:minigrid_results_full:n12s10}
\end{subfigure}
\begin{subfigure}{.32\textwidth}
  \centering
  \includegraphics[width=0.99\textwidth]{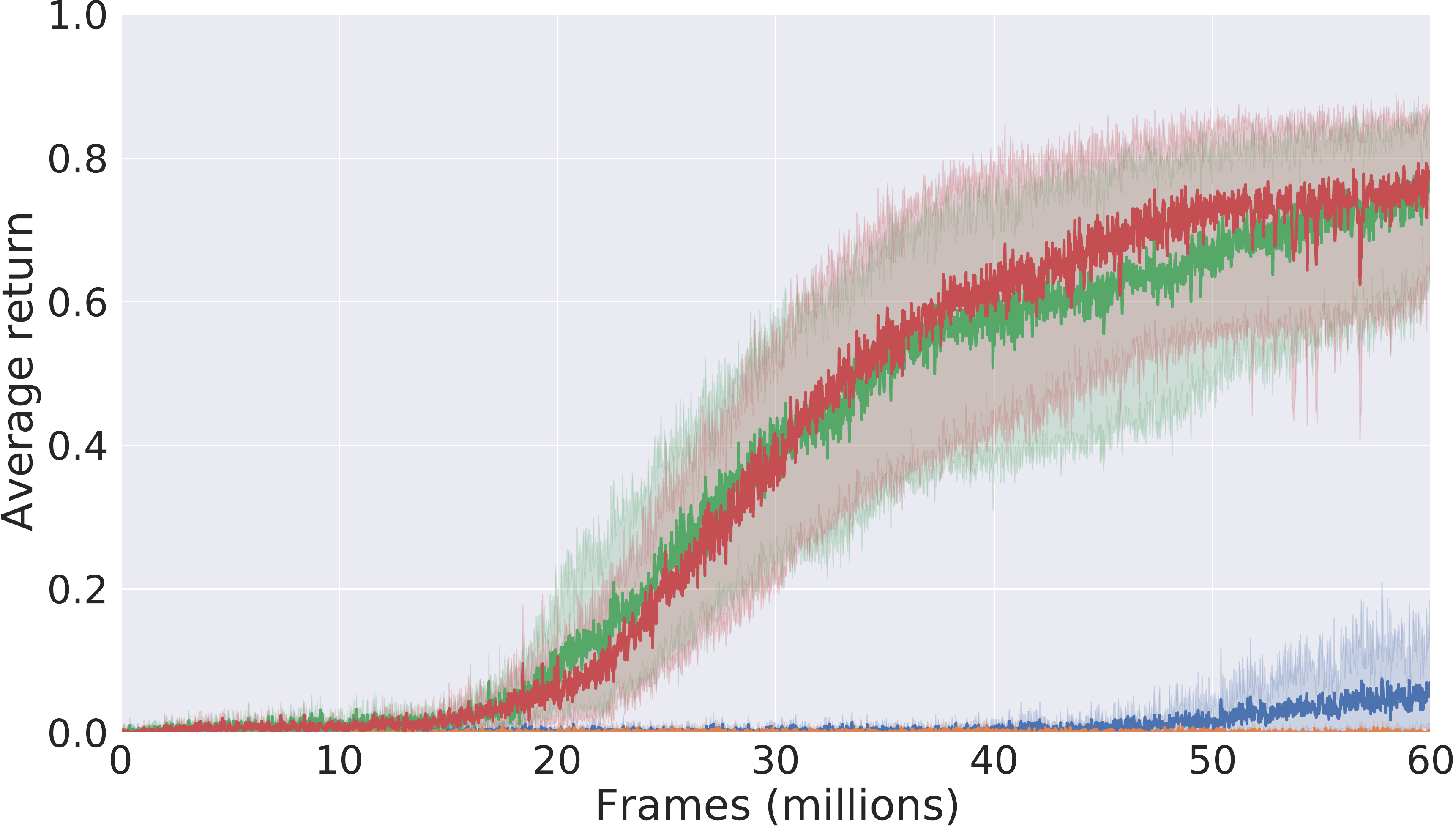}
  \caption{ObstructedMaze-2Dlhb}
  \label{fig:minigrid_results_full:om2dlhb}
\end{subfigure}
\caption{
Average return of {\approach} and RND with panoramic observations. The performance of baselines (AGAC and NovelD) improves significantly with the inclusion of the episodic count component while {\approach} (using panoramic observations) achieves similar performance without.
\vspace{-5mm}
}
\label{fig:minigrid_results_full}
\end{figure}

We now demonstrate how using episodic counts from the simulator is crucial to the success of the baselines.
In \autoref{fig:minigrid_results_full} it can be seen how AGAC and NovelD improve with episodic counts, and also solve the more difficult problems (KeyCorridor-S5R3, MultiRoom-N12-S10 and ObstructedMaze-2Dlhb).\looseness=-1

\paragraph{Panoramic observations} 
In order to increase performance of {\approach} further, without introducing episodic counts, we investigate the use of augmented panoramic views as previously explored by \citet{parisi2021interesting}\footnote{We verified in a preliminary experiment that incorporating episodic counts for RC-GVF similarly improves performance (see \autoref{fig:minigrid_rcgvf_episodic_counts} in Appendix \ref{sec:appendix_rcgvf_epcounts}). However, since we believe that this constitutes privileged information that is not readily accessible outside of a simulator, we will not explore this direction further.}.
In the MiniGrid environments, the observation changes almost in its entirety when the agent changes the direction it faces.
In contrast, moving to the next cell leads to fewer sudden changes in the observation.

The lack of gradual changes of egocentric observations through rotations is a consequence of having four discrete angles for orientation $(0^{\circ}, 90^{\circ}, 180^{\circ}, 270^{\circ})$, which may be an unrealistic depiction of how rotations affect observations for agents situated in the real world.
As a consequence, prediction errors are dominated by predictions of the outcomes of turning (rather than other actions). 
To address this, prior work proposed to make the agent's observations invariant to rotation by augmenting the observation with all directional observations~\citep{parisi2021interesting}.
In effect, this assumes that the agent rotates $360^{\circ}$ after moving to a cell, or alternatively that it is equipped with additional sensors on its sides and back.
Similar to \citet{parisi2021interesting}, we will consider panoramic views only for generating intrinsic rewards, and egocentric observations for the base PPO agent.\looseness=-1

In \autoref{fig:minigrid_results_full}, we demonstrate how RC-GVF with panoramic views and without privileged information about the underlying state of the MDP can solve harder problems and is competitive with the baselines that use episodic counts in five out of the six settings.
We see that RC-GVF with panoramic observations comfortably succeeds in the KeyCorridor-S5R3 and MultiRoom-N12-S10 environments where it was previously unsuccessful. RC-GVF did not succeed in solving the harder ObstructedMaze-2Dlhb within the given frames despite the inclusion of panoramic views.
\autoref{fig:minigrid_results_full} shows that the use of panoramic views does not make the exploration problem trivial; RND also uses panoramic views (but no episodic counts). It improves with panoramic views on KeyCorridor-S4R3 and MultiRoom-N7S8 but struggles with the harder instances.

\subsection{Analysis}

\paragraph{Changing the temporal prediction horizon}

\begin{figure}[t]
\begin{subfigure}{.4\textwidth}
  \centering
  \includegraphics[width=1.0\textwidth]{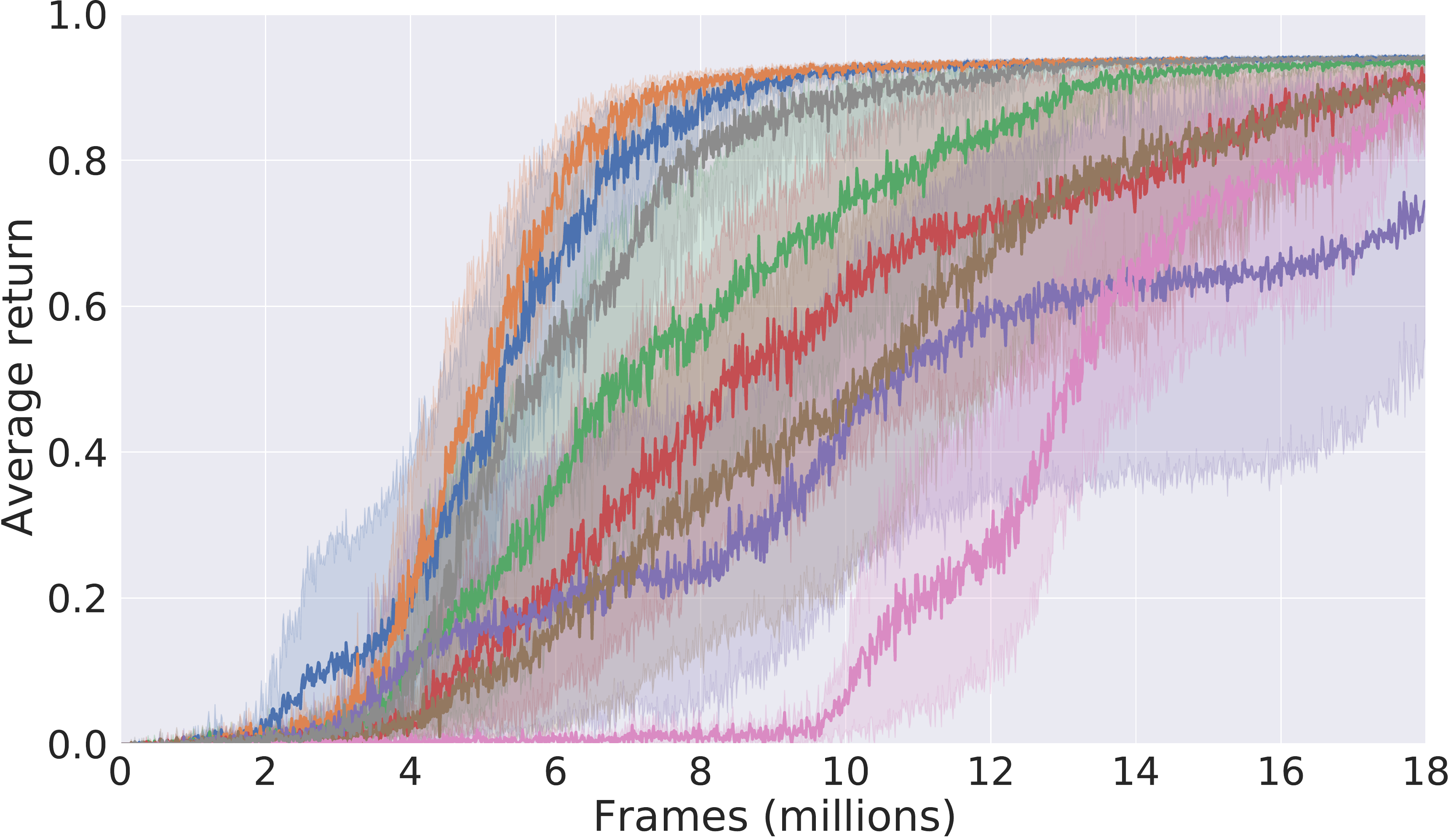}
  \caption{KeyCorridor-S5R3}
  \label{fig:minigrid_discount_analysis:kcs5r3}
\end{subfigure} \quad\quad
\begin{subfigure}{.4\textwidth}
  \centering
  \includegraphics[width=1.0\textwidth]{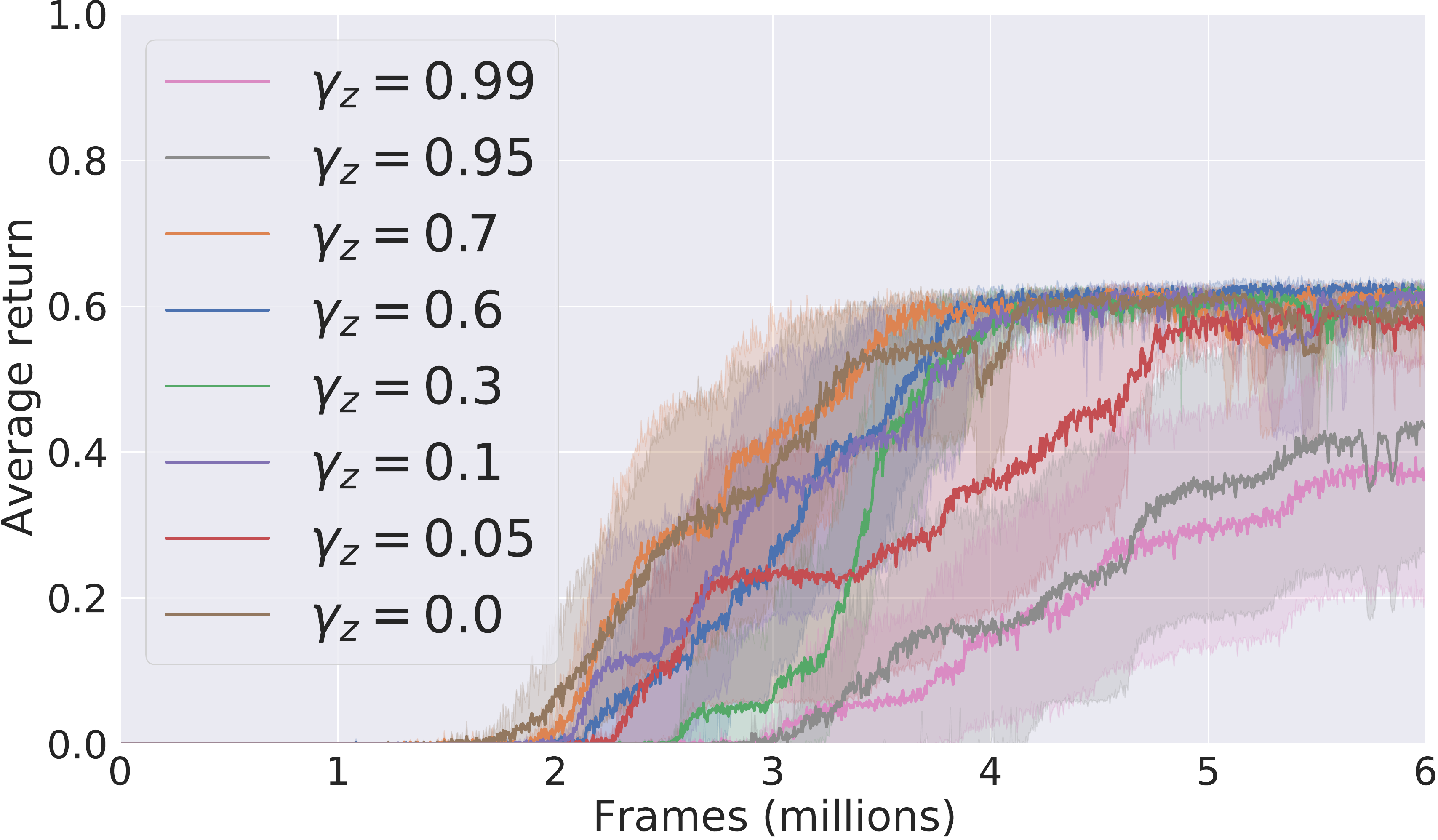}
  \caption{MultiRoom-N12-S10}
  \label{fig:minigrid_discount_analysis:mrn12s10}
\end{subfigure}
\caption{
Analysis reveals that: (a) larger discount factors (except $0.99$) are useful in KeyCorridor-S5R3 (b) low and intermediate discount factors perform well in MultiRoom-N12-S10.
\vspace{-5mm}
}
\label{fig:minigrid_discount_ablation}
\end{figure}

We study the effect of changing the temporal prediction horizon by evaluating RC-GVF with different discount factors.
Recall from \autoref{eq:discount_horizon} how the discount factor can me mapped to the prediction horizon, which lets us estimate a lower bound on the horizon of about $11$ time steps for $\gamma_z=0.6$ with $\epsilon=0.01$.  
In \autoref{fig:minigrid_discount_ablation} it can be seen how a larger prediction horizon is desirable for KeyCorridorS5R3.
Among the higher discount factors, $0.95$ is a good choice for this environment, but not $0.99$.
Compared to our default value of $\gamma_z=0.6$, a short horizon using $0.3$ reduces performance further, while discount factors closest to $0$ yield worse performance.
In contrast, for MultiRoom-N12-S10, we find that low and intermediate values of $\gamma_z$ perform well.
The high GVF discount factors of $0.99$ and $0.95$ perform worse.
Comparing $\gamma_z=0$ to RND shows how the use of recurrent networks and the variance term derived from the ensemble of predictors contributes most to the performance of RC-GVF in this domain.\looseness=-1

\begin{figure}[htpb]
\begin{subfigure}{.4\textwidth}
  \centering
  \includegraphics[width=1.0\textwidth]{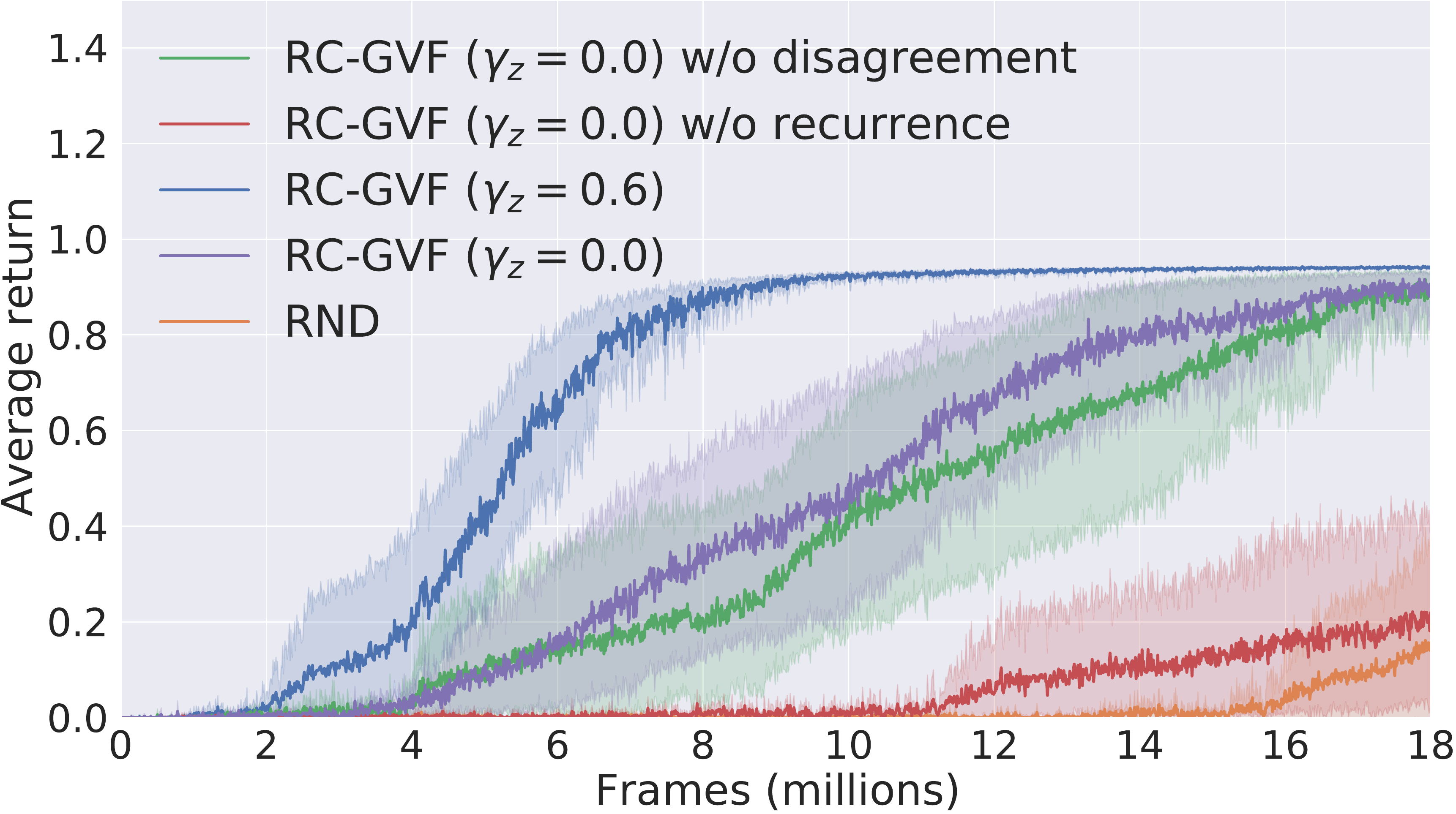}
  \caption{Ablation with $\gamma_z = 0.0$}
  \label{fig:minigrid:kcs5r3_ablation}
\end{subfigure}\quad\quad
\begin{subfigure}{.4\textwidth}
  \centering
  \includegraphics[width=1.0\textwidth]{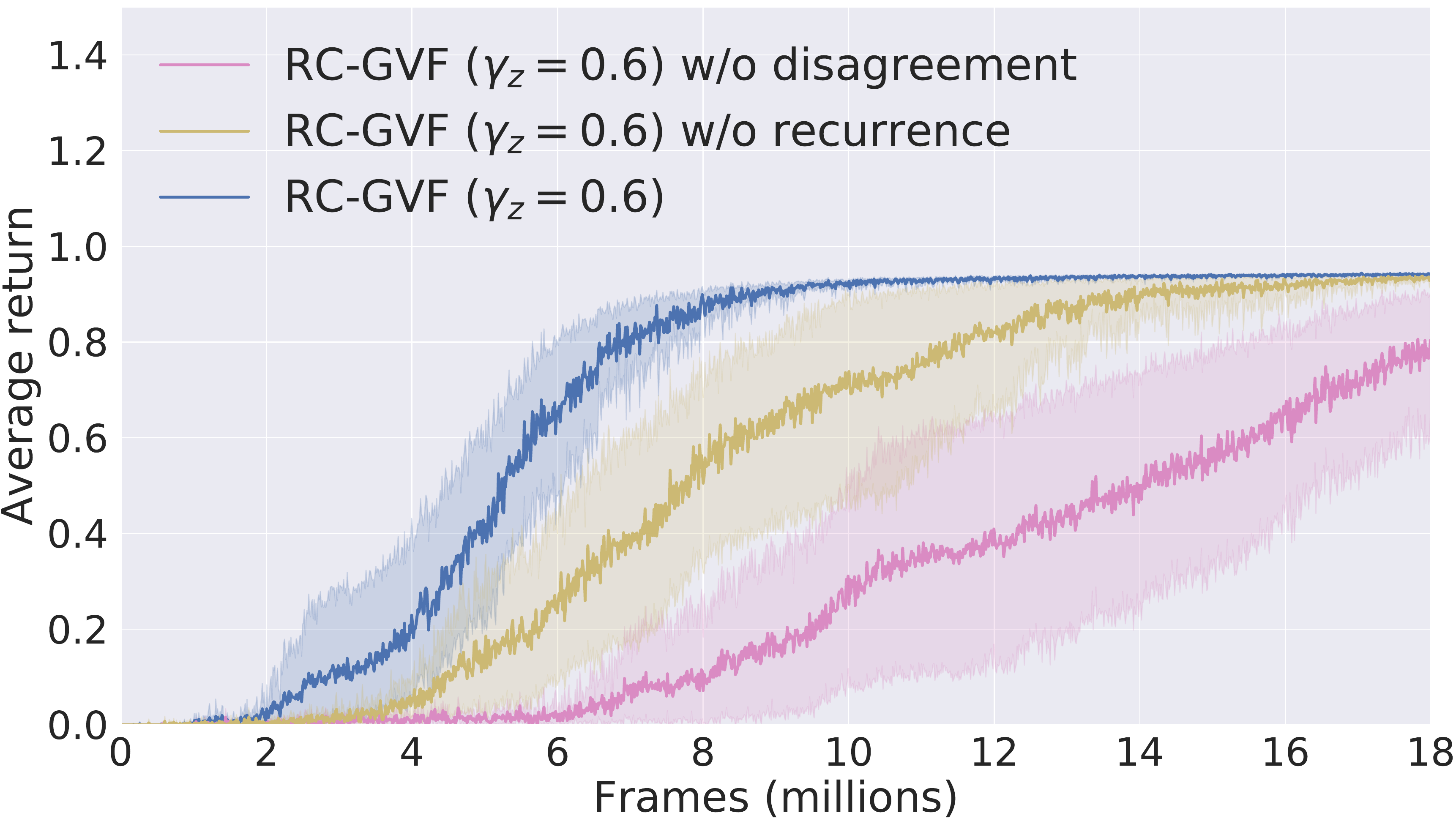}
  \caption{Ablation with $\gamma_z = 0.6$}
  \label{fig:minigrid_component_ablation_nonzerogamma}
\end{subfigure}
\caption{
Ablation of RC-GVF components in the KeyCorridor-S5R3 environment (a) the variance term, recurrent predictor and discount factor contribute most to the performance of RC-GVF (b) the variance term appears to be more important in the case of $\gamma_z=0.6$.
\vspace{-5mm}
}
\label{fig:minigrid_component_ablation}
\end{figure}

\paragraph{Ablation}
To better understand which of these components are contributing most to the success of RC-GVF, we consider the following additional ablations: (1) RC-GVF ($\gamma_z=0$) without the variance term in \autoref{eq:intrinsic_reward}, and (2) RC-GVF ($\gamma_z=0$) without the recurrent predictor.
\autoref{fig:minigrid:kcs5r3_ablation} compares these variations to RC-GVF ($\gamma_z=0$), RC-GVF ($\gamma_z=0.6$), and RND.
It can be seen that the inclusion of the history-conditioned recurrent predictor and the variance term individually contribute to the improved performance of RC-GVF.

Using the history conditioned recurrent predictor sharply improves the performance of RC-GVF with $\gamma_z=0$. 
We hypothesise that this effect arises from the previous pseudo-rewards available to the predictor, which might enable better predictions on input observations it was not explicitly trained on (which is often the case in procedurally generated environments).
In support of this hypothesis, we observed that using a recurrent predictor with solely observations as inputs does not produce such an improvement (\autoref{fig:minigrid:kcs5r3_rnn_ablation} in Appendix \ref{sec:appendix_history_conditioning}). %

In Figure \ref{fig:minigrid_component_ablation_nonzerogamma}, we present ablations for $\gamma_z=0.6$.
We consider RC-GVF without the variance term in \autoref{eq:intrinsic_reward}, and without the recurrent predictor.
Removing either the disagreement term or the recurrent predictor worsens the performance of RC-GVF with $\gamma_z = 0.6$.
Interestingly, unlike as was observed for $\gamma_z = 0$, here we find that RC-GVF with only disagreement (no recurrent predictor) performs better than RC-GVF with only recurrent predictor (no disagreement). 
This further emphasises the importance of handling the aleatoric uncertainty for non-zero GVF discounts.

\paragraph{Increasing the ensemble size} 
We conduct experiments with larger ensemble sizes in two MiniGrid environments.
We obtain comparable results for ensemble sizes of 2, 4, 6 and 8 predictors in  \autoref{fig:minigrid_rcgvf_ensemble_size} in Appendix \ref{sec:appendix_ensemble size analysis}, indicating that two member ensembles usually suffice for these problems.

\section{Limitations and Societal Impact}
\label{discussion}

\paragraph{Limitations}
The incorporation of the current policy into the prediction task of GVF prediction can have benefits with regards to behavioural exploration.
However, it may also encourage over-exploration due to the same state appearing `interesting' under different policies.
Possible mitigation strategies include off-policy variants or explicitly policy-conditioned (general) value functions~\citep{faccio2020parameter,harb2020policy}.\looseness=-1

In some situations it may be difficult to ascertain what is a good choice of the GVF discount factor $\gamma_z$, which effectively determines the prediction horizon length, and can be influenced by optimisation effects under function approximation~\citep{van2019using}.
The best choice may be problem-dependent and could further benefit from dynamic adjustment during learning~\citep{badia2020agent57, moskovitz2021tactical}.
Nonetheless, our results in Figures \ref{fig:minigrid_results} and \ref{fig:minigrid_results_full} indicate that a default of $\gamma_z=0.6$ works well in many of the considered settings.

Several other intrinsic reward formulations using RC-GVF are possible (eg. based on information/prediction gain, learning progress, or only the disagreement term), which we did not investigate.
Our specific formulation in \autoref{eq:intrinsic_reward} was motivated by the prediction error in RND, which does not separate the aleatoric uncertainty when $\gamma_z > 0$.
Starting with the prediction error term also allowed us to closely study the connection between RND and RC-GVF.

A general limitation of curiosity-based exploration for task-specific RL is that more information is gathered than is necessary to solve the given task~\citep{osband2018randomized, lu2021reinforcement}.
Indeed, the space of potential GVFs is enormous. In the case of RC-GVF one could aim at focusing it by integrating task-specific information~\citep{sutton2022reward} or by considering predictions using a smaller set of policies that adequately cover the state space~\citep{agarwal2020pc}.\looseness=-1

\paragraph{Societal impact}
This paper does not focus on real-world applications of exploration in RL and thus does not have a direct societal impact.
In applications, advanced exploration methods may lead to unexpected policy behaviour, which could be mitigated by incorporating safety constraints~\citep{garcia2015comprehensive}.
\section{Conclusion}
\label{conclusion}
We introduced \textit{{\fullapproach} ({\approach})}, an exploration method that intrinsically rewards a reinforcement learning agent based on errors and uncertainty in predicting random transformations of observation sequences generated through the agent's actions.
Unlike state-novelty approaches, ours takes into account multi-step future behaviours of policy and environment. 
Unlike commonly used artificial curiosity approaches, {\approach} does not rely on model rollouts and does not need to predict all details of future observations.

On the diabolical lock problem~\citep{misra2020kinematic} and the MiniGrid environments~\citep{gym_minigrid} we demonstrate that predicting abstract quantities of extended time intervals can improve exploration in POMDPs.
Compared to recent methods such as AGAC~\citep{flet2021adversarially} and NovelD~\citep{zhang2021noveld} that rely on privileged episodic state-visitation counts in MiniGrid, our RC-GVF achieves 
competitive results even when using only panoramic observations.
Importantly, in the natural setting where episodic counts are not available, RC-GVF significantly outperforms all baseline methods.

Our approach can be generalized  by moving beyond random pseudo-rewards, considering general value functions under a set of different policies \citep{agarwal2020pc}, and introducing time or state-dependent discounting. This offers exciting avenues for future research.\looseness=-1

\begin{ack}
We would like to thank Kenny Young, Francesco Faccio, Anand Gopalakrishnan, and Dylan Ashley  for valuable comments. This research was supported by the ERC Advanced Grant (742870), the Swiss National Science Foundation grant (200021\_192356), and by the Swiss National Supercomputing Centre (CSCS projects s1090 and s1127).
\end{ack}

\bibliographystyle{abbrvnat}
\bibliography{neurips_2022}

\begin{thebibliography}{78}
\providecommand{\natexlab}[1]{#1}
\providecommand{\url}[1]{\texttt{#1}}
\expandafter\ifx\csname urlstyle\endcsname\relax
  \providecommand{\doi}[1]{doi: #1}\else
  \providecommand{\doi}{doi: \begingroup \urlstyle{rm}\Url}\fi

\bibitem[Agarwal et~al.(2020)Agarwal, Henaff, Kakade, and Sun]{agarwal2020pc}
A.~Agarwal, M.~Henaff, S.~Kakade, and W.~Sun.
\newblock Pc-pg: Policy cover directed exploration for provable policy gradient
  learning.
\newblock \emph{Advances in Neural Information Processing Systems},
  33:\penalty0 13399--13412, 2020.

\bibitem[Amin et~al.(2021)Amin, Gomrokchi, Satija, van Hoof, and
  Precup]{amin2021survey}
S.~Amin, M.~Gomrokchi, H.~Satija, H.~van Hoof, and D.~Precup.
\newblock A survey of exploration methods in reinforcement learning.
\newblock \emph{arXiv preprint arXiv:2109.00157}, 2021.

\bibitem[{\AA}str{\"o}m(1965)]{aastrom1965optimal}
K.~J. {\AA}str{\"o}m.
\newblock Optimal control of markov processes with incomplete state
  information.
\newblock \emph{Journal of mathematical analysis and applications}, 10\penalty0
  (1):\penalty0 174--205, 1965.

\bibitem[Auer et~al.(2008)Auer, Jaksch, and Ortner]{auer2008near}
P.~Auer, T.~Jaksch, and R.~Ortner.
\newblock Near-optimal regret bounds for reinforcement learning.
\newblock \emph{Advances in neural information processing systems}, 21, 2008.

\bibitem[Badia et~al.(2020)Badia, Piot, Kapturowski, Sprechmann, Vitvitskyi,
  Guo, and Blundell]{badia2020agent57}
A.~P. Badia, B.~Piot, S.~Kapturowski, P.~Sprechmann, A.~Vitvitskyi, Z.~D. Guo,
  and C.~Blundell.
\newblock Agent57: Outperforming the atari human benchmark.
\newblock In \emph{International Conference on Machine Learning}, pages
  507--517. PMLR, 2020.

\bibitem[Barto and Singh(1991)]{barto1991computational}
A.~G. Barto and S.~P. Singh.
\newblock On the computational economics of reinforcement learning.
\newblock In \emph{Connectionist Models}, pages 35--44. Elsevier, 1991.

\bibitem[Bellemare et~al.(2016)Bellemare, Srinivasan, Ostrovski, Schaul,
  Saxton, and Munos]{bellemare2016unifying}
M.~Bellemare, S.~Srinivasan, G.~Ostrovski, T.~Schaul, D.~Saxton, and R.~Munos.
\newblock Unifying count-based exploration and intrinsic motivation.
\newblock \emph{Advances in neural information processing systems},
  29:\penalty0 1471--1479, 2016.

\bibitem[Bellemare et~al.(2019)Bellemare, Dabney, Dadashi, Ali~Taiga, Castro,
  Le~Roux, Schuurmans, Lattimore, and Lyle]{bellemare2019geometric}
M.~Bellemare, W.~Dabney, R.~Dadashi, A.~Ali~Taiga, P.~S. Castro, N.~Le~Roux,
  D.~Schuurmans, T.~Lattimore, and C.~Lyle.
\newblock A geometric perspective on optimal representations for reinforcement
  learning.
\newblock \emph{Advances in neural information processing systems}, 32, 2019.

\bibitem[Burda et~al.(2019{\natexlab{a}})Burda, Edwards, Pathak, Storkey,
  Darrell, and Efros]{burda2018large}
Y.~Burda, H.~Edwards, D.~Pathak, A.~J. Storkey, T.~Darrell, and A.~A. Efros.
\newblock Large-scale study of curiosity-driven learning.
\newblock In \emph{7th International Conference on Learning Representations},
  2019{\natexlab{a}}.

\bibitem[Burda et~al.(2019{\natexlab{b}})Burda, Edwards, Storkey, and
  Klimov]{burda2018exploration}
Y.~Burda, H.~Edwards, A.~J. Storkey, and O.~Klimov.
\newblock Exploration by random network distillation.
\newblock In \emph{7th International Conference on Learning Representations},
  2019{\natexlab{b}}.

\bibitem[Campero et~al.(2020)Campero, Raileanu, K{\"u}ttler, Tenenbaum,
  Rockt{\"a}schel, and Grefenstette]{campero2020learning}
A.~Campero, R.~Raileanu, H.~K{\"u}ttler, J.~B. Tenenbaum, T.~Rockt{\"a}schel,
  and E.~Grefenstette.
\newblock Learning with amigo: Adversarially motivated intrinsic goals.
\newblock \emph{arXiv preprint arXiv:2006.12122}, 2020.

\bibitem[Chentanez et~al.(2004)Chentanez, Barto, and
  Singh]{chentanez2004intrinsically}
N.~Chentanez, A.~Barto, and S.~Singh.
\newblock Intrinsically motivated reinforcement learning.
\newblock \emph{Advances in neural information processing systems}, 17, 2004.

\bibitem[Chevalier-Boisvert et~al.(2018)Chevalier-Boisvert, Willems, and
  Pal]{gym_minigrid}
M.~Chevalier-Boisvert, L.~Willems, and S.~Pal.
\newblock Minimalistic gridworld environment for openai gym.
\newblock \url{https://github.com/maximecb/gym-minigrid}, 2018.

\bibitem[Ciosek et~al.(2020)Ciosek, Fortuin, Tomioka, Hofmann, and
  Turner]{ciosek2020conservative}
K.~Ciosek, V.~Fortuin, R.~Tomioka, K.~Hofmann, and R.~E. Turner.
\newblock Conservative uncertainty estimation by fitting prior networks.
\newblock In \emph{8th International Conference on Learning Representations,
  {ICLR}}, 2020.

\bibitem[Der~Kiureghian and Ditlevsen(2009)]{der2009aleatory}
A.~Der~Kiureghian and O.~Ditlevsen.
\newblock Aleatory or epistemic? does it matter?
\newblock \emph{Structural safety}, 31\penalty0 (2):\penalty0 105--112, 2009.

\bibitem[Espeholt et~al.(2018)Espeholt, Soyer, Munos, Simonyan, Mnih, Ward,
  Doron, Firoiu, Harley, Dunning, et~al.]{espeholt2018impala}
L.~Espeholt, H.~Soyer, R.~Munos, K.~Simonyan, V.~Mnih, T.~Ward, Y.~Doron,
  V.~Firoiu, T.~Harley, I.~Dunning, et~al.
\newblock Impala: Scalable distributed deep-rl with importance weighted
  actor-learner architectures.
\newblock In \emph{International Conference on Machine Learning}, pages
  1407--1416. PMLR, 2018.

\bibitem[Faccio et~al.(2021)Faccio, Kirsch, and
  Schmidhuber]{faccio2020parameter}
F.~Faccio, L.~Kirsch, and J.~Schmidhuber.
\newblock Parameter-based value functions.
\newblock In \emph{9th International Conference on Learning Representations},
  2021.

\bibitem[Flet{-}Berliac et~al.(2021)Flet{-}Berliac, Ferret, Pietquin, Preux,
  and Geist]{flet2021adversarially}
Y.~Flet{-}Berliac, J.~Ferret, O.~Pietquin, P.~Preux, and M.~Geist.
\newblock Adversarially guided actor-critic.
\newblock In \emph{9th International Conference on Learning Representations},
  2021.

\bibitem[Garc{\i}a and Fern{\'a}ndez(2015)]{garcia2015comprehensive}
J.~Garc{\i}a and F.~Fern{\'a}ndez.
\newblock A comprehensive survey on safe reinforcement learning.
\newblock \emph{Journal of Machine Learning Research}, 16\penalty0
  (1):\penalty0 1437--1480, 2015.

\bibitem[Harb et~al.(2020)Harb, Schaul, Precup, and Bacon]{harb2020policy}
J.~Harb, T.~Schaul, D.~Precup, and P.-L. Bacon.
\newblock Policy evaluation networks.
\newblock \emph{arXiv preprint arXiv:2002.11833}, 2020.

\bibitem[Hochreiter and Schmidhuber(1997)]{hochreiter1997long}
S.~Hochreiter and J.~Schmidhuber.
\newblock Long short-term memory.
\newblock \emph{Neural computation}, 9\penalty0 (8):\penalty0 1735--1780, 1997.

\bibitem[Houthooft et~al.(2016)Houthooft, Chen, Duan, Schulman, De~Turck, and
  Abbeel]{houthooft2016vime}
R.~Houthooft, X.~Chen, Y.~Duan, J.~Schulman, F.~De~Turck, and P.~Abbeel.
\newblock Vime: Variational information maximizing exploration.
\newblock \emph{Advances in Neural Information Processing Systems},
  29:\penalty0 1109--1117, 2016.

\bibitem[H{\"u}llermeier and Waegeman(2021)]{hullermeier2021aleatoric}
E.~H{\"u}llermeier and W.~Waegeman.
\newblock Aleatoric and epistemic uncertainty in machine learning: An
  introduction to concepts and methods.
\newblock \emph{Machine Learning}, 110\penalty0 (3):\penalty0 457--506, 2021.

\bibitem[Jaderberg et~al.(2016)Jaderberg, Mnih, Czarnecki, Schaul, Leibo,
  Silver, and Kavukcuoglu]{jaderberg2016reinforcement}
M.~Jaderberg, V.~Mnih, W.~M. Czarnecki, T.~Schaul, J.~Z. Leibo, D.~Silver, and
  K.~Kavukcuoglu.
\newblock Reinforcement learning with unsupervised auxiliary tasks.
\newblock \emph{arXiv preprint arXiv:1611.05397}, 2016.

\bibitem[Jain et~al.(2021)Jain, Lahlou, Nekoei, Butoi, Bertin, Rector-Brooks,
  Korablyov, and Bengio]{jain2021deup}
M.~Jain, S.~Lahlou, H.~Nekoei, V.~Butoi, P.~Bertin, J.~Rector-Brooks,
  M.~Korablyov, and Y.~Bengio.
\newblock Deup: Direct epistemic uncertainty prediction.
\newblock \emph{arXiv preprint arXiv:2102.08501}, 2021.

\bibitem[Kaelbling et~al.(1998)Kaelbling, Littman, and
  Cassandra]{kaelbling1998planning}
L.~P. Kaelbling, M.~L. Littman, and A.~R. Cassandra.
\newblock Planning and acting in partially observable stochastic domains.
\newblock \emph{Artificial intelligence}, 101\penalty0 (1-2):\penalty0 99--134,
  1998.

\bibitem[Kearney et~al.(2021)Kearney, Koop, G{\"u}nther, and
  Pilarski]{kearney2021finding}
A.~Kearney, A.~Koop, J.~G{\"u}nther, and P.~M. Pilarski.
\newblock Finding useful predictions by meta-gradient descent to improve
  decision-making.
\newblock \emph{arXiv preprint arXiv:2111.11212}, 2021.

\bibitem[Kearns et~al.(2002)Kearns, Mansour, and Ng]{kearns2002sparse}
M.~Kearns, Y.~Mansour, and A.~Y. Ng.
\newblock A sparse sampling algorithm for near-optimal planning in large markov
  decision processes.
\newblock \emph{Machine learning}, 49\penalty0 (2):\penalty0 193--208, 2002.

\bibitem[Kingma and Ba(2015)]{kingma2015adam}
D.~P. Kingma and J.~Ba.
\newblock Adam: {A} method for stochastic optimization.
\newblock In \emph{3rd International Conference on Learning Representations},
  2015.

\bibitem[Kocsis and Szepesv{\'a}ri(2006)]{kocsis2006bandit}
L.~Kocsis and C.~Szepesv{\'a}ri.
\newblock Bandit based monte-carlo planning.
\newblock In \emph{European conference on machine learning}, pages 282--293.
  Springer, 2006.

\bibitem[Lambert et~al.(2022)Lambert, Wulfmeier, Whitney, Byravan, Bloesch,
  Dasagi, Hertweck, and Riedmiller]{lambert2022challenges}
N.~Lambert, M.~Wulfmeier, W.~Whitney, A.~Byravan, M.~Bloesch, V.~Dasagi,
  T.~Hertweck, and M.~Riedmiller.
\newblock The challenges of exploration for offline reinforcement learning.
\newblock \emph{arXiv preprint arXiv:2201.11861}, 2022.

\bibitem[Littman and Sutton(2001)]{littman2001predictive}
M.~Littman and R.~S. Sutton.
\newblock Predictive representations of state.
\newblock \emph{Advances in neural information processing systems}, 14, 2001.

\bibitem[Lu et~al.(2021)Lu, Van~Roy, Dwaracherla, Ibrahimi, Osband, and
  Wen]{lu2021reinforcement}
X.~Lu, B.~Van~Roy, V.~Dwaracherla, M.~Ibrahimi, I.~Osband, and Z.~Wen.
\newblock Reinforcement learning, bit by bit.
\newblock \emph{arXiv preprint arXiv:2103.04047}, 2021.

\bibitem[Lyle et~al.(2021)Lyle, Rowland, Ostrovski, and Dabney]{lyle2021effect}
C.~Lyle, M.~Rowland, G.~Ostrovski, and W.~Dabney.
\newblock On the effect of auxiliary tasks on representation dynamics.
\newblock In \emph{International Conference on Artificial Intelligence and
  Statistics}, pages 1--9. PMLR, 2021.

\bibitem[McLeod et~al.(2021)McLeod, Lo, Schlegel, Jacobsen, Kumaraswamy, White,
  and White]{mcleod2021continual}
M.~McLeod, C.~Lo, M.~Schlegel, A.~Jacobsen, R.~Kumaraswamy, M.~White, and
  A.~White.
\newblock Continual auxiliary task learning.
\newblock \emph{Advances in Neural Information Processing Systems},
  34:\penalty0 12549--12562, 2021.

\bibitem[Misra et~al.(2020)Misra, Henaff, Krishnamurthy, and
  Langford]{misra2020kinematic}
D.~Misra, M.~Henaff, A.~Krishnamurthy, and J.~Langford.
\newblock Kinematic state abstraction and provably efficient rich-observation
  reinforcement learning.
\newblock In \emph{International conference on machine learning}, pages
  6961--6971. PMLR, 2020.

\bibitem[Moskovitz et~al.(2021)Moskovitz, Parker-Holder, Pacchiano, Arbel, and
  Jordan]{moskovitz2021tactical}
T.~Moskovitz, J.~Parker-Holder, A.~Pacchiano, M.~Arbel, and M.~Jordan.
\newblock Tactical optimism and pessimism for deep reinforcement learning.
\newblock \emph{Advances in Neural Information Processing Systems}, 34, 2021.

\bibitem[Osband et~al.(2016{\natexlab{a}})Osband, Blundell, Pritzel, and
  Van~Roy]{osband2016deep}
I.~Osband, C.~Blundell, A.~Pritzel, and B.~Van~Roy.
\newblock Deep exploration via bootstrapped dqn.
\newblock \emph{Advances in neural information processing systems}, 29,
  2016{\natexlab{a}}.

\bibitem[Osband et~al.(2016{\natexlab{b}})Osband, Van~Roy, and
  Wen]{osband2016generalization}
I.~Osband, B.~Van~Roy, and Z.~Wen.
\newblock Generalization and exploration via randomized value functions.
\newblock In \emph{International Conference on Machine Learning}, pages
  2377--2386. PMLR, 2016{\natexlab{b}}.

\bibitem[Osband et~al.(2018)Osband, Aslanides, and
  Cassirer]{osband2018randomized}
I.~Osband, J.~Aslanides, and A.~Cassirer.
\newblock Randomized prior functions for deep reinforcement learning.
\newblock \emph{Advances in Neural Information Processing Systems}, 31, 2018.

\bibitem[Ostrovski et~al.(2017)Ostrovski, Bellemare, Oord, and
  Munos]{ostrovski2017count}
G.~Ostrovski, M.~G. Bellemare, A.~Oord, and R.~Munos.
\newblock Count-based exploration with neural density models.
\newblock In \emph{International conference on machine learning}, pages
  2721--2730. PMLR, 2017.

\bibitem[Parisi et~al.(2021)Parisi, Dean, Pathak, and
  Gupta]{parisi2021interesting}
S.~Parisi, V.~Dean, D.~Pathak, and A.~Gupta.
\newblock Interesting object, curious agent: Learning task-agnostic
  exploration.
\newblock \emph{Advances in Neural Information Processing Systems}, 34, 2021.

\bibitem[Pathak et~al.(2017)Pathak, Agrawal, Efros, and
  Darrell]{pathak2017curiosity}
D.~Pathak, P.~Agrawal, A.~A. Efros, and T.~Darrell.
\newblock Curiosity-driven exploration by self-supervised prediction.
\newblock In \emph{International conference on machine learning}, pages
  2778--2787. PMLR, 2017.

\bibitem[Pathak et~al.(2019)Pathak, Gandhi, and Gupta]{pathak2019self}
D.~Pathak, D.~Gandhi, and A.~Gupta.
\newblock Self-supervised exploration via disagreement.
\newblock In \emph{International conference on machine learning}, pages
  5062--5071. PMLR, 2019.

\bibitem[Rahimi and Recht(2008)]{rahimi2008weighted}
A.~Rahimi and B.~Recht.
\newblock Weighted sums of random kitchen sinks: Replacing minimization with
  randomization in learning.
\newblock \emph{Advances in neural information processing systems}, 21, 2008.

\bibitem[Raileanu and Rockt{\"{a}}schel(2020)]{raileanu2020ride}
R.~Raileanu and T.~Rockt{\"{a}}schel.
\newblock {RIDE:} rewarding impact-driven exploration for
  procedurally-generated environments.
\newblock In \emph{8th International Conference on Learning Representations},
  2020.

\bibitem[Schaul and Ring(2013)]{schaul2013better}
T.~Schaul and M.~Ring.
\newblock Better generalization with forecasts.
\newblock In \emph{Twenty-Third International Joint Conference on Artificial
  Intelligence}, 2013.

\bibitem[Schaul et~al.(2018)Schaul, van Hasselt, Modayil, White, White, Bacon,
  Harb, Mourad, Bellemare, and Precup]{schaul2018barbados}
T.~Schaul, H.~van Hasselt, J.~Modayil, M.~White, A.~White, P.-L. Bacon,
  J.~Harb, S.~Mourad, M.~Bellemare, and D.~Precup.
\newblock The barbados 2018 list of open issues in continual learning.
\newblock \emph{arXiv preprint arXiv:1811.07004}, 2018.

\bibitem[Schmidhuber(1990)]{schmidhuber1990making}
J.~Schmidhuber.
\newblock Making the world differentiable: On using fully recurrent
  self-supervised neural networks for dynamic reinforcement learning and
  planning in non-stationary environments.
\newblock Technical Report FKI-126-90, TUM, 1990.

\bibitem[Schmidhuber(1991{\natexlab{a}})]{Schmidhuber:91nips}
J.~Schmidhuber.
\newblock Reinforcement learning in {M}arkovian and non-{M}arkovian
  environments.
\newblock In D.~S. Lippman, J.~E. Moody, and D.~S. Touretzky, editors,
  \emph{Advances in Neural Information Processing Systems 3 (NIPS 3)}, pages
  500--506. Morgan Kaufmann, 1991{\natexlab{a}}.

\bibitem[Schmidhuber(1991{\natexlab{b}})]{Schmidhuber:91singaporecur}
J.~Schmidhuber.
\newblock Curious model-building control systems.
\newblock In \emph{Proceedings of the International Joint Conference on Neural
  Networks, Singapore}, volume~2, pages 1458--1463. IEEE press,
  1991{\natexlab{b}}.

\bibitem[Schmidhuber(1991{\natexlab{c}})]{schmidhuber1991curiosity}
J.~Schmidhuber.
\newblock A possibility for implementing curiosity and boredom in
  model-building neural controllers.
\newblock In \emph{Proc. of the international conference on simulation of
  adaptive behavior: From animals to animats}, pages 222--227,
  1991{\natexlab{c}}.

\bibitem[Schmidhuber(1997)]{Schmidhuber:97interesting}
J.~Schmidhuber.
\newblock What's interesting?
\newblock Technical Report IDSIA-35-97, IDSIA, 1997.
\newblock ftp://ftp.idsia.ch/pub/juergen/interest.ps.gz; extended abstract in
  Proc. Snowbird'98, Utah, 1998; see also \cite{Schmidhuber:02predictable}.

\bibitem[Schmidhuber(2002)]{Schmidhuber:02predictable}
J.~Schmidhuber.
\newblock Exploring the predictable.
\newblock In A.~Ghosh and S.~Tsuitsui, editors, \emph{Advances in Evolutionary
  Computing}, pages 579--612. Springer, 2002.

\bibitem[Schmidhuber(2007)]{schmidhuber2007simple}
J.~Schmidhuber.
\newblock Simple algorithmic principles of discovery, subjective beauty,
  selective attention, curiosity \& creativity.
\newblock In \emph{International conference on discovery science}, pages
  26--38. Springer, 2007.

\bibitem[Schmidhuber(2010)]{schmidhuber2010formal}
J.~Schmidhuber.
\newblock Formal theory of creativity, fun, and intrinsic motivation
  (1990--2010).
\newblock \emph{IEEE transactions on autonomous mental development}, 2\penalty0
  (3):\penalty0 230--247, 2010.

\bibitem[Schulman et~al.(2017)Schulman, Wolski, Dhariwal, Radford, and
  Klimov]{schulman2017proximal}
J.~Schulman, F.~Wolski, P.~Dhariwal, A.~Radford, and O.~Klimov.
\newblock Proximal policy optimization algorithms.
\newblock \emph{arXiv preprint arXiv:1707.06347}, 2017.

\bibitem[Sekar et~al.(2020)Sekar, Rybkin, Daniilidis, Abbeel, Hafner, and
  Pathak]{sekar2020planning}
R.~Sekar, O.~Rybkin, K.~Daniilidis, P.~Abbeel, D.~Hafner, and D.~Pathak.
\newblock Planning to explore via self-supervised world models.
\newblock In \emph{International Conference on Machine Learning}, pages
  8583--8592. PMLR, 2020.

\bibitem[Shyam et~al.(2019)Shyam, Ja{\'s}kowski, and Gomez]{shyam2019model}
P.~Shyam, W.~Ja{\'s}kowski, and F.~Gomez.
\newblock Model-based active exploration.
\newblock In \emph{International conference on machine learning}, pages
  5779--5788. PMLR, 2019.

\bibitem[Storck et~al.(1995)Storck, Hochreiter, Schmidhuber,
  et~al.]{storck1995reinforcement}
J.~Storck, S.~Hochreiter, J.~Schmidhuber, et~al.
\newblock Reinforcement driven information acquisition in non-deterministic
  environments.
\newblock In \emph{Proceedings of the international conference on artificial
  neural networks, Paris}, volume~2, pages 159--164. Citeseer, 1995.

\bibitem[Strehl and Littman(2008)]{strehl2008analysis}
A.~L. Strehl and M.~L. Littman.
\newblock An analysis of model-based interval estimation for markov decision
  processes.
\newblock \emph{Journal of Computer and System Sciences}, 74\penalty0
  (8):\penalty0 1309--1331, 2008.

\bibitem[Strens(2000)]{strens2000bayesian}
M.~Strens.
\newblock A bayesian framework for reinforcement learning.
\newblock In \emph{ICML}, volume 2000, pages 943--950, 2000.

\bibitem[Sun et~al.(2011)Sun, Gomez, and Schmidhuber]{sun2011planning}
Y.~Sun, F.~Gomez, and J.~Schmidhuber.
\newblock Planning to be surprised: Optimal bayesian exploration in dynamic
  environments.
\newblock In \emph{International conference on artificial general
  intelligence}, pages 41--51. Springer, 2011.

\bibitem[Sutton(1988)]{sutton1988learning}
R.~S. Sutton.
\newblock Learning to predict by the methods of temporal differences.
\newblock \emph{Machine learning}, 3\penalty0 (1):\penalty0 9--44, 1988.

\bibitem[Sutton(1990)]{sutton1990integrated}
R.~S. Sutton.
\newblock Integrated architectures for learning, planning, and reacting based
  on approximating dynamic programming.
\newblock In \emph{Machine learning proceedings 1990}, pages 216--224.
  Elsevier, 1990.

\bibitem[Sutton and Tanner(2005)]{sutton2005temporal}
R.~S. Sutton and B.~Tanner.
\newblock Temporal-difference networks.
\newblock In \emph{Advances in neural information processing systems}, pages
  1377--1384, 2005.

\bibitem[Sutton et~al.(2011)Sutton, Modayil, Delp, Degris, Pilarski, White, and
  Precup]{sutton2011horde}
R.~S. Sutton, J.~Modayil, M.~Delp, T.~Degris, P.~M. Pilarski, A.~White, and
  D.~Precup.
\newblock Horde: a scalable real-time architecture for learning knowledge from
  unsupervised sensorimotor interaction.
\newblock In L.~Sonenberg, P.~Stone, K.~Tumer, and P.~Yolum, editors,
  \emph{10th International Conference on Autonomous Agents and Multiagent
  Systems {(AAMAS} 2011), Taipei, Taiwan, May 2-6, 2011, Volume 1-3}, pages
  761--768, 2011.

\bibitem[Sutton et~al.(2022)Sutton, Machado, Holland, Timbers, Tanner, and
  White]{sutton2022reward}
R.~S. Sutton, M.~C. Machado, G.~Z. Holland, D.~S.~F. Timbers, B.~Tanner, and
  A.~White.
\newblock Reward-respecting subtasks for model-based reinforcement learning.
\newblock \emph{arXiv preprint arXiv:2202.03466}, 2022.

\bibitem[Talvitie(2014)]{talvitie2014model}
E.~Talvitie.
\newblock Model regularization for stable sample rollouts.
\newblock In \emph{UAI}, pages 780--789, 2014.

\bibitem[Thrun(1992)]{thrun1992efficient}
S.~B. Thrun.
\newblock Efficient exploration in reinforcement learning.
\newblock 1992.

\bibitem[Ulyanov et~al.(2018)Ulyanov, Vedaldi, and Lempitsky]{ulyanov2018deep}
D.~Ulyanov, A.~Vedaldi, and V.~Lempitsky.
\newblock Deep image prior.
\newblock In \emph{Proceedings of the IEEE conference on computer vision and
  pattern recognition}, pages 9446--9454, 2018.

\bibitem[Van~Seijen et~al.(2019)Van~Seijen, Fatemi, and Tavakoli]{van2019using}
H.~Van~Seijen, M.~Fatemi, and A.~Tavakoli.
\newblock Using a logarithmic mapping to enable lower discount factors in
  reinforcement learning.
\newblock \emph{Advances in Neural Information Processing Systems}, 32, 2019.

\bibitem[Veeriah et~al.(2019)Veeriah, Hessel, Xu, Lewis, Rajendran, Oh, van
  Hasselt, Silver, and Singh]{veeriah2019discovery}
V.~Veeriah, M.~Hessel, Z.~Xu, R.~Lewis, J.~Rajendran, J.~Oh, H.~van Hasselt,
  D.~Silver, and S.~Singh.
\newblock Discovery of useful questions as auxiliary tasks.
\newblock \emph{arXiv preprint arXiv:1909.04607}, 2019.

\bibitem[Willems(2017)]{rl_starter_files}
L.~Willems.
\newblock Rl starter files.
\newblock \url{https://github.com/lcswillems/rl-starter-files}, 2017.

\bibitem[Williams and Zipser(1995)]{williams1995gradient}
R.~J. Williams and D.~Zipser.
\newblock Gradient-based learning algorithms for recurrent networks and their
  computational complexity.
\newblock \emph{Backpropagation: Theory, architectures, and applications},
  433:\penalty0 17, 1995.

\bibitem[Zha et~al.(2021)Zha, Ma, Yuan, Hu, and Liu]{zha2021rank}
D.~Zha, W.~Ma, L.~Yuan, X.~Hu, and J.~Liu.
\newblock Rank the episodes: A simple approach for exploration in
  procedurally-generated environments.
\newblock \emph{arXiv preprint arXiv:2101.08152}, 2021.

\bibitem[Zhang et~al.(2021)Zhang, Xu, Wang, Wu, Keutzer, Gonzalez, and
  Tian]{zhang2021noveld}
T.~Zhang, H.~Xu, X.~Wang, Y.~Wu, K.~Keutzer, J.~E. Gonzalez, and Y.~Tian.
\newblock Noveld: A simple yet effective exploration criterion.
\newblock \emph{Advances in Neural Information Processing Systems}, 34, 2021.

\bibitem[Zheng et~al.(2021)Zheng, Veeriah, Vuorio, Lewis, and
  Singh]{zheng2021learning}
Z.~Zheng, V.~Veeriah, R.~Vuorio, R.~L. Lewis, and S.~Singh.
\newblock Learning state representations from random deep action-conditional
  predictions.
\newblock \emph{Advances in Neural Information Processing Systems}, 34, 2021.

\end{thebibliography}

\clearpage
\section*{Checklist}

\begin{enumerate}

\item For all authors...
\begin{enumerate}
  \item Do the main claims made in the abstract and introduction accurately reflect the paper's contributions and scope?
    \answerYes{}
  \item Did you describe the limitations of your work?
    \answerYes{See \autoref{discussion}.}
  \item Did you discuss any potential negative societal impacts of your work?
    \answerYes{See \autoref{discussion}.}
  \item Have you read the ethics review guidelines and ensured that your paper conforms to them?
    \answerYes{}
\end{enumerate}

\item If you are including theoretical results...
\begin{enumerate}
  \item Did you state the full set of assumptions of all theoretical results?
    \answerNA{}
	\item Did you include complete proofs of all theoretical results?
    \answerNA{}
\end{enumerate}

\item If you ran experiments...
\begin{enumerate}
  \item Did you include the code, data, and instructions needed to reproduce the main experimental results (either in the supplemental material or as a URL)?
    \answerYes{Included in the supplementary material. An open-source implementation will be made available online.}
  \item Did you specify all the training details (e.g., data splits, hyperparameters, how they were chosen)?
    \answerYes{Implementation details and hyperparameters are provided in Appendices \ref{sec:appendix_implementation_db_locks} and \ref{sec:appendix_implementation}}
	\item Did you report error bars (e.g., with respect to the random seed after running experiments multiple times)?
    \answerYes{We include 95\% bootstrapped confidence intervals over 5 or 10 seeds for all of our results.}
	\item Did you include the total amount of compute and the type of resources used (e.g., type of GPUs, internal cluster, or cloud provider)?
    \answerYes{Details provided in Appendix~\ref{sec:appendix_implementation}}
\end{enumerate}

\item If you are using existing assets (e.g., code, data, models) or curating/releasing new assets...
\begin{enumerate}
  \item If your work uses existing assets, did you cite the creators?
    \answerYes{}
  \item Did you mention the license of the assets?
    \answerYes{Provided in Appendix \ref{sec:appendix_envs:minigrid} with the description of the environment.}
  \item Did you include any new assets either in the supplemental material or as a URL?
    \answerNA{}
  \item Did you discuss whether and how consent was obtained from people whose data you're using/curating?
    \answerNA{}
  \item Did you discuss whether the data you are using/curating contains personally identifiable information or offensive content?
    \answerNA{}
\end{enumerate}

\item If you used crowdsourcing or conducted research with human subjects...
\begin{enumerate}
  \item Did you include the full text of instructions given to participants and screenshots, if applicable?
    \answerNA{}
  \item Did you describe any potential participant risks, with links to Institutional Review Board (IRB) approvals, if applicable?
    \answerNA{}
  \item Did you include the estimated hourly wage paid to participants and the total amount spent on participant compensation?
    \answerNA{}
\end{enumerate}

\end{enumerate}

\clearpage
\appendix
\section{Environments}
\label{sec:appendix_envs}

In the following sections we describe the environments used in our experiments.

\subsection{Diabolical lock}
\label{sec:appendix_envs:db_lock}

\begin{figure}[t]
\centering
    \begin{floatrow}[1]
      \ffigbox{\caption{Based on Figure 5 from \citet{misra2020kinematic}. The diabolical lock environment is organised in 3 rows $\{a,b,c\}$ and $H$ columns. The green states of rows $a$ and $b$ correspond to the `good' states. 
      The red states of row $c$ correspond to the `dead' states. 
      The grey square denotes the terminal state. 
      The black arrows represent deterministic transitions to the `dead' row of the next column. 
      The green dotted arrows denote the equally likely transitions to one of two good states by taking the `good' action. 
      The yellow arrows mark the final transitions which provide the optimal sparse reward upon termination. 
      \svs{keep as figure 5}}\label{fig:db_lock}}
      {
        \includegraphics[width=0.9\textwidth]{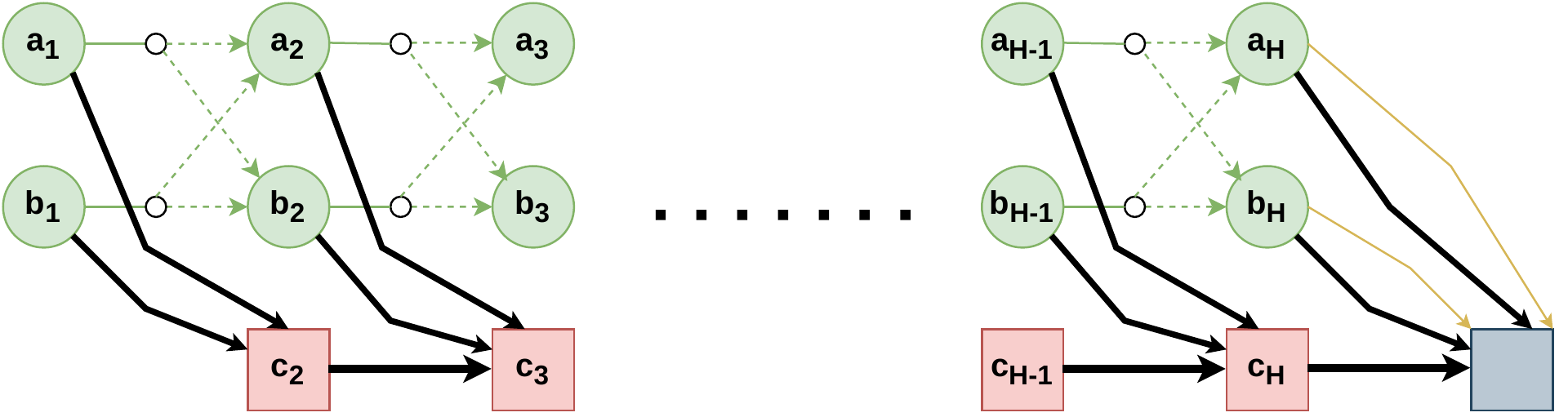}
      }
      \end{floatrow}
\end{figure}

The diabolical lock problem~\citep{misra2020kinematic} consists of states organised in 3 rows $\{a,b,c\}$ and $H$ columns from $\{1, 2, \dots H \}$ (see \autoref{fig:db_lock}).
In each episode, the agent is initialised in one of two possible starting states, either at $a_1$ or $b_1$ with equal probability.
Each episode lasts for $H$ time steps.

\paragraph{Transition dynamics}
In each state the agent has $L$ available actions, only one of which is `good' and transitions to a `good' state (green) in the next column with equal probability. 
For instance, the good action in state $a_t$ leads to $a_{t+1}$ or $b_{t+1}$ with equal probability.
Taking the good action gives a negative (anti-shaped) reward of $-1/H$, except at the end of the lock (states $a_H$ or $b_H$), where the sparse optimal reward of $10$ is received.
The remaining $L-1$ actions in any state are `bad,' causing a deterministic transition to the `dead' row $c$ (red) at the next column (with zero reward). 
The agent remains in the `dead' row $c$ till the end of the episode, i.e., all actions from $c_t$ transition to $c_{t+1}$ with zero reward.
The good action in each state is assigned uniformly at random from the $L$ available actions upon initialising the environment.
The good action in every state remains fixed across episodes.

\paragraph{Noisy observations}
The diabolical lock environment provides a high-dimensional noisy observation based on the environmental state.
The observation is obtained by concatenating one-hot vectors encoding the row and column, adding independent Gaussian noise with variance $\sigma_o^2$ to each component of the vector,
padding this vector with zeroes to the nearest higher power of two and then applying a Hadamard rotation matrix.
~\lk{what's the purpose of the hadamard?}

In our experiments, we select $H=100$, $L=10$, and $\sigma_o=0.1$, which is similar to the configuration used by~\citet{misra2020kinematic}. 
We were unable to find a publicly available implementation of the environment introduced by~\citet{misra2020kinematic} and therefore used our own (no license).

\subsection{MiniGrid}
\label{sec:appendix_envs:minigrid}
The MiniGrid \citep{gym_minigrid} (Apache License 2.0) set of environments are partially observable, procedurally generated, and provide a sparse reward upon successful completion of the task.

The agent receives a $7 \times 7$ egocentric view of its surroundings, which is a $7 \times 7 \times 3$ integer encoding that describes the contents of the visible grid cells. 
The agent cannot see behind walls or closed doors.

For the study with \emph{panoramic observations}, we follow the implementation of \citet{parisi2021interesting} where a rotation invariant view is created by concatenating the four observations from when the agent faces north, east, south and west to create a $7 \times 7 \times 12$ panoramic observation.

At every step, the agent has seven available actions: turn left, turn right, move forward, pick up, drop, toggle (used to open doors), and done.

A new configuration of the environment is generated for each episode. Episodes finish when the agent has interacted with the environment for the maximum number of steps permitted, or if the agent successfully completes the task, in which case it also receives a non-zero reward depending on the number of steps taken to complete the task.

\paragraph{KeyCorridor}
The agent has to pick up a ball behind a locked door. 
To unlock the door, the agent must first find the key, which is also hidden behind closed (but unlocked) doors. 
We consider two levels of difficulty in this class of environments.
The KeyCorridor-S5R3 environment poses a harder exploratory challenge due to larger rooms and hallways in comparison to KeyCorridor-S4R3.

\paragraph{ObstructedMaze}
Like with the KeyCorridor environments, the goal of the agent is to pick up a ball behind one of the locked doors. 
We again consider two levels of difficulty in this class of environments.
In the ObstructedMaze-2Dlh environment the keys are hidden inside chests. 
In ObstructedMaze-2Dlhb the keys are hidden inside chests and the doors are blocked by obstacles that need to be displaced.

\paragraph{MultiRoom}
The goal of the agent is to reach the goal state in the farthest room. 
The notation N12-S10 signifies that the environment consists of 12 rooms with a maximum size of 10.
These environments (MultiRoom-N7-S8 and MultiRoom-N12-S10) are not available in the default set of MiniGrid environments. 
We borrow the implementation of these environments from previous work~\citep{raileanu2020ride}(Creative Commons
Attribution-NonCommercial 4.0 International Public License).

\section{Implementation details for diabolical lock experiments}
\label{sec:appendix_implementation_db_locks}

This section presents the implementation details for our experiments with the diabolical lock environment.
In the following, we describe the architectures and hyperparameters of our actor, critic, pseudo-reward generator, and ensemble of predictors used in these experiments.

\subsection{Neural network architectures}
\paragraph{Base agent}

We use the PPO~\citep{schulman2017proximal} implementation from the \texttt{rl-starter-files} repository~\citep{rl_starter_files}.
The actor and critic share a multi-layer perceptron (MLP) of three layers with 256 units and ReLU non-linearities.
Separate linear heads for the actor and critic operate on the representation provided by the shared architecture.

\paragraph{Pseudo-reward generator}
The randomly initialised fixed network for generating pseudo-rewards is a two layer MLP with ReLU non-linearities in the hidden layer of 128 units. 
The linear output layer provides a vector of 128 pseudo-rewards. 

\paragraph{Predictor networks}

The predictor network for RND has the same architecture as the pseudo-reward generator.

The predictor network for RC-GVF is an MLP (no recurrence here) with wider layers and one extra layer compared to the pseudo-reward generating network.
Specifically, a predictor consists of a network of three layers with 256 units each with ReLU non-linearities in the two hidden layers.
We use an ensemble of $K=2$ predictors in the case of RC-GVF. 
In these experiments with the diabolical lock, the individual predictors in the ensemble are completely separate neural networks which do not share parameters. 

\subsection{Grid search and hyperparameters}
\label{sec:appendix_implementation_db_locks:hparams}

Hyperparameters of the base PPO agent are provided in \autoref{tab:hgrid_ppo_locks}.
These were selected based on previous work~\citep{misra2020kinematic} and apply to our method and RND.

\begin{table*}[ht]
    \small
    \centering
    \caption{PPO hyperparameters for diabolical lock experiments.}
    \label{tab:hgrid_ppo_locks}
    \begin{tabular}{l c c}
      \toprule
      Hyperparameter & Value \\
      \midrule
      Discount factor & 0.99 \\
      GAE lambda & 0.95 \\
      Optimization epochs &  5 \\
      Learning rate ($\eta$) &  0.0005 \\
      Clipping epsilon &  0.2 \\
      Rollout length (frames per process) & $H=100$ \\
      Batch size & 256 \\
      Parallel processes & 16 \\
      Entropy coefficient & 0.01 \\
      Value loss coefficient & 0.5 \\
      \bottomrule
    \end{tabular}
\end{table*}

Utilising an intrinsic reward introduces additional hyperparameters.
We conducted a grid-search to select the values for hyperparameters introduced by RND and RC-GVF.
The best hyperparameter configuration was selected by evaluating average lock completion for each candidate in the grid across 5 seeds.
In \autoref{tab:db_lock_results}, we report the final results with 10 independent runs of the selected hyperparameters.

For RND, the hyperparameters include the intrinsic coefficient $\beta$ and the learning rate of the predictor $\eta_p$.
The search candidates and the final selected values are described in \autoref{tab:hgrid_rnd_locks}.

\begin{table*}[ht]
    \small
    \centering
    \caption{RND specific hyperparameters for diabolical lock experiments.}
    \label{tab:hgrid_rnd_locks}
    \begin{tabular}{l c c}
      \toprule
      Hyperparameter & Candidates & Selected \\
      \midrule
      Predictor learning rate $\eta_p$ & $\{\eta, 0.75 \eta, 0.5 \eta, 0.25 \eta, 0.1 \eta \}$ & $0.25 \eta$ ($\eta=0.0005$) \\
      Intrinsic reward coefficient $\beta$ & $\{4.0, 2.0, 1.0, 0.5, 0.25\}$ & 0.5 \\
      \bottomrule
    \end{tabular}
\end{table*}

In comparison to RND, RC-GVF introduces additional hyperparameters such as the GVF discount factor $\gamma_z$, the associated bias-variance tradeoff parameter for the TD-targets $\lambda_z$, and the number of predictors in the ensemble. 
We observed good results with our default values of $\gamma_z=0.6$ and an ensemble of 2 predictors.
The search candidates and the final selected values for RC-GVF are described in \autoref{tab:hgrid_rcgvf_locks}.

\begin{table*}[ht]
    \small
    \centering
    \caption{RC-GVF specific hyperparameters for diabolical lock experiments.}
    \label{tab:hgrid_rcgvf_locks}
    \begin{tabular}{l c c}
      \toprule
      Hyperparameter & Candidates & Selected \\
      \midrule
      Number of predictors & $\{2\}$ & 2 \\
      Predictor learning rate $\eta_p$ & $\{\eta , 0.5\eta , 0.25\eta \}$ & $0.5 \eta$ ($\eta=0.0005$) \\
      Intrinsic reward coefficient $\beta$ & $\{4.0, 2.0, 1.0, 0.5\}$ & 2.0 \\
      GVF discount factor $\gamma_z$ & $\{0.6\}$ & 0.6 \\
      \bottomrule
    \end{tabular}
\end{table*}

The learning rates of the base agent and predictor are linearly annealed to zero over 100 million frames.
We used the Adam optimiser~\citep{kingma2015adam}
for the PPO agent and the predictor in all experiments. 

\section{Implementation details for Minigrid experiments}
\label{sec:appendix_implementation}

This section presents the implementation details for our experiments with the MiniGrid environments.

\subsection{Neural network architectures}
\label{sec:appendix_implementation:nn_archs}
\paragraph{Base agent}
We use the PPO implementation from the \texttt{rl-starter-files}~\citep{rl_starter_files} 
repository.
We retain the same architecture for the actor and critic as provided in their implementation. 

The actor and the critic share three convolution layers with 16, 32 and 64 output channels, with $2 \times 2$ kernels and a stride of 1. 
Each convolution layer is followed by a ReLU non-linearity. A $2 \times 2$ max-pooling is applied after the first convolution layer. 
The output from the convolution layers is flattened and provided to an LSTM~\citep{hochreiter1997long} with a hidden dimension of 64.
Separate multi-layer perceptron (MLP) heads for the actor and critic operate on the output of the LSTM to output the action logits and the value. 
Each MLP has 64 hidden units with a hyperbolic tangent (tanh) activation function.

\paragraph{Pseudo-reward generator}

The pseudo-reward generator is implemented as a convolutional neural network whose output is flattened to provide a vector of pseudo-rewards $z_{t+1} \in \mathbb{R}^d$. 
It has a similar architecture to the convolutional layers used in the base agent.

Concretely, this network consists of three convolution layers, with 32, 64, and $d$ output channels respectively, with $2 \times 2$ kernels and a stride of 1. 
Here, $d$ is the number of pseudo-rewards, which we set as $d=128$ for our experiments. 
ReLU non-linearity is applied after the first two convolution layers. 
Additionally, a $2 \times 2$ max-pooling is applied after the first convolution layer.

\paragraph{Predictor networks}

In the case of RND, the predictor network has the same architecture as the pseudo-reward generator.

For RC-GVF, we have an ensemble of predictors with $K$ members. 
An ensemble of $K$ two layer MLP heads (with 256 units and ReLU non-linearity) operate on a common representation of 128 dimensions generated by a shared recurrent (LSTM) core. The LSTM operates on histories of observations, actions, and pseudo-rewards.

The observation at time step $o_t$ is fed through three convolution layers with 32, 64 and 128 output channels, with $2 \times 2$ kernels and a stride of 1. 
Each convolution layer is followed by a ReLU non-linearity. 
A $2 \times 2$ max-pooling is applied after the first convolution layer.
The output from the convolution layers is flattened and provided to a fully connected layer which provides a 64 dimensional representation of the input observation.

The action at the previous time step $a_{t-1}$ (as a one-hot input) is transformed into a 32 dimensional vector by a fully connected layer.
A fully connected layer is also applied to the previous pseudo-reward vector $z_t$ to obtain a 32 dimensional vector. The observation, action and pseudo-reward representations are concatenated into a 128 dimensional vector which is fed as input to the LSTM (with hidden dimension of 128). 

\subsection{Details about baseline implementations}
\label{sec:appendix_implementation:baselines}

We employed the same base PPO agent (see Appendix \ref{sec:appendix_implementation:nn_archs}) for all approaches to maintain consistency. 
This means that our implementation of baselines deviate slightly from the original implementations. 
Nevertheless, we obtain comparable or better results than the reported ones on the considered environments.
We describe the key details of our AGAC~\citep{flet2021adversarially} and NovelD~\citep{zhang2021noveld} implementations below.

\paragraph{AGAC}
The PPO implementation of AGAC utilised independent convolutional neural networks for the actor and critic with frame stacking rather than a shared recurrent neural network (for actor and critic) as used by us.

In our implementation, the policy predictor (or adversary) has the same architecture as the recurrent actor-critic (see Appendix \ref{sec:appendix_implementation:nn_archs}), but without the MLP head for the critic.

\paragraph{NovelD}

NovelD derives intrinsic rewards using RND as a novelty measure. The neural network architecture details for NovelD therefore follow those described for RND (see Appendix \ref{sec:appendix_implementation:nn_archs}).

A major difference is that the original implementation of NovelD used IMPALA~\citep{espeholt2018impala}
as the base agent instead of PPO.

Another deviation in our version of NovelD is that we do not use the \emph{Episodic Restriction on Intrinsic Reward (ERIR)} multiplier for using episodic counts. 
The ERIR multiplier uses episodic counts from the simulator to ensure that intrinsic rewards are provided only when the agent visits a state for the first time in an episode.
For consistency, we use episodic counts in a similar manner to RIDE~\citep{raileanu2020ride} and AGAC~\citep{flet2021adversarially}, where the square root of episodic count divides the main intrinsic reward term. Our NovelD intrinsic reward is given by\footnote{We note that the authors of NovelD mention that replacing the ERIR term with scaling based on the square root of episodic counts still outperforms previous baselines (\url{https://openreview.net/forum?id=CYUzpnOkFJp}).}
\begin{equation}
\label{eq:noveld_ir}
    R_i(o_t; s_t, o_{t+1}) = \frac{\max \left [ novelty(o_{t+1}) - \alpha \cdot novelty(o_{t}), 0 \right ]}{\sqrt{N_e(s_t)}},
\end{equation}
where $novelty(.)$ is measured as the RND prediction error (see Equation \ref{eq:rnd}), $\alpha$ is a hyperparameter called the scaling factor, and $N_e(s_t)$ is the number of times the agent has been in that state in that episode.

\paragraph{On the use of episodic counts}
Note that in our experiments the episodic counts refer to agent positions alone, following AGAC~\citep{flet2021adversarially}. 
Other approaches have used different strategies to obtain episodic counts from the simulator.
For example, RIDE~\citep{raileanu2020ride} and NovelD~\citep{zhang2021noveld} derive counts from fully observable views which provide count information based on agent position, direction or whether an object has been picked up etc. 

\subsection{Grid search and selected hyperparameters}
\label{sec:appendix_implementation:hyperparameters}

In this section we describe our hyperparameter selection strategy for the MiniGrid experiments. 

Our implementation utilises the PPO implementation from the \texttt{rl-starter-files}~\citep{rl_starter_files} 
repository. 
The hyperparameters for the common PPO base agent were selected via preliminary experiments and are similar to the ones provided in the \texttt{rl-starter-files} repository.
These are detailed in \autoref{tab:hgrid_ppo_minigrid}. 

\begin{table*}[ht]
    \small
    \centering
    \caption{PPO hyperparameters for MiniGrid experiments.}
    \label{tab:hgrid_ppo_minigrid}
    \begin{tabular}{l c c}
      \toprule
      Hyperparameter & Value \\
      \midrule
      Discount factor & 0.99 \\
      GAE lambda & 0.95 \\
      Optimization epochs &  4 \\
      Learning rate ($\eta$) &  0.0002 \\
      Clipping epsilon &  0.2 \\
      Rollout length (frames per process) & 128 \\
      Batch size & 256 \\
      Parallel processes & 16 \\
      Value loss coefficient & 0.5 \\
      Recurrence (for truncated BPTT~\citep{williams1995gradient}) & 4 \\
      \bottomrule
    \end{tabular}
\end{table*}

We choose the hyperparameters introduced by each intrinsic reward approach via a grid search. 
We select separate hyperparameters for each environment type: MultiRoom, KeyCorridor and ObstructedMaze. 
To select hyperparameters, we evalaute each candidate configuration with 5 seeds on the simpler versions of the considered environments (MultiRoom-N7S8, KeyCorridor-S4R3 and ObstructedMaze-2Dlh) and select the configuration with the highest average return.

Final results are reported with separate independent runs on these environments (usually with 10 seeds) and also the harder versions of the considered environments (MultiRoom-N12S10, KeyCorridor-S5R3 and ObstructedMaze-2Dlhb).

All experiments used the Adam optimiser~\citep{kingma2015adam} for the PPO agent and the respective predictors.

Similar to previous approaches~\citep{raileanu2020ride, flet2021adversarially}, the learning rates are linearly annealed to zero while interacting with the environment during 10 million frames for the MultiRoom environments, 30 million frames for the KeyCorridor and ObstructedMaze-2Dlh environments, and 90 million frames for the ObstructedMaze-2Dlhb environment.

\subsubsection{RC-GVF}

We selected the GVF discount factor $\gamma_z=0.6$ and the associated $\lambda$-return parameter $\lambda_z=0.9$ as default values for all experiments. We found that $K=2$ predictors in the ensemble worked sufficiently well in preliminary experiments.
We identified the entropy coefficient, intrinsic reward coefficient and learning rate of the predictor through a grid search. The candidate and selected values are described in Tables \ref{tab:hgrid_rcgvf_minigrid} and \ref{tab:hgrid_rcgvf_minigrid_pano} for the setting with egocentric and panoramic observations respectively.

\begin{table*}[!h]
    \small
    \centering
    \caption{RC-GVF specific hyperparameters for MiniGrid experiments with egocentric observations.}
    \label{tab:hgrid_rcgvf_minigrid}
    \begin{tabular}{l c cccc}
      \toprule
      & & \multicolumn{3}{c}{Egocentric observations} \\
      \cmidrule(lr){3-5}
      Hyperparameter & Candidates & MultiRoom & KeyCorridor & ObstructedMaze \\
      \midrule
      Number of predictors & $\{2\}$ & 2 & 2 & 2 \\
      GVF discount factor $\gamma_z$ & $\{0.6\}$ & 0.6 & 0.6 & 0.6 \\
      GVF $\lambda_z$ (for TD-$\lambda$ targets) & $\{ 0.9\}$ & 0.9 & 0.9 & 0.9 \\
      Predictor learning rate $\eta_p$ & $\{0.5\eta,  0.75 \eta, \eta \}$ & $ 0.75 \eta$ & $ 0.75 \eta$ & $ 0.75 \eta$ \\
      Intrinsic reward coefficient $\beta$ & $\{0.1, 0.5, 1.0 \}$ & 1.0 & 0.1 & 0.1 \\
      Entropy coefficient & $\{0, 1e-5, 1e-4, 1e-3 \}$ & 1e-5 & 1e-5 & 1e-4 \\
      \bottomrule
    \end{tabular}
\end{table*}

\begin{table*}[!h]
    \small
    \centering
    \caption{RC-GVF specific hyperparameters for MiniGrid experiments with panoramic observations.}
    \label{tab:hgrid_rcgvf_minigrid_pano}
    \begin{tabular}{l c cccc}
      \toprule
      & & \multicolumn{3}{c}{Panoramic observations} \\
      \cmidrule(lr){3-5}
      Hyperparameter & Candidates & MultiRoom & KeyCorridor & ObstructedMaze \\
      \midrule
      Number of predictors & $\{2\}$ & 2 & 2 & 2 \\
      GVF discount factor $\gamma_z$ & $\{0.6\}$ & 0.6 & 0.6 & 0.6 \\
      GVF $\lambda_z$ (for TD-$\lambda$ targets) & $\{ 0.9\}$ & 0.9 & 0.9 & 0.9 \\
      Predictor learning rate $\eta_p$ & $\{0.5\eta,  0.75 \eta, \eta \}$ & $0.5 \eta$ & $0.75 \eta$ & $0.5 \eta$ \\
      Intrinsic reward coefficient $\beta$ & $\{0.1, 0.5, 1.0 \}$ & 0.1 & 1.0 & 0.1\\
      Entropy coefficient & $\{0, 1e-5, 1e-4, 1e-3 \}$ & 1e-5 & 1e-5 & 1e-4 \\
      \bottomrule
    \end{tabular}
\end{table*}

\subsubsection{RND}

Tables \ref{tab:hgrid_rnd_minigrid} and \ref{tab:hgrid_rnd_minigrid_pano} describe the search grid and selected hyperparameters for RND with egocentric and panoramic observations respectively.

\begin{table*}[!h]
    \small
    \centering
    \caption{RND specific hyperparameters for MiniGrid experiments with egocentric observations.}
    \label{tab:hgrid_rnd_minigrid}
    \begin{tabular}{l c cccc}
      \toprule
      & & \multicolumn{3}{c}{Egocentric observations} \\
      \cmidrule(lr){3-5}
      Hyperparameter & Candidates & MultiRoom & KeyCorridor & ObstructedMaze \\
      \midrule
      Predictor learning rate $\eta_p$ & $\{0.5\eta,  0.75 \eta, \eta \}$ & $0.75 \eta$ & $0.75 \eta$ & $0.75 \eta$\\
      Intrinsic reward coefficient $\beta$ &$\{0.0001, 0.0005, 0.001, 0.005\}$  & 0.0005 & 0.0005 & 0.0005 \\
      Entropy coefficient & $\{0, 1e-5, 1e-4, 1e-3 \}$ & 0 & 1e-5 & 1e-4\\
      \bottomrule
    \end{tabular}
\end{table*}

\begin{table*}[!h]
    \small
    \centering
    \caption{RND specific hyperparameters for MiniGrid experiments with panoramic observations.}
    \label{tab:hgrid_rnd_minigrid_pano}
    \begin{tabular}{l c cccc}
      \toprule
      & & \multicolumn{3}{c}{Panoramic observations} \\
      \cmidrule(lr){3-5}
      Hyperparameter & Candidates & MultiRoom & KeyCorridor & ObstructedMaze \\
      \midrule
      Predictor learning rate $\eta_p$ & $\{0.5\eta,  0.75 \eta, \eta \}$ & $0.75 \eta$ & $0.75 \eta$ & $0.75 \eta$ \\
      Intrinsic reward coefficient $\beta$ & $\{0.0001, 0.0005, 0.001, 0.005\}$ & 0.0005 & 0.001 & 0.0001 \\
      Entropy coefficient & $\{0, 1e-5, 1e-4, 1e-3 \}$ & 1e-5 & 1e-5 & 1e-4 \\
      \bottomrule
    \end{tabular}
\end{table*}

\subsubsection{NovelD}

NovelD relies on RND as a novelty measure and therefore has similar hyperparameters. 
We identified the entropy coefficient, intrinsic coefficient and learning rate of the RND predictor through a grid search. 

A new hyperparameter introduced by NovelD is the scaling factor ($\alpha$ in \autoref{eq:noveld_ir}) which was selected as 0.5 in previous work~\citep{zhang2021noveld}. We also found the value of 0.5 to work well in our preliminary experiments and did not search over other values for this hyperparameter.

Tables \ref{tab:hgrid_noveld_minigrid} and \ref{tab:hgrid_noveld_minigrid_counts} describe the search grid and selected hyperparameters for NovelD.

\begin{table*}[!h]
    \small
    \centering
    \caption{NovelD specific hyperparameters for MiniGrid experiments with egocentric observations (without counts).}
    \label{tab:hgrid_noveld_minigrid}
    \begin{tabular}{l c cccc}
      \toprule
      & & \multicolumn{3}{c}{Without episodic counts} \\
      \cmidrule(lr){3-5}
      Hyperparameter & Candidates & MultiRoom & KeyCorridor & ObstructedMaze \\
      \midrule
      Scaling factor $\alpha$ & $\{ 0.5 \}$ & 0.5 & 0.5 & 0.5 \\
      Predictor learning rate $\eta_p$ & $\{0.5 \eta, 0.75 \eta, \eta \}$ & $0.5 \eta$  & $0.75 \eta$ & $0.75 \eta$ \\
      Intrinsic reward coefficient & $\{0.0005, 0.001, 0.005, 0.01\}$ & 0.01 & 0.005 & 0.001\\
      Entropy coefficient & $\{0, 1e-5, 1e-4, 1e-3 \}$ & 0 & 1e-5 & 1e-4 \\
      \bottomrule
    \end{tabular}
\end{table*}

\begin{table*}[!h]
    \small
    \centering
    \caption{NovelD specific hyperparameters for MiniGrid experiments with episodic counts.}
    \label{tab:hgrid_noveld_minigrid_counts}
    \begin{tabular}{l c cccc}
      \toprule
      & & \multicolumn{3}{c}{With episodic counts} \\
      \cmidrule(lr){3-5}
      Hyperparameter & Candidates & MultiRoom & KeyCorridor & ObstructedMaze \\
      \midrule
      Scaling factor $\alpha$ & $\{ 0.5 \}$ & 0.5 & 0.5 & 0.5 \\
      Predictor learning rate $\eta_p$ & $\{0.5 \eta, 0.75 \eta, \eta \}$ & $0.75 \eta$ & $0.75 \eta$ & $0.75 \eta$ \\
      Intrinsic reward coefficient $\beta$ & $\{0.005, 0.01, 0.05, 0.1\}$ & 0.05 & 0.01 & 0.01\\
      Entropy coefficient & $\{0, 1e-5, 1e-4, 1e-3 \}$& 1e-5 & 1e-5 & 1e-4 \\
      \bottomrule
    \end{tabular}
\end{table*}

\subsubsection{AGAC}

We identified the entropy coefficient, intrinsic coefficient and learning rate of the predictor (`adversary' in the original terminology) through a grid search.

AGAC also introduces a hyperparameter called the `episodic count coefficient' to scale the bonus from the episodic counts term. 

Our main results in \autoref{fig:minigrid_results} are obtained without the use of episodic counts (this coefficient is set to zero).
To report results with episodic counts in \autoref{fig:minigrid_results_full} we search over values of this hyperparameter.

Tables \ref{tab:hgrid_agac_minigrid} and \ref{tab:hgrid_agac_minigrid_counts} describe the search grid and selected hyperparameters for AGAC.

Note that in the experiments without episodic counts, AGAC did not succeed (with any search configuration) on any environment other than ObstructedMaze-2Dlh.

\begin{table*}[!h]
    \small
    \centering
    \caption{AGAC specific hyperparameters for MiniGrid experiments with egocentric observations (without counts). }
    \label{tab:hgrid_agac_minigrid}
    \begin{tabular}{l c cccc}
      \toprule
      & & \multicolumn{3}{c}{Without episodic counts} \\
      \cmidrule(lr){3-5}
      Hyperparameter & Candidates & MultiRoom & KeyCorridor & ObstructedMaze \\
      \midrule
      Predictor learning rate $\eta_p$ & $\{0.25 \eta, 0.5 \eta \}$ & $0.25 \eta$ & $0.25 \eta$ & $0.25 \eta$ \\
      Intrinsic coefficient &  $\{0.00001, 0.00005, 0.0001, 0.0005 \}$ & 0.00005 & 0.00005 & 0.00005 \\
      Episodic count coefficient & $\{0 \}$ & 0 & 0 & 0\\
      Entropy coefficient & $\{0, 1e-5, 1e-4, 1e-3 \}$ & 1e-5 & 1e-4 & 1e-4 \\
      \bottomrule
    \end{tabular}
\end{table*}

\begin{table*}[!h]
    \small
    \centering
    \caption{AGAC specific hyperparameters for MiniGrid experiments with episodic counts}
    \label{tab:hgrid_agac_minigrid_counts}
    \begin{tabular}{l c cccc}
      \toprule
      & & \multicolumn{3}{c}{With episodic counts} \\
      \cmidrule(lr){3-5}
      Hyperparameter & Candidates & MultiRoom & KeyCorridor & ObstructedMaze \\
      \midrule
      Predictor learning rate $\eta_p$ & $\{0.25 \eta, 0.5 \eta \}$ & $0.25 \eta$ & $0.25 \eta$ & $0.25 \eta$ \\
      Intrinsic coefficient & $\{0.00001, 0.00005, 0.0001, 0.0005 \}$ & 0.00005 & 0.00005 & 0.00005 \\
      Episodic count coefficient & $\{0.0005, 0.001, 0.005, 0.01\}$ & 0.005, & 0.001 & 0.001 \\
      Entropy coefficient & $\{0, 1e-5, 1e-4, 1e-3 \}$ & 1e-5 & 1e-4 & 1e-4\\
      \bottomrule
    \end{tabular}
\end{table*}

\subsection{Grid search and selected hyperparameters for analysis and ablation}

For the analysis of discount factors presented in Figures \ref{fig:minigrid_discount_analysis:kcs5r3} and \ref{fig:minigrid_discount_analysis:mrn12s10} we conducted a search over best intrinsic reward coefficients ($\beta$) for each GVF discount factor $\gamma_z$.
This was necessary as the magnitudes of the TD-error are sensitive to the choice of  $\gamma_z$.
The candidates and selected intrinsic reward coefficients ($\beta$) are reported in \autoref{tab:hgrid_rcgvf_discount_analysis}.
The remaining hyperparameters are the same as the ones selected for RC-GVF with panoramic views (see \autoref{tab:hgrid_rcgvf_minigrid_pano}).
Final results are reported with 10 independent runs of the selected hyperparameters.

\begin{table*}[ht]
    \small
    \centering
    \caption{Search over intrinsic coefficients for different GVF discount factors. KCS5R3 refers to the KeyCorridor-S5R3 environment and MRN12S10 refers to the MultiRoom-N12-S10 environment.}
    \label{tab:hgrid_rcgvf_discount_analysis}
    \begin{tabular}{c c c c}
      \toprule
      GVF $\gamma_z$ & Intrinsic coefficients ($\beta$) considered & Selected $\beta$ (KCS5R3) & Selected $\beta$ (MRN12S10) \\
      \midrule
      0.99 & $\{$ 1e-6, 5e-6, 1e-5, 5e-5, 1e-4, 5e-4, 1e-3, 5e-3$\}$ & 1e-5 & 1e-4  \\
      0.95 & $\{0.0005, 0.001, 0.005, 0.01, 0.05 \}$ & 0.001 & 0.001 \\
      0.7 & $\{0.1, 0.5, 1 \}$ & 0.5 & 0.1 \\
      0.6 & $\{0.1, 0.5, 1 \}$ & 1 & 0.1 \\
      0.3 & $\{0.5, 1, 5, 10, 25, 50\}$ & 10 & 1 \\
      0.1 & $\{1, 5, 10, 25, 50, 100 \}$& 25 & 5 \\
      0.05 & $\{1, 5, 10, 25, 50, 100 \}$ & 25 & 10 \\
      0.0 & $\{5, 10, 25, 50, 100 \}$ & 50 & 10 \\
      \bottomrule
    \end{tabular}
\end{table*}

We conduct a similar search over choices of intrinsic coefficients for the ablation with components of RC-GVF (see \autoref{fig:minigrid_component_ablation}).
The candidates and selected intrinsic reward coefficients are reported in \autoref{tab:hgrid_rcgvf_ablation}. 
The hyperparameters for RC-GVF ($\gamma_z = 0.6$) and RND are the same as the ones reported in Tables \ref{tab:hgrid_rcgvf_minigrid_pano} and \ref{tab:hgrid_rnd_minigrid_pano} respectively.
Final results are reported with 10 independent runs of the selected hyperparameters.

\begin{table*}[ht]
    \small
    \centering
    \caption{Search over intrinsic coefficients for different variants of RC-GVF. KCS5R3 refers to the KeyCorridor-S5R3 environment.}
    \label{tab:hgrid_rcgvf_ablation}
    \begin{tabular}{c c c}
      \toprule
      Approach & Intrinsic coefficients ($\beta$) considered & Selected $\beta$ (KCS5R3) \\
      \midrule
      RC-GVF ($\gamma_z=0.0$) & $\{5, 10, 25, 50, 100 \}$ & 50  \\
      RC-GVF ($\gamma_z=0.0$) w/o disagreement & $\{ 0.0005, 0.001, 0.005, 0.01 \}$ & 0.005  \\
      RC-GVF ($\gamma_z=0.0$) w/o recurrence & $\{10, 25, 50, 100 \}$ & 50 \\
      RC-GVF ($\gamma_z=0.6$) w/o disagreement & $\{ 0.00001, 0.00005, 0.0001, 0.0005\}$ & 0.00005  \\
      RC-GVF ($\gamma_z=0.6$) w/o recurrence & $\{ 0.1, 0.5, 1\}$&  0.5 \\
      \bottomrule
    \end{tabular}
\end{table*}

\subsection{Computational resources}

All our experiments were conducted on a cluster with NVIDIA Pascal P100 GPUs. Due to low GPU usage of a single run, we could accommodate multiple concurrent runs (usually 5) on the same node (CPU + GPU). Our final experiments should be easy to replicate even on a single GPU machine. A single run of RC-GVF on a desktop machine with an NVIDIA GeForce RTX 2080 GPU completes about 30 million frames in MiniGrid (with egocentric observations) in about 9 hours (equivalent to our KeyCorridor experiments).

\section{Additional experiments with the diabolical lock problem}
\label{sec:appendix:db_lock_analysis}

\subsection{Further baselines}
Our aim with the diabolical lock experiment is to illustrate the differences between RND and RC-GVF. 
For completeness, in this section, we additionally report the performance of AGAC and NovelD.
We also investigate the performance of RND with a higher capacity predictor, as used by RC-GVF (see Appendix \ref{sec:appendix_implementation_db_locks}).

We conduct a grid-search to select the values for hyperparameters specific to the new baselines.
These are described in Tables \ref{tab:hgrid_rnd_big_locks}, \ref{tab:hgrid_noveld_locks}, and \ref{tab:hgrid_agac_locks}.
The best configuration is selected by evaluating average lock completion for each candidate in the grid across 5 seeds.
The remaining hyperparameters are the same as the ones detailed in Appendix \ref{sec:appendix_implementation_db_locks:hparams}.

The farthest completions at different points in training (with 10 independent runs of the selected hyperparameters) are reported in \autoref{tab:db_lock_full_results}.
We observe that neither AGAC nor NovelD manage to complete the lock. 
AGAC does not  make much progress towards solving the problem.
NovelD fares better and performs roughly as well as RND.
We also observe that the higher capacity predictor does not improve RND (on average). 

\begin{table*}[htpb]
    \vspace{-5pt}
    \footnotesize
    \centering
    \caption{
    Mean over 10 seeds and (min, max) of the farthest column reached (of $H=100$) at different points on the diabolical lock problem.}
    \begin{tabular}{lccccc}
      \toprule
      Frames & 5M & 10M & 15M & 20M \\
      \midrule
      RC-GVF &  61 (47, 81) & 94 (83, 100)& 100 (100, 100) & 100 (100, 100)  \\
      RND & 28 (21, 41) & 37 (27, 47) & 41 (30, 53) & 46 (33, 58)\\
      RND (with high capacity predictor) & 22 (8, 41) & 28 (8, 62) & 32 (8, 70) & 34 (8, 72)\\
      NovelD & 28 (22, 35) & 35 (26, 50) & 41 (27, 54)  & 45 (32, 60) \\
      AGAC & 5 (5, 7) & 6 (5, 7) & 6 (5, 9) & 6 (5, 9)\\
      \bottomrule
    \end{tabular}
    \vspace{-5pt}
    \label{tab:db_lock_full_results}
\end{table*}

\begin{table*}[htpb]
    \small
    \centering
    \caption{RND (with high capacity predictor) hyperparameters for diabolical lock experiments.}
    \label{tab:hgrid_rnd_big_locks}
    \begin{tabular}{l c c}
      \toprule
      Hyperparameter & Candidates & Selected \\
      \midrule
      Predictor learning rate $\eta_p$ & $\{\eta, 0.75 \eta, 0.5 \eta, 0.25 \eta \}$ &  $0.25 \eta$ ($\eta=0.0005$) \\
      Intrinsic reward coefficient $\beta$ & $\{4.0, 2.0, 1.0, 0.5, 0.25\}$ & 1.0 \\
      \bottomrule
    \end{tabular}
\end{table*}

\begin{table*}[htpb]
    \small
    \centering
    \caption{NovelD specific hyperparameters for diabolical lock experiments.}
    \label{tab:hgrid_noveld_locks}
    \begin{tabular}{l c c}
      \toprule
      Hyperparameter & Candidates & Selected \\
      \midrule
      Predictor learning rate $\eta_p$ & $\{\eta, 0.75 \eta, 0.5 \eta, 0.25 \eta \}$ &  $0.25 \eta$ ($\eta=0.0005$) \\
      Intrinsic reward coefficient $\beta$ & $\{4.0, 2.0, 1.0, 0.5, 0.25\}$ & $1.0$ \\
      Scaling factor $\alpha$ & $\{ 0.5 \}$ & 0.5 \\
      \bottomrule
    \end{tabular}
\end{table*}

\begin{table*}[h!]
    \small
    \centering
    \caption{AGAC specific hyperparameters for diabolical lock experiments.}
    \label{tab:hgrid_agac_locks}
    \begin{tabular}{l c c}
      \toprule
      Hyperparameter & Candidates & Selected \\
      \midrule
      Predictor learning rate $\eta_p$ & $\{0.5 \eta, 0.25 \eta\}$ &  $0.25 \eta$ ($\eta=0.0005$) \\
      Intrinsic reward coefficient $\beta$ & $\{4.0, 2.0, 1.0, 0.5, 0.25\}$ &  1.0 \\
      Shared architecture for actor and critic & $\{ True, False \}$ & $True$\\
      \bottomrule
    \end{tabular}
\end{table*}

\subsection{Ensemble size analysis}
In this section, we analyse the effect of increasing the ensemble size in RC-GVF on the performance in the diabolical lock problem. We consider 2, 4, 6 and 8 predictors in the ensemble. The other hyperparameters are the same as those reported in \autoref{tab:hgrid_rcgvf_locks}. 

The results are reported across 10 seeds in \autoref{tab:db_lock_full_results}.
For the most part, we obtain similar results for the different ensemble sizes. The minor exception is with 6 predictors, where one run out of the ten fails to reach the end of the lock.

\begin{table*}[bhtp]
    \vspace{-5pt}
    \footnotesize
    \centering
    \caption{
    Mean over 10 seeds and (min, max) of the farthest column reached (of $H=100$) at different points during training on the diabolical lock problem.
    Higher is better.}
    \begin{tabular}{ccccc}
      \toprule
      Number of predictors / Frame & 5M & 10M & 15M & 20M \\
      \midrule
      2 &  61 (47, 81) & 94 (83, 100)& 100 (100, 100) & 100 (100, 100)  \\
      4 & 66 (62, 78) & 96 (82, 100) & 100 (100, 100) & 100 (100, 100) \\
      6 &  61 (39, 75) & 90 (59, 100) & 96 (63, 100) & 97 (67, 100) \\
      8 &  69 (57, 84) & 94 (77, 100)  & 100 (99, 100)  & 100 (100, 100) \\
      \bottomrule
    \end{tabular}
    \vspace{-5pt}
    \label{tab:db_lock_full_results}
\end{table*}
\FloatBarrier
\section{Additional experiments in MiniGrid}
\subsection{Intrinsic reward from solely episodic counts}\label{sec:appendix_only_counts}

\begin{figure}[htpb]
\begin{subfigure}{.49\textwidth}
  \centering
  \includegraphics[width=1.0\textwidth]{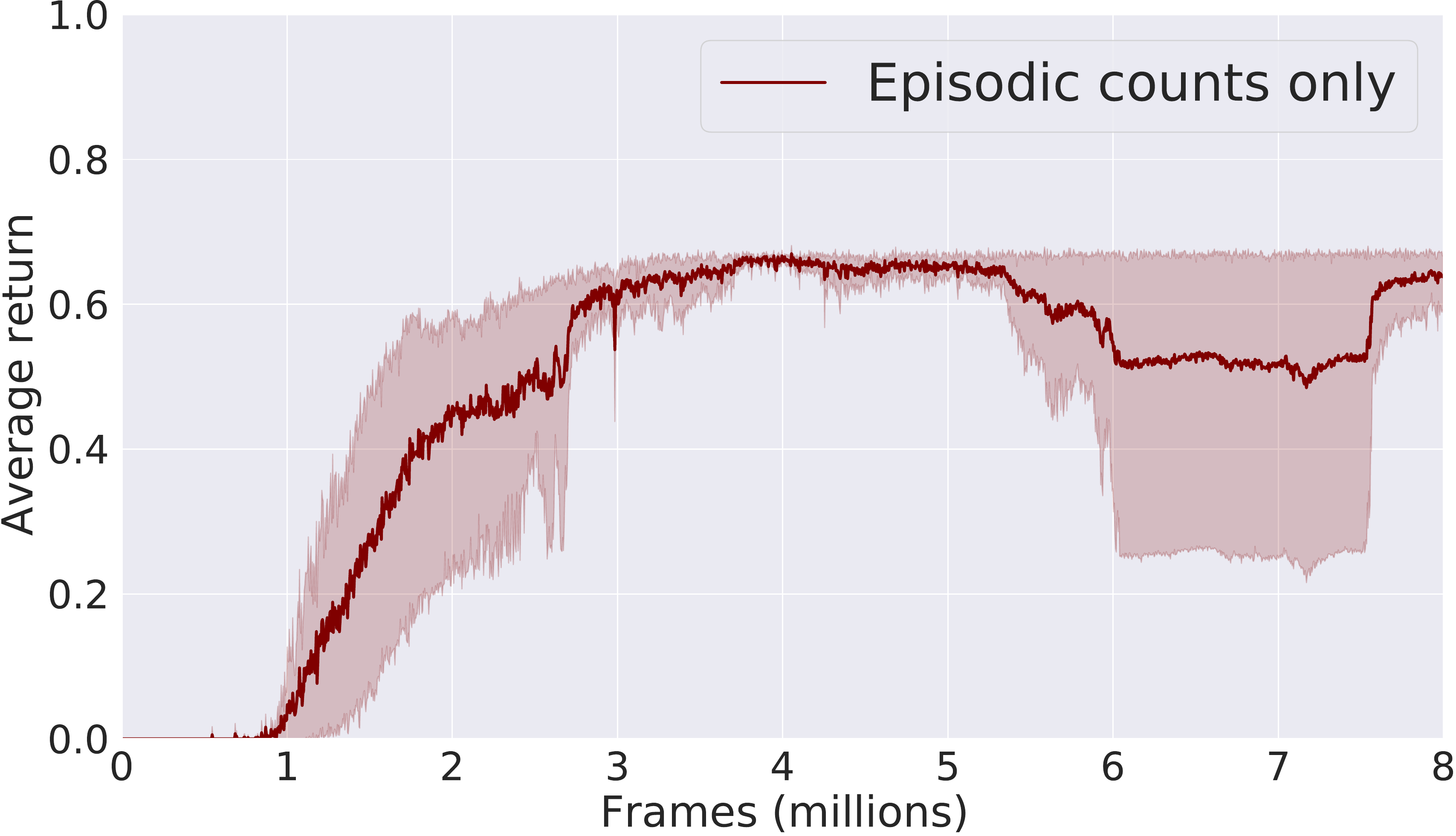}
  \caption{MultiRoom-N7-S8}
  \label{fig:minigrid_episodic_counts:mrn7s8}
\end{subfigure}
\begin{subfigure}{.49\textwidth}
  \centering
  \includegraphics[width=1.0\textwidth]{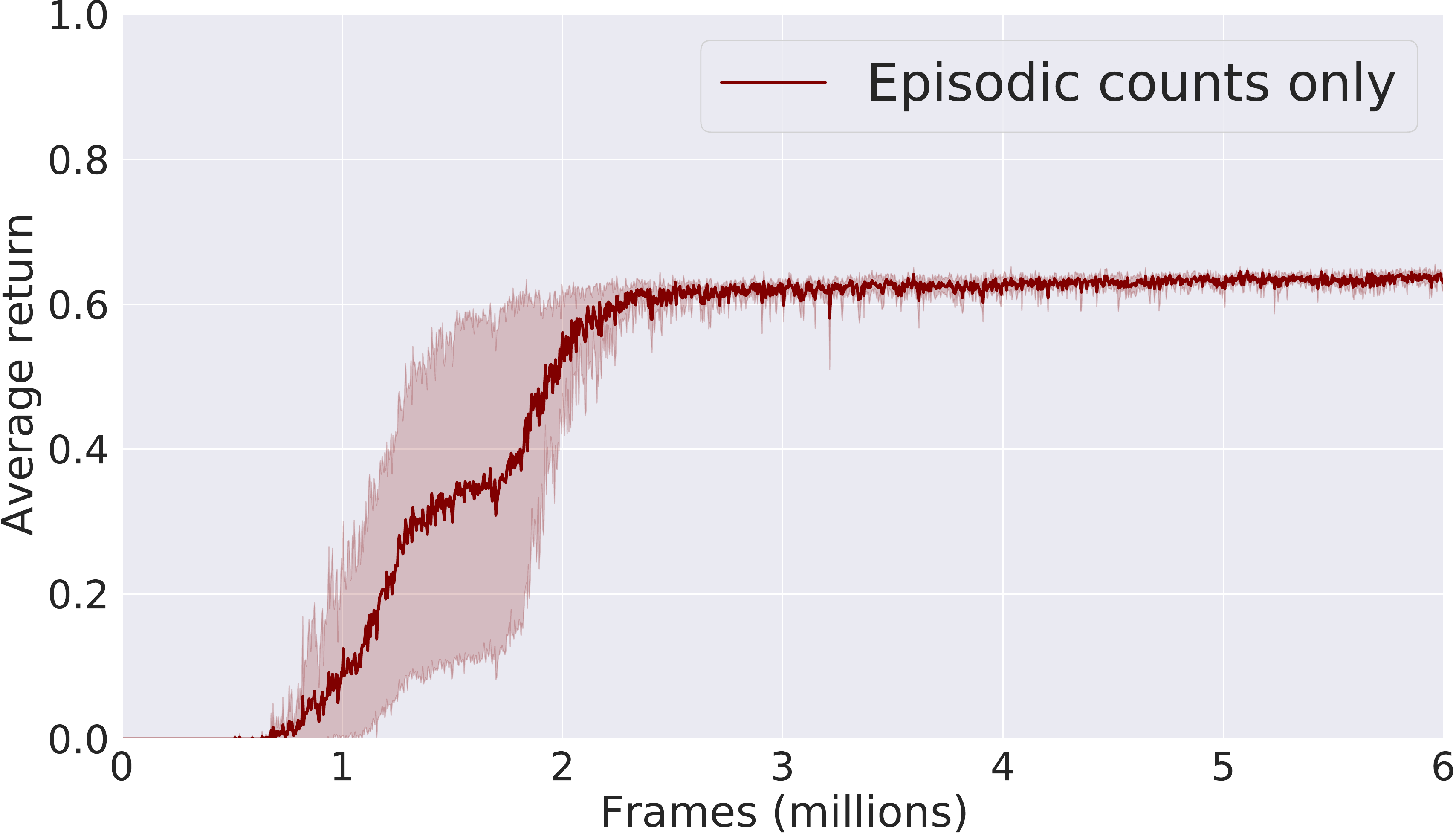}
  \caption{MultiRoom-N12-S10}
  \label{fig:minigrid_episodic_counts:mrn12s10}
\end{subfigure}
\caption{Performance with an intrinsic reward derived from episodic counts. These results show that a bonus term from solely episodic counts can guide useful exploration in MiniGrid environments. 95\% bootstrapped confidence intervals are shown for 5 seeds. \svs{keep as figure 6}
}
\label{fig:minigrid_episodic_counts}
\end{figure}

\autoref{fig:minigrid_episodic_counts} shows that intrinsic rewards from episodic state counts are sufficient for solving the MultiRoom-N7S8 and MultiRoom-N12S10 environments.

The exact form of the intrinsic bonus is $R_i(s_t) =  \frac{\beta}{\sqrt{N_e(s_t)}}$, where $N_e(s_t)$ is the number of times the agent has been in that position in that episode, and $\beta$ is the intrinsic reward coefficient.

We perform a search over the intrinsic reward coefficients $\beta \in \{0.0005, 0.001, 0.005, 0.01\}$ and select $\beta=0.005$ as a suitable value. The entropy coefficient for this experiment is set to 0.00001.

\subsection{RC-GVF with episodic counts}
\label{sec:appendix_rcgvf_epcounts}

Several recent works incorporate episodic counts from the simulator as part of their intrinsic reward in MiniGrid~\citep{raileanu2020ride, flet2021adversarially, zhang2021noveld}.
Our main experiments with RC-GVF do not utilise episodic counts as we do not wish to provide agents with access to privileged information.
For completeness, this section reports the result from a preliminary experiment, where we study the performance of RC-GVF with the use of episodic counts.
We incorporate episodic counts in the same way as for the other baselines (see Appendix \ref{sec:appendix_implementation:baselines}).

To account for change in scale of the intrinsic reward with the use of the counts, we perform a search over the intrinsic reward coefficients $\beta \in \{1, 5, 10\}$ and select $\beta=5$ for KeyCorridor-S5R3 and $\beta=1$ for MultiRoom-N12-S10.
All other hyperparameters are the same as those reported in \autoref{tab:hgrid_rcgvf_minigrid}. 

\begin{figure}[htpb]
\begin{subfigure}{.49\textwidth}
  \centering
  \includegraphics[width=1.0\textwidth]{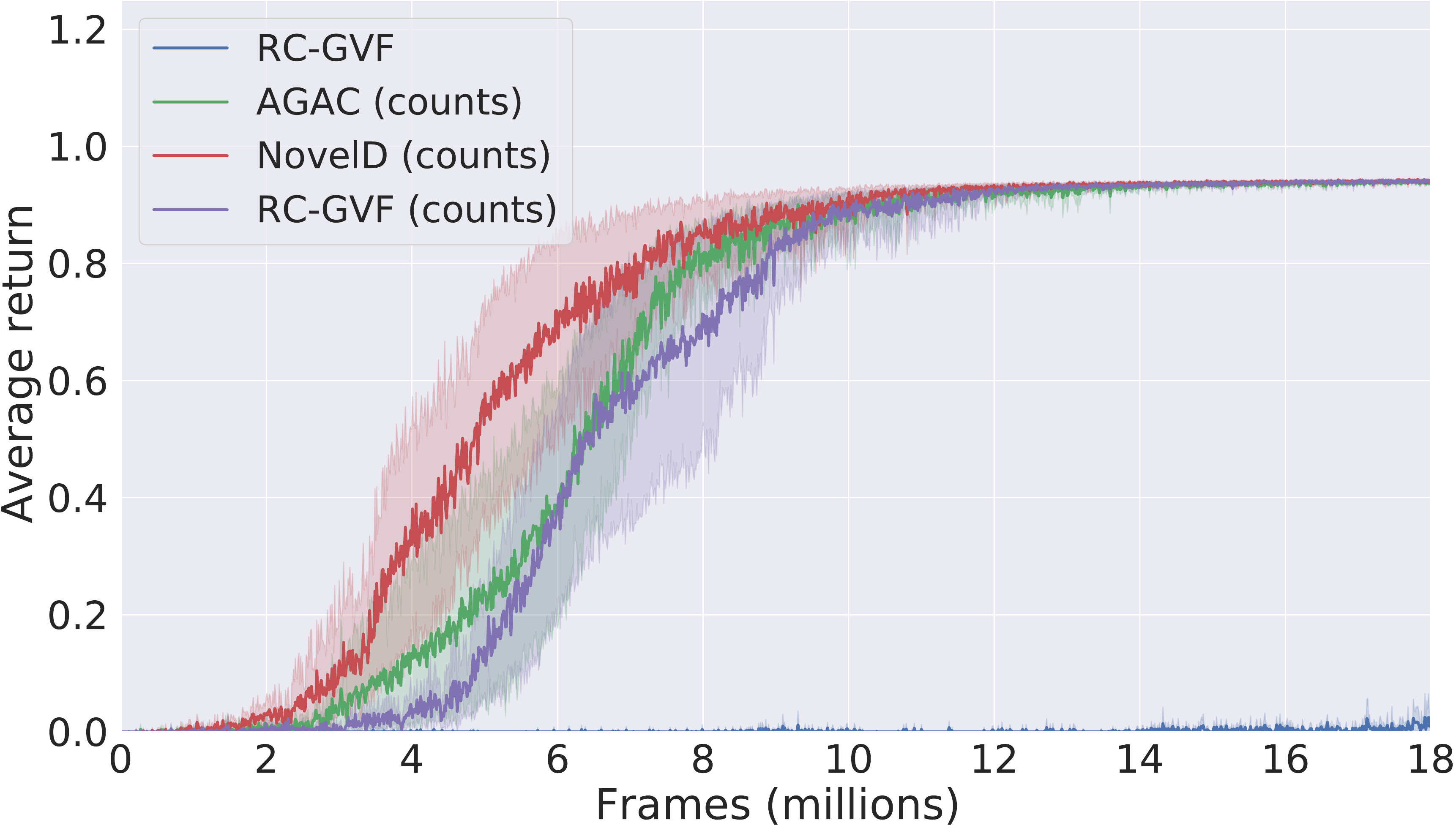}
  \caption{KeyCorridor-S5R3}
  \label{fig:minigrid_rcgvf_episodic_counts:kcs5r3}
\end{subfigure}
\begin{subfigure}{.49\textwidth}
  \centering
  \includegraphics[width=1.0\textwidth]{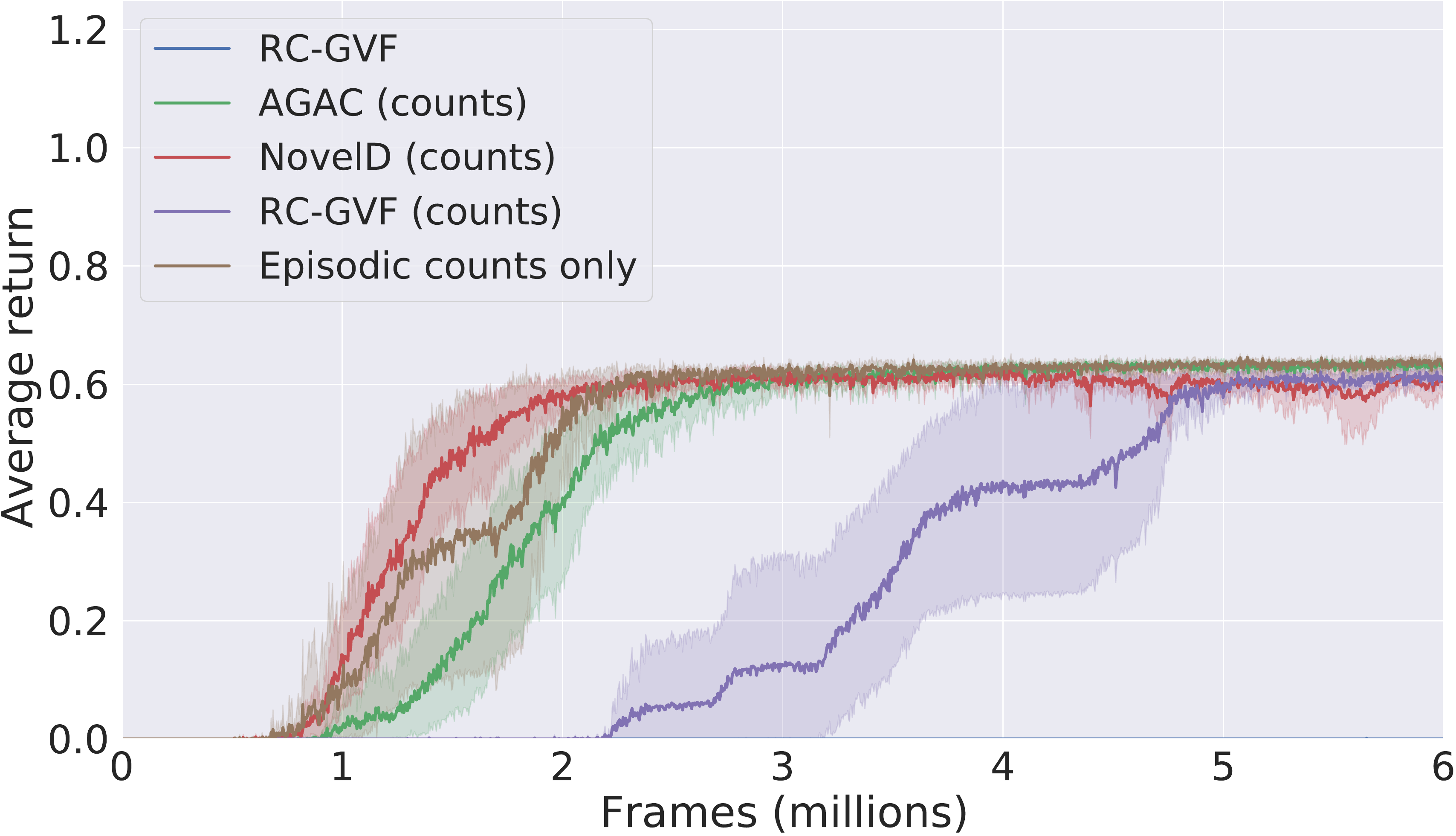}
  \caption{MultiRoom-N12-S10}
  \label{fig:minigrid_rcgvf_episodic_counts:mrn12s10}
\end{subfigure}
\caption{%
These results show that the performance of RC-GVF can also be improved through the use of episodic counts. All approaches use the standard egocentric observations. 95\% bootstrapped confidence intervals are shown for 10 seeds.
}
\label{fig:minigrid_rcgvf_episodic_counts}
\end{figure}

As shown in \autoref{fig:minigrid_rcgvf_episodic_counts}, RC-GVF can solve the harder environments (KeyCorridor-S5R3 and MultiRoom-N12-S10) without panoramic observations when including episodic counts.
In KeyCorridor-S5R3 we observe that RC-GVF adapted in this way is competitive with the baselines, while in MultiRoom-N12-S10 it converges more slowly.
This is unsurprising since we have not attempted to make our approach amenable to incorporating episodic counts.
Indeed, in MultiRoom-N12-S10, we observe that RC-GVF performs worse compared to when only using episodic state-counts (and no value function prediction), indicating that some interference takes place between these two signals.
Thus, we expect that other ways of combining (or weighting) these signals (such as by using an additive term for the state-counts as opposed to a multiplicative one) should lead to improved performance when also including state-counts.

\subsection{Improvement from history conditioning}
\label{sec:appendix_history_conditioning}

\begin{figure}[htpb]
\centering
    \begin{floatrow}[1]
      \ffigbox{\caption{Studying the performance of RC-GVF ($\gamma_z=0$) without the disagreement term on the KeyCorridor-S5R3 environment. We see that our recurrent predictor conditioned on histories of observations, actions and pseudo-rewards performs better than one conditioned on just histories of observations. 95\% bootstrapped confidence intervals are shown for 10 seeds. }\label{fig:minigrid:kcs5r3_rnn_ablation}}
      {
        \includegraphics[width=0.5\textwidth]{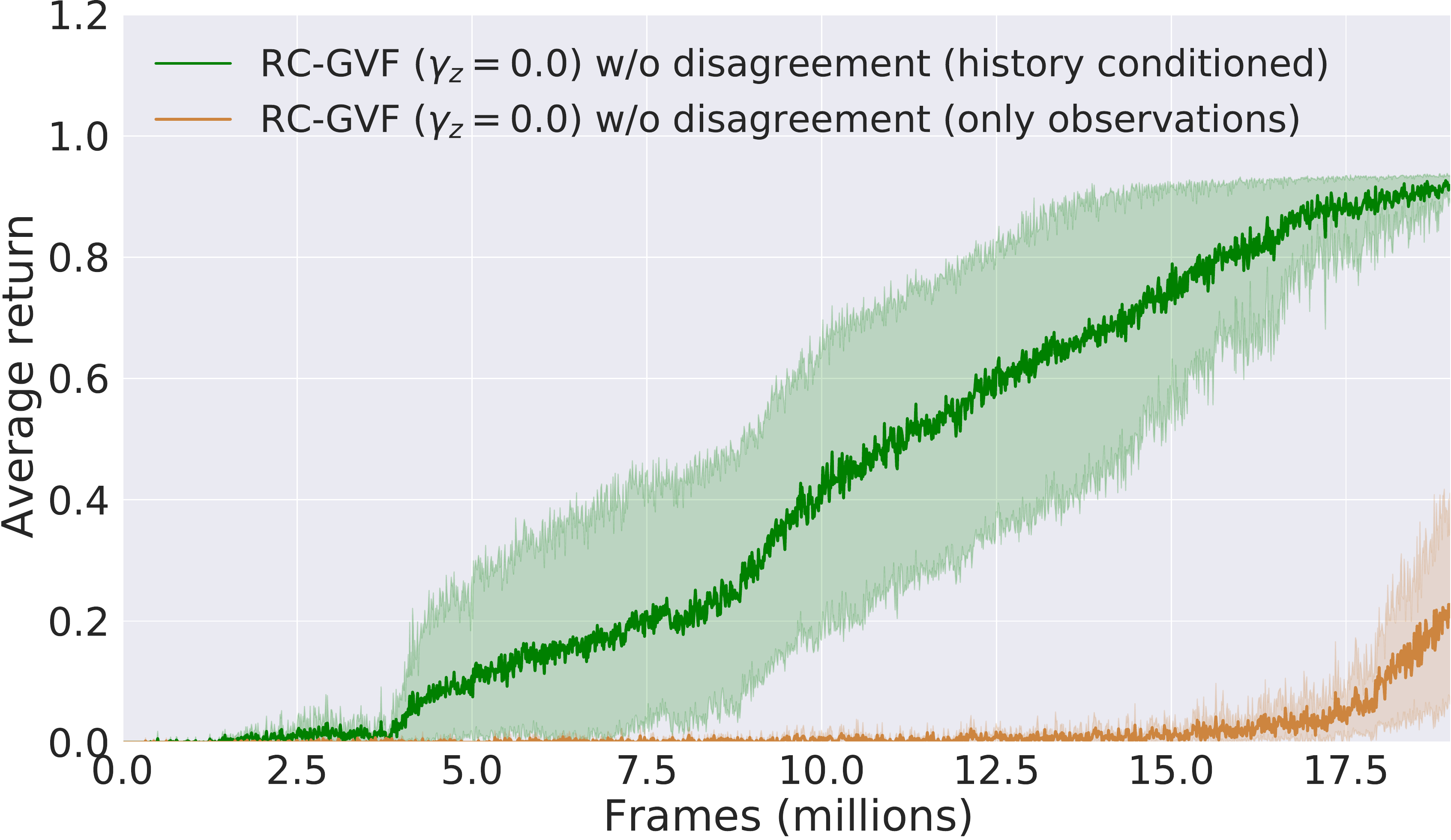}
      }
      \end{floatrow}
\end{figure}
As indicated in \autoref{fig:minigrid:kcs5r3_ablation}, the recurrent predictor plays an important role in the functioning of RC-GVF.
Here we present a result which shows that a recurrent predictor conditioned on histories of observations, actions and pseudo-rewards outperforms a recurrent predictor operating on histories of observations.
\autoref{fig:minigrid:kcs5r3_rnn_ablation} compares the performance of the two variants on the KeyCorridor-S5R3 environment.
We hypothesise that the improved results with history conditioning arise from previous pseudo-rewards being made available to the predictor. In a procedurally generated environment, having access to previous pseudo-rewards could support predictions of the new input observations on which the predictor has not yet been trained.

\subsection{Ensemble size analysis}
\label{sec:appendix_ensemble size analysis}
\begin{figure}[htpb]
\begin{subfigure}{.49\textwidth}
  \centering
  \includegraphics[width=1.0\textwidth]{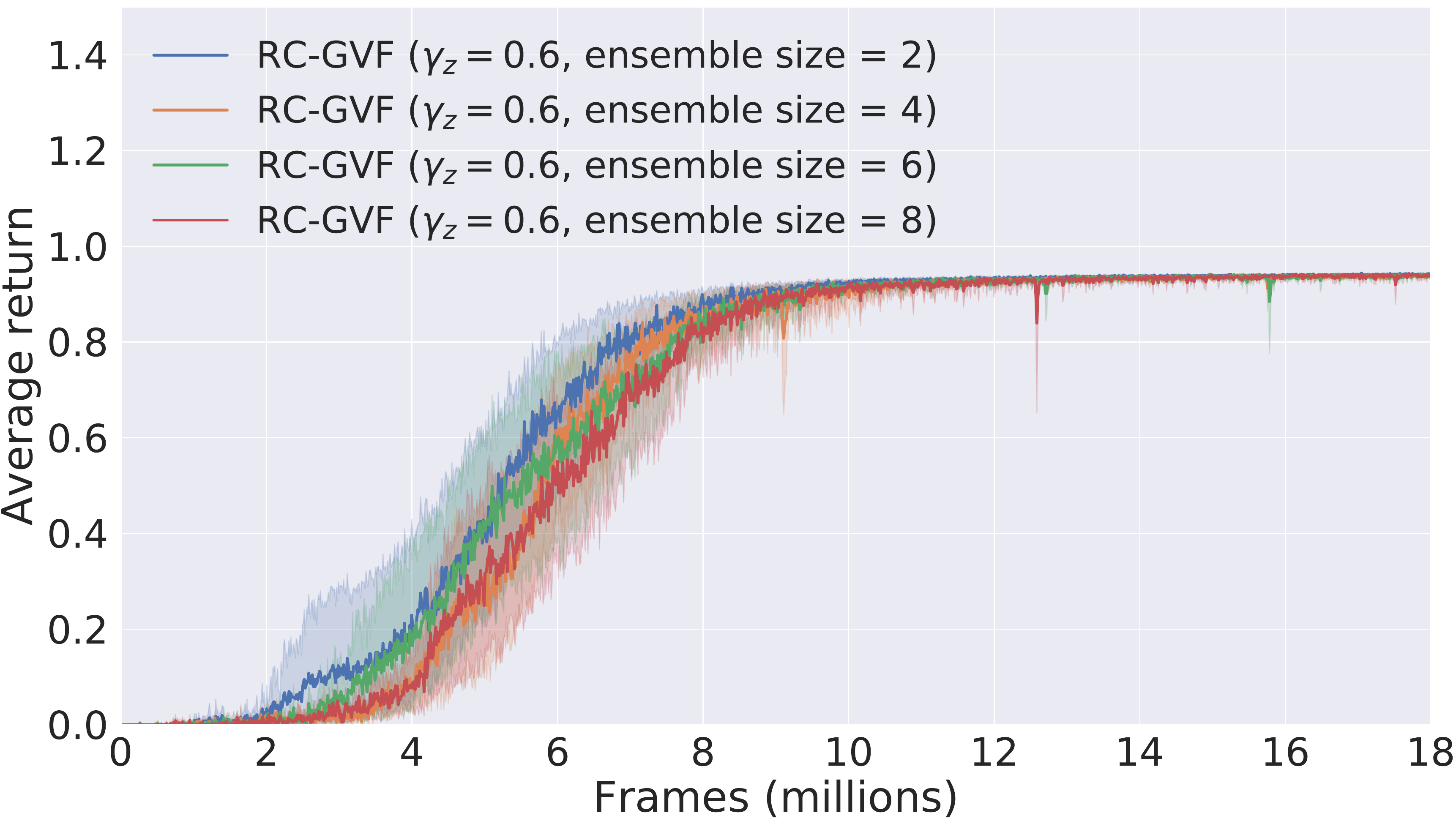}
  \caption{KeyCorridor-S5R3}
  \label{fig:minigrid_rcgvf_ensemble_size:kcs5r3}
\end{subfigure}
\begin{subfigure}{.49\textwidth}
  \centering
  \includegraphics[width=1.0\textwidth]{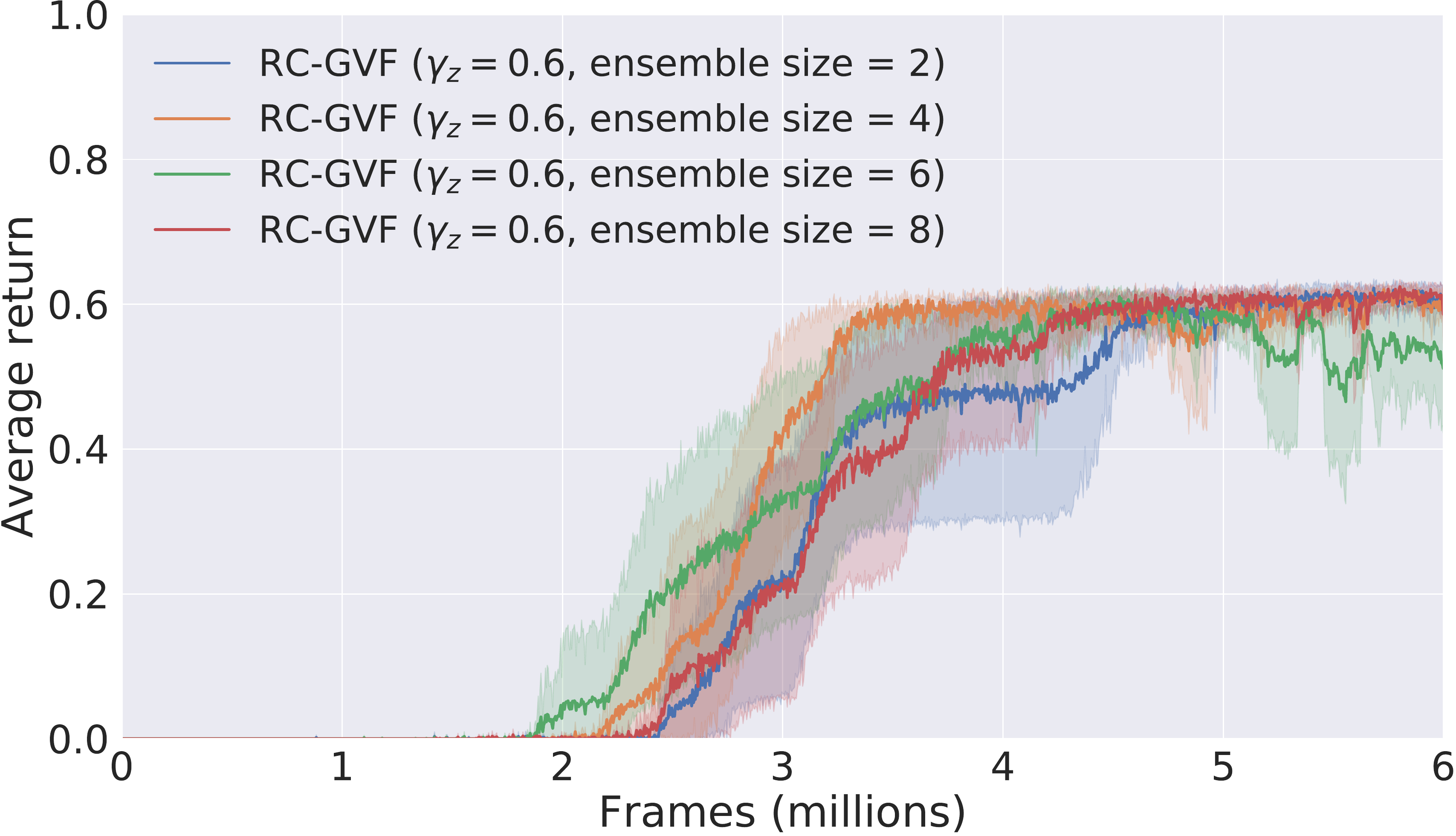}
  \caption{MultiRoom-N12-S10}
  \label{fig:minigrid_rcgvf_ensemble_size:mrn12s10}
\end{subfigure}
\caption{%
Comparison of RC-GVF with different numbers of predictors in the ensemble.
95\% bootstrapped confidence intervals are shown for 10 seeds.
}
\label{fig:minigrid_rcgvf_ensemble_size}
\end{figure}

\autoref{fig:minigrid_rcgvf_ensemble_size} shows the performance of larger ensemble sizes with RC-GVF in two MiniGrid environments (KeyCorridor-S5R3 and MultiRoom-N12-S10).
We observe similar results for ensemble sizes of 2, 4, 6 and 8 predictors.
The intrinsic coefficient $\beta=1$ in KeyCorridor-S5R3 and $\beta=0.5$ in MultiRoom-N12-S10. The remaining hyperparameters are the same as the ones in \autoref{tab:hgrid_rcgvf_minigrid_pano}.

These results indicate that two member ensembles usually suffice in these problems. We speculate how measuring disagreement with an ensemble of two predictors might be sufficient when also including the prediction error term.
However, note that the ensemble members have a common recurrent core and only differ in MLP heads; perhaps having separate RNNs for each member would lead to greater variation with ensemble size.
\end{document}